\documentclass[10pt]{article}

\usepackage[preprint]{rlc}

\usepackage[utf8]{inputenc} 
\usepackage[T1]{fontenc}    
\usepackage{url}            
\usepackage{booktabs}       
\usepackage{amsfonts}       
\usepackage{nicefrac}       
\usepackage{microtype}      
\usepackage{xcolor}
\usepackage{tcolorbox}[1]
\newtcolorbox{mybox}{colback=red!5!white,colframe=black}
\usepackage{amsmath} 
\usepackage{amssymb}  
\usepackage{dsfont}

\usepackage{times}
\usepackage{helvet}
\usepackage{courier}
\usepackage{url}
\usepackage{blkarray}
\usepackage{bbm}

\usepackage{amsthm}
\usepackage[font=small]{caption}
\usepackage{natbib}
\usepackage[capitalize]{cleveref}
\usepackage{autonum}
\usepackage{mathtools}
\usepackage{siunitx}
\usepackage{enumerate}
\usepackage{siunitx}
\usepackage{dsfont}
\usepackage[english]{babel}
\usepackage[parfill]{parskip}
\usepackage{bm}
\usepackage{mathrsfs}
\usepackage{multicol}
\usepackage{relsize}
\usepackage{nicefrac}
\usepackage{float}

\usepackage{pdflscape}
\usepackage{afterpage}
\usepackage{graphicx}
\usepackage{adjustbox}
\usepackage{makecell}
\usepackage{array}
\usepackage{subcaption}
\usepackage{tablefootnote}
\usepackage[flushleft]{threeparttable}
\usepackage{multirow}
\usepackage{wrapfig}
\graphicspath{{Figures/}}

\usepackage{algorithm}
\usepackage{algcompatible}
\usepackage{changepage}
\usepackage{bbm}

\newtheorem{assumption}{Assumption}

\newtheorem{theorem}{Theorem}
\newtheorem{lemma}{Lemma}

\DeclareMathOperator*{\argmax}{arg\,max}

\DeclarePairedDelimiterX{\infdivx}[2]{(}{)}{%
	#1\;\delimsize\|\;#2%
}

\usepackage{tikz}
\usetikzlibrary{arrows.meta,calc,fit,decorations.pathreplacing, hobby}
\usetikzlibrary{automata} 
\usetikzlibrary{positioning} 
\usetikzlibrary{arrows} 

\usepackage{scrwfile}
\TOCclone[\appendixname]{toc}{atoc}
\newcommand\StartAppendixEntries{}
\AfterTOCHead[toc]{%
  \renewcommand\StartAppendixEntries{\value{tocdepth}=-10000\relax}%
}
\AfterTOCHead[atoc]{%
  \edef\maintocdepth{\the\value{tocdepth}}%
  \value{tocdepth}=-10000\relax%
  \renewcommand\StartAppendixEntries{\value{tocdepth}=\maintocdepth\relax}%
}
\newcommand*\appendixwithtoc{%
  \appendix
  \renewcommand{\baselinestretch}{1.3}\normalsize  
  \addtocontents{toc}{\protect\StartAppendixEntries}
  \listofatoc
  \renewcommand{\baselinestretch}{1.0}\normalsize 
}

\title{Evolution Strategies at the Hyperscale}

\author{Bidipta Sarkar$^*$$^{1,2}$, Mattie Fellows$^*$$^{1}$, Juan Agustin Duque$^*$$^{2,3}$,\\ \textbf{Alistair Letcher$^\dagger$$^{1}$, Antonio León Villares$^\dagger$$^{1}$, Anya Sims$^\dagger$$^{1}$, Clarisse Wibault$^\dagger$$^{1}$,}
\\ \textbf{Dmitry Samsonov$^\dagger$$^{6}$,   Dylan Cope$^\dagger$$^{1}$, Jarek Liesen$^\dagger$$^{1}$, Kang Li$^\dagger$$^{1}$, Lukas Seier$^\dagger$$^{1}$, Theo Wolf$^\dagger$$^{1}$, }
\\ \textbf{Uljad Berdica$^\dagger$$^{1}$, Valentin Mohl$^\dagger$$^{1}$,}\\
\textbf{Alexander David Goldie$^{1,2}$, Aaron Courville$^{3,5}$, Karin Sevegnani$^4$,}\\
\textbf{Shimon Whiteson$^\ddagger$$^{2}$, Jakob Nicolaus Foerster$^\ddagger$$^{1}$.}\\
$^1$ FLAIR - University of Oxford, $^2$ WhiRL - University of Oxford, $^3$ MILA– Qu\'ebec AI Institute\\
$^4$ NVIDIA AI Technology Center, $^5$ CIFAR AI Chair, $^6$ NormaCore.dev\\
\texttt{\{bidipta.sarkar,matthew.fellows,jakob.foerster\}@eng.ox.ac.uk} \\
\texttt{juan.duque@mila.quebec},\ \texttt{shimon.whiteson@cs.ox.ac.uk} \\
}

\begin{document}
	
	\maketitle
    \def\thefootnote{*}\footnotetext{Equal Contribution\quad{$^\dagger$} Core Contributor, sorted by alphabetical order in first names\quad$^\ddagger$Equal Senior Authors}

    \begin{abstract}
Evolution Strategies (ES) is a class of powerful black-box optimisation methods that are highly parallelisable and can handle non-differentiable and noisy objectives. However, naïve ES becomes prohibitively expensive at scale on GPUs due to the low arithmetic intensity of batched matrix multiplications with unstructured random perturbations. We introduce Evolution Guided GeneRal Optimisation via Low-rank Learning (EGGROLL), which improves arithmetic intensity by structuring individual perturbations as rank-$r$ matrices, resulting in a hundredfold increase in training speed for billion-parameter models at large population sizes, achieving up to 91\% of the throughput of pure batch inference. We provide a rigorous theoretical analysis of Gaussian ES for high-dimensional parameter objectives, investigating conditions needed for ES updates to converge in high dimensions. Our results reveal a linearising effect, and proving consistency between EGGROLL and ES as parameter dimension increases. Our experiments show that EGGROLL: (1) enables the stable pretraining of nonlinear recurrent language models that operate purely in integer datatypes, (2) is competitive with GRPO for post-training LLMs on reasoning tasks, and (3) does not compromise performance compared to ES in tabula rasa RL settings, despite being faster. Our code is available at \url{https://eshyperscale.github.io/}.
    
\end{abstract}

\begin{figure}[htb!]
    \centering

    \vspace{1.5ex}

        \resizebox{0.9\linewidth}{!}{
        \begin{tikzpicture}[
          font=\normalsize,
          >=Latex,
          node distance=14mm and 18mm,
          block/.style={
            draw, rounded corners, thick, align=center,
            inner sep=5pt,
            minimum width=28mm, minimum height=10mm,
            fill=white
          },
          headImg/.style={
            inner sep=0pt,
            outer sep=0pt,
            minimum size=0pt
          },
          encImg/.style={
            inner sep=0pt,
            outer sep=0pt,
            minimum size=0pt
          },
          headEns/.style={
            inner sep=0pt,
            outer sep=0pt,
            minimum size=0pt
          },
          line/.style={->, thick}
        ]

        \def\NUMHEADS{4} 
        \def\STACKSTEP{12} 

        \newlength{\HeadSize}
        \setlength{\HeadSize}{30mm}

        \newlength{\RewardImgSize}
        \setlength{\RewardImgSize}{10mm} 

        \node[encImg] (enc) {%
          \includegraphics[
            width=\HeadSize,
            trim=20pt 20pt 20pt 20pt,
            clip
          ]{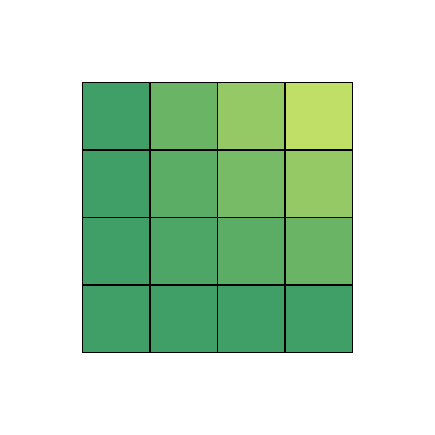}
        };

        \tikzset{
          ret/.style={
            align=center,
            inner sep=0pt,
            outer sep=0pt,
            font=\footnotesize
          },
        }

        \pgfmathsetmacro{\half}{(\NUMHEADS-1)/2.0}
        \foreach \i in {1,...,\NUMHEADS} {
          \pgfmathsetmacro{\shift}{(\half - (\i-1)) * \STACKSTEP}
          \pgfmathtruncatemacro{\imgindex}{\i-1} 

          \ifnum\i=3
            \node[headImg, right=37mm of enc, yshift=\shift mm] (pi\i) {\Huge$\vdots$};
            \node[ret, right=43mm of pi\i] (R\i) {\raisebox{0.3ex}{\Huge$\vdots$}};
          \else
            \node[headImg, right=25mm of enc, yshift=\shift mm] (pi\i) {%
              $\vcenter{\hbox{%
                \raisebox{1ex}{\includegraphics[width=\RewardImgSize, trim=30pt 30pt 30pt 50pt, clip]{Figures/Diagram/matrix_product_\imgindex.pdf}}%
                \;\raisebox{2.6ex}{$=$}\;%
                \includegraphics[width=\RewardImgSize]{Figures/Diagram/perturbation_\imgindex.pdf}%
              }}$%
            };

            \node[ret, right=15mm of pi\i] (R\i) {%
                $f\left(
                  \vcenter{\hbox{%
                    \includegraphics[width=\RewardImgSize]{Figures/Diagram/weight_matrix_v2.pdf}%
                    \;\raisebox{3.5ex}{$+$}\;%
                    \;\raisebox{3.5ex}{$\sigma$}\;%
                    \includegraphics[width=\RewardImgSize]{Figures/Diagram/perturbation_\imgindex.pdf}%
                  }}
                \right)$%
              };

            \draw[line] (enc) -- (pi\i);
            \draw[line] (pi\i) -- (R\i);
          \fi
        }

        \usetikzlibrary{backgrounds}

        \coordinate (topH) at (pi1.north);
        \coordinate (botH) at (pi\NUMHEADS.south);

        \coordinate (Rref) at (R1.east);

        \def\GAUSSAMP{12mm}
        \def\GAUSSSIG{0.22}
        \def\GAUSSOFF{8mm}

        \coordinate (gBase) at ($(Rref)+(\GAUSSOFF,0)$);

        \draw[very thick]
          let \p1=(topH), \p2=(botH), \p3=(gBase) in
          plot[
            domain=0:1,
            samples=160,
            smooth,
            variable=\q 
          ]
          ({\x3 + \GAUSSAMP*exp(-((\q-0.5)^2)/(2*(\GAUSSSIG)^2))},
           {\y2 + \q*(\y1-\y2)});

        \draw[densely dashed] (gBase|-botH) -- (gBase|-topH);

        \coordinate (midH) at ($(topH)!0.5!(botH)$);
        \coordinate (gaussRight) at ($(gBase)+(\GAUSSAMP,0)$);
        \def\ENSEMBLEOFF{25mm}
        \coordinate (ensX) at ($(gaussRight)+(\ENSEMBLEOFF,0)$);

        \node[headEns] (piEns) at (ensX|-midH){
          \includegraphics[
            width=\HeadSize,
            trim=20pt 20pt 20pt 20pt,
            clip
          ]{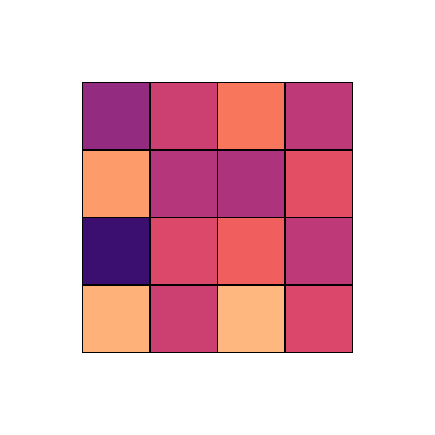}
        };

        \def\GPEAKYSHIFT{-18.2mm} 
        \coordinate (gPeak) at ($($(gBase)+(\GAUSSAMP,0)$ |- midH)+(0,\GPEAKYSHIFT)$);
        \draw[line] (gPeak) -- (piEns.west);


        \node[above=3mm of enc] {Initial weights};

        \node[above=3mm of pi1] {Rank-one perturbation $E_i$};

        \node[above=4mm of R1] {Fitness evaluation};

        \coordinate (gaussMid) at ($(gBase)!0.5!(gaussRight)$);
        \node[above=8.5mm of gaussMid] {Weighted average};

        \node[above=3mm of piEns] {Final rank-$N$ update};

        \end{tikzpicture}
        }

    \caption{Schematic visualisation of EGGROLL using $N$ workers. }
    \label{fig:header_figure}
\end{figure}

\begin{figure}[htb!]
\centering
    \begin{subfigure}[t]{0.36\linewidth}
        \centering
        \includegraphics[width=\linewidth]{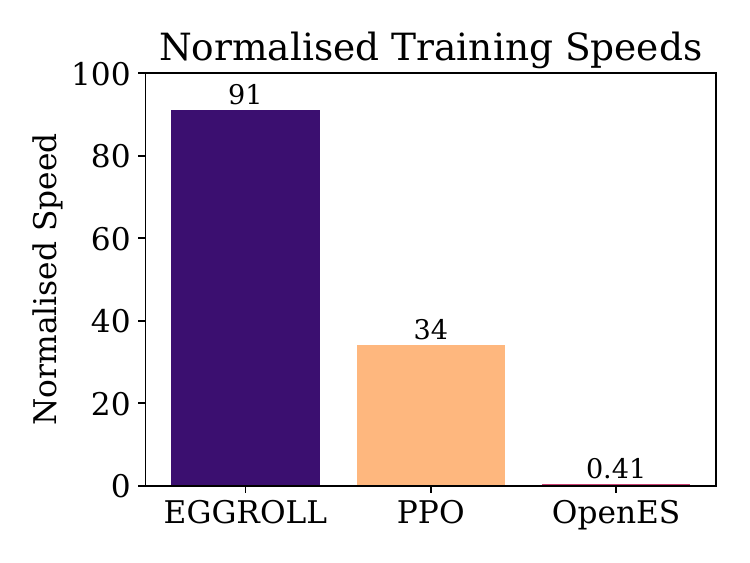}
        \phantomsubcaption
        \label{fig:batch_timings}
    \end{subfigure}
      \hfill
    \begin{subfigure}[t]{0.54\linewidth}
        \centering
        \includegraphics[width=\linewidth]{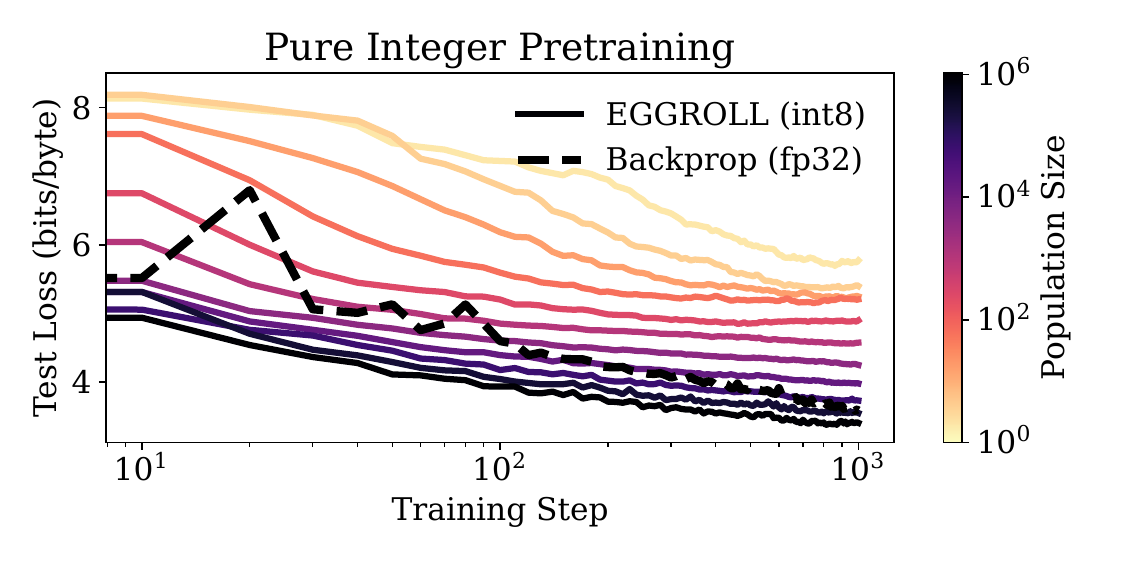}
        \phantomsubcaption
        \label{fig:pretrain_loss}
    \end{subfigure}
  \vspace{-0.5cm}
    \caption{(a) Relative speed of our method, EGGROLL, in terms of experience throughput versus prior methods, where 100 is the maximum batch inference throughput. See Appendix~\ref{sec:speed_appendix} for more details. (b) We use EGGROLL to train an int8 RNN language model from scratch, scaling population size from 2 to 1,048,576 with a fixed data batch size of 16. The dotted line is a fp32 Transformer trained with backprop SGD. EGGROLL's test next-token cross-entropy of 3.40 bits/byte while backprop only gets 3.58 bits/byte.}
  \vspace{-0.4cm}
\end{figure}
    \section{Introduction}
\label{sec:intro}
Evolution Strategies (ES) \citep{Rechenberg78, Beyer1995,Beyer02} is an attractive alternative to first-order methods based on gradient backpropagation for several reasons. First, ES does not require differentiability; it can optimise a broader class of models, like those with discrete parametrisations (cellular automata) or objectives for which gradients are unavailable or noisy, such as outcome-only rewards in LLM fine-tuning \citep{qiu2025evolutionstrategiesscalellm}. Second, ES can be more robust to noisy and ill-conditioned optimisation landscapes \citep{wierstra2011naturalevolutionstrategies,ijcai2021p443}. Population-based exploration smooths irregularities \citep{salimans2017}, tolerates discontinuities, and mitigates issues like ill-conditioned curvature or vanishing and exploding gradients in long-range or recurrent settings \citep{hansen2023cmaevolutionstrategytutorial}. Third, ES is highly amenable to parallel scaling, since fitness evaluations are independent across population members and require only the communication of scalar fitnesses, which maps cleanly onto modern inference infrastructure and yields near-linear speedups on large clusters \citep{salimans2017}. By contrast, backpropagation requires communicating and aggregating gradients across devices, yielding updates with high memory and computational costs. Furthermore, backpropagation requires special care when training models with low-precision datatypes~\citep{fishman2025scalingfp8trainingtrilliontoken}, whereas ES can directly optimise any model with the same datatypes used at inference time. Together, these properties position ES as a potentially powerful tool for training large, discrete, or hybrid architectures, and end-to-end systems with non-differentiable components, including LLMs \citep{brown2020language,Chowdhery2022PaLM,Du2022GLaM,Fedus2022SwitchTransformers}.

However, there are currently practical obstacles to employing ES at scale.
In deep learning architectures \citep{Goodfellow-et-al-2016}, the majority of trainable parameters form linear mappings represented by matrices \citep{Rosenblatt62,Hochreiter96,Bengio00a, Krizhevsky2012,Goodfellow14,kingma14,Vaswani2017}. Na{\"i}vely adapting ES therefore requires generating full-rank matrix perturbations that replicate the entire parameter set for every population member. This inflates memory costs and forces frequent movement of large weight tensors. Evaluating these perturbations then requires a separate sequence of matrix multiplications per member, so the total compute and wall-clock time scale roughly with the population size and sequence length since batched matrix multiplication has a low arithmetic intensity, i.e., the ratio of arithmetic operations to memory traffic~\citep{arithmeticintensity}. In billion-parameter regimes, these two costs dominate, limiting ES to small models and small populations \citep{qiu2025evolutionstrategiesscalellm, korotyshova2025essaevolutionarystrategiesscalable}.

To mitigate both memory and computational bottlenecks, we introduce Evolution Guided GeneRal Optimisation via Low-rank Learning (EGGROLL), an ES algorithm that allows for the efficient training of neural network architectures with billions of parameters. Analogous to LoRA’s low-rank adapters in gradient-based training \citep{hu2022lowrank}, EGGROLL generates \emph{low-rank} parameter-space perturbations for ES; instead of sampling a full-rank matrix \(E\in\mathbb{R}^{m\times n}\), we sample \(A\in\mathbb{R}^{m\times r}\) and \(B\in\mathbb{R}^{n\times r}\) with \(r\ll \min(m,n)\) and form \(E=\frac{1}{\sqrt{r}}AB^{\top}\). This reduces auxiliary perturbation matrix storage from \(mn\) to \((m+n)r\) per layer, and proportionally reduces tensor movement. 

Moreover, we use a counter-based deterministic random number generator (RNG) \citep{threefry, jax2018github} to reconstruct noise on demand, so matrix perturbations need not persist in memory. When evaluating the fitness of members of multiple perturbations in parallel, EGGROLL batches a population of low-rank adapters and shares the base activations, enabling a single forward pass that applies all \(AB^{\top}\) updates via specialised batched matrix multiplications with significantly higher arithmetic intensity, resulting in over a hundredfold increase in training throughput for large neural networks at large population sizes, as shown in~\cref{fig:batch_timings}. Crucially, EGGROLL does not restrict updates to be low-rank, as the overall update is a weighted average of rank $r$ matrices across the population, making the matrix parameter update rank $\min(Nr,m,n)$ . 

To understand ES when applied to large parameter models, we analyse the convergence properties of general Gaussian ES in high dimensions, showing there exists a critical noise scaling $\sigma_d = o(d^{-1/2})$ under which the update provably linearises and converges to the first-order derivative for a broad class of (possibly discontinuous) objectives. We identify three distinct regimes---linearisation, critical, and divergence---and establish provably tight conditions for stable ES optimisation in large models. Building on this, we extend the analysis to EGGROLL and prove that even fixed low-rank updates (including rank-1) converge to the true ES gradient as dimension grows, despite heavier-tailed perturbations. Our results explain the empirical success of EGGROLL in high-dimensional neural networks and connect its behaviour to neural tangent kernel-style linearisation \citep{Jacot18}, yielding explicit convergence rates under standard overparameterised regimes. We also provide a rigorous theoretical analysis of the low-rank approximation accuracy, proving that EGGROLL updates converge to the full-rank Gaussian ES updates at a fast $\mathcal{O}(r^{-1})$ rate. 

Furthermore, in our extensive empirical evaluation, we test this hypothesis across a wide range of domains. In tabula rasa and multi-agent RL (MARL) settings, we show that EGGROLL does not compromise performance compared to na\"ive ES despite being faster. We demonstrate the scalability of EGGROLL for LLM fine-tuning with experiments on pretrained RWKV7~\citep{peng2025rwkv7gooseexpressivedynamic} models, modern recurrent language models that enable large batch inference due to their constant state size. Finally, we develop a nonlinear RNN language model that operates purely in integer datatypes, and demonstrate that EGGROLL can stably pretrain this language model, a feat which is only feasible due to the large population sizes enabled by EGGROLL.

    \section{Preliminaries}

\subsection{Low-Rank Matrix Approximations}
\label{sec:low_rank_networks}
When adapting high-dimensional foundation models for specific tasks, updating the parameters using gradient-based methods has high memory requirements. LoRA~\citep{hu2022lowrank} applies low-rank approximations to the matrix multiplications to reduce these costs. For each matrix $M_i\in\mathbb{R}^{m\times n}$ in the model, a low-rank approximation can be made by decomposing each matrix:
\begin{align}
    M_i\approx M_i^0+A_i B_i^\top, 
\end{align}
 where $M_i^0\coloneq \textrm{StopGrad}(M_i)$ is the imported matrix from the foundation model with frozen parameters and $A_i\in \mathbb{R}^{m\times r}$ and $B_i\in \mathbb{R}^{n\times r}$ are low-width column matrices (i.e., $r\ll \min(m,n)$) whose parameters are updated through gradient-based optimisation during task-specific adaptation. This reduces the number of optimisation parameters for each matrix  from $mn$ to $r (m+ n)$. EGGROLL uses a similar low-rank approximation for evolutionary strategies.

\subsection{Evolution Strategies}
Evolution strategies (ES) \citep{Rechenberg78, Beyer1995,Beyer02} is a set of black-box optimisation methods that has emerged as a useful alternative to first-order gradient-based methods like stochastic gradient descent (SGD), particularly for noisy or non-differentiable systems. Let $f:\mathbb{R}^d\rightarrow \mathbb{R}$ denote an objective to be optimised, known as the $\emph{fitness}$, where the goal is to find an optimising set of parameters $x^\star\in\argmax_{x\in\mathbb{R}^d} f(x)$. Each set of  parameters is collected into a $d$-dimensional vector known as a genotype. We denote the derivative of the fitness $\nabla_x f(x)\vert_{x=a}$ evaluated at $x=a$ as $\nabla f(a)$. Unlike first-order gradient-based methods, which query derivatives $\nabla f(x)$ to update the vector of parameters $x$, evolutionary methods update a parametric population distribution over the fitness parameter space $\pi(x\vert \theta)$, which is smoothly parametrised by a separate set of parameters $\theta\in \Theta$. The population distribution generates perturbations $x\sim \pi(x\vert \theta)$ known as mutations. The problem of optimising the fitness $f(x)$ for $x$ reduces to optimising the parameters of the population distribution $\theta$. This is achieved by solving a \emph{secondary} optimisation problem to maximise the expected fitness under $\pi(x\vert \theta)$ for $\theta$:
\begin{align}
    J(\theta)=\mathbb{E}_{x\sim \pi(x\vert \theta)}\left[ f(x) \right].\label{eq:gauss_es_objective}
\end{align}
Introducing a population distribution \emph{smooths} the fitness landscape; since $\pi(x\vert \theta)$ is smooth in $\theta$, the resulting objective $J(\theta)$ is also smooth in $\theta$, provided $f(x)$ is measurable and integrable but not necessarily differentiable.  Evolution strategies can therefore optimise black-box problems that may be non-differentiable as the derivatives of $J(\theta)$ exist for  fitness functions that are discontinuous, yielding a gradient with respect to $\theta$: 
\begin{align}
    \nabla_\theta J(\theta)=\mathbb{E}_{x\sim \pi(x\vert \theta)}\left[\nabla_{\theta} \log \pi(x\vert \theta)f(x)\right],\label{eq:es_grad}
\end{align}
where $\nabla_{\theta} \log \pi(x\vert \theta)$ is known as the score function. A Monte Carlo estimate  is formed by sampling $N$ search mutations $x_i\sim \pi(x_i\vert \theta)$ and computing an average of the score-weighted fitnesses: 
\begin{align}
    \hat{\nabla}_\theta J(\theta)=\frac{1}{ N}\sum_{i=1}^N\nabla_\theta \log \pi(x_i\vert \theta)f(x_i),\label{eq:monte_carlo}
\end{align}
with which we update $\theta$ via stochastic gradient ascent with a suitable stepsize $\alpha_t$: 
\begin{align}
    \theta_{t+1}\gets \theta_t+\alpha_t \hat{\nabla}_\theta J(\theta_t).
\end{align}

 ES does not require taking derivatives directly through the fitness function; instead the Monte Carlo update in \cref{eq:monte_carlo}  only requires evaluation of $f(x_i)$ for each mutation $x_i$ to estimate $\nabla_\theta J(\theta)$. As ES only queries $f(x)$ and not $\nabla f(\mu)$, it is a \emph{zeroth-order} optimisation method. 

 In this paper, we study ES using Gaussian population distributions:
$\pi(x\vert \theta)= \mathcal{N}(\mu,I_d\sigma^2)$. In addition to its mathematical convenience, the central limit theorem means that the Gaussian distribution emerges naturally from the EGGROLL low-rank approximation as rank increases, even if the matrices $A$ and $B$ are themselves non-Gaussian. Moreover, most widely-used ES algorithms assume Gaussian population distributions \citep{Rechenberg78,schwefel1995evolution,hansen2001completely,Beyer02,auger2011theory,wierstra2011naturalevolutionstrategies,salimans2017}. In our setting, ES optimises over the population mean $\mu\in\mathbb{R}^d$, which acts as a proxy for the true maximum of the fitness function, and the variance parameter $\sigma^2\ge 0$ is treated as a hyperparameter to be tuned.  

For the Gaussian population distribution we study in this paper, the ES update can be written using an expectation under a standard normal distribution by making a transformation of variables $v= \frac{x-\mu}{\sigma}$ \citep{wierstra2011naturalevolutionstrategies,salimans2017}:
\begin{align}
    \nabla_\mu J(\theta)&=-\frac{1}{\sigma}\mathbb{E}_{v\sim \mathcal{N}(0,I_d)}\left[\nabla_v \log p(v)\cdot f(\mu+\sigma v)\right],\\
    &=\frac{1}{\sigma}\mathbb{E}_{v\sim \mathcal{N}(0,I_d)}\left[v\cdot f(\mu+\sigma v)\right],\label{eq:standard_normal}
\end{align}
where $v\sim  P(v)=\mathcal{N}(0,I_d)$ and $p(v)$ denotes the density of $P(v)$. In this form, \cref{eq:standard_normal} shows that Gaussian ES methods optimise the fitness by generating search vectors from a standard normal distribution $\mathcal{N}(0,I_d)$ around the mean parameter $\mu$.

\subsection{Evolution Strategies for Matrix Parameters}\label{sec:matrix_es}
A key focus of this paper is to develop efficient methods for evolution strategies that target \emph{matrix parameters}. When working in matrix space, it is convenient to use the matrix Gaussian distribution \citep{Dawid81}, which is defined directly over matrices $X\in \mathbb{R}^{m\times n}$:
\begin{align}
    \mathcal{N}(M,U,V)=\frac{1}{(2\pi)^\frac{mn}{2}\det( U)^\frac{n}{2}\det(V)^\frac{m}{2} }\exp\left(-\frac{1}{2} \textrm{tr}\left(V^{-1}(X-M)^\top U^{-1}(X-M)\right)\right),\label{eq:gaussian_matrix}
\end{align}
where $M\in\mathbb{R}^{m\times n}$ is the mean matrix, $U\in \mathbb{R}^{m\times m}$ is the row covariance matrix and $V\in \mathbb{R}^{n\times n}$ is the column covariance matrix. We use $\textrm{vec}(\cdot)$ to denote the vectorisation operator:
\begin{align}
    \textrm{vec}(X)\coloneqq[x_{1,1},\dots x_{m,1},x_{1,2},\dots x_{m,n}]^\top.
\end{align}
The matrix Gaussian distribution is a generalisation of the multivariate Gaussian distribution $\mathcal{N}(\mu,\Sigma)$ defined over vector space. Sampling a matrix $X\sim \mathcal{N}(M,U,V)$ from a matrix Gaussian distribution is equivalent to sampling a vector $\textrm{vec}(X)\sim \mathcal{N}(\mu,\Sigma)$ from a multivariate Gaussian distribution  with mean $\mu=\textrm{vec}(M)$ and covariance matrix $\Sigma=V \otimes U$ where $\otimes$ denotes the Kronecker product. For isotropic matrix Gaussian distributions with covariance matrices $ U=\sigma I_m$ and $ V=\sigma I_n$, the equivalent multivariate Gaussian distribution is also isotropic with $\Sigma=\sigma^2 I_{mn}$. We denote the $\ell^2$ vector norm as $\lVert \cdot \rVert$ and to measure distance between matrices, we use the Frobenius norm:
\begin{align}
    \lVert M \rVert_F\coloneqq\sqrt{\sum_{i,j}\ {m_{i,j}}^2}=\left\lVert \textrm{vec}(M)\right\rVert,
\end{align}
 which provides an upper bound on the matrix 2-norm \citep{Petersen12}. Let $W\in\mathbb{R}^{m\times n}$ be a set of matrix parameters where $\textrm{vec}(W)$ forms a subset of the full parameter vector $x$, typically parametrising the weights of a linear layer in a neural network. As we derive in \cref{app:es_matrix}, the Gaussian ES update associated with the matrix $W$ is:
\begin{align}
    \nabla_M J(\theta)&=-\frac{1}{\sigma}\mathbb{E}_{E\sim P(E)}\left[\nabla_E \log p(E) \cdot f(W=M+\sigma E)\right],\\
    &= \frac{1}{\sigma}\mathbb{E}_{E\sim P(E)}\left[E \cdot f(W=M+\sigma E)\right],\label{eq:ES_update}
\end{align}
where $M$ is the mean matrix associated with $W$, i.e. $\textrm{vec}(M)$ forms a subset of $\mu$, and 
 $P(E)$ is a zero-mean standard normal matrix distribution: $p(E)=\mathcal{N}(0,I_m,I_n)$. The gradient in \cref{eq:ES_update} is estimated using the Monte Carlo estimate:
 \begin{align}
      \hat{\nabla}_M J(\theta)=\frac{1}{\sigma N}\sum_{i=1}^N E_i \cdot f(W=M+\sigma E_i),\label{eq:matrix_mc}
 \end{align}
 by sampling $N$ search matrices $E_i\sim P(E_i)$ from a standard matrix normal distribution $\mathcal{N}(0,I_m,I_n)$ around the mean parameter matrix $M$, which is updated via stochastic gradient ascent: 
 \begin{align}
     M_{t+1}\leftarrow M_t+\alpha_t \hat{\nabla}_M J(\theta_t).
 \end{align}

    \section{Related Work}

\subsection{Evolutionary Algorithms}

Evolutionary algorithms have long been a compelling alternative to backpropagation-based training methods (e.g., genetic algorithms \citep{such_deep_2018} or symbolic evolution \citep{koza_genetic_1994}). Much research in evolution has focused on developing algorithms for deep learning that scale well to distributed parallel computation \citep{jaderberg_population_2017, hansen_completely_2001, salimans2017}. These approaches have increased in popularity following the application of ES to policy learning in deep RL environments~\citep{salimans2017}. Since then, evolution has been widely applied in other domains, such as meta-learning (e.g., \citep{chris_ppo, metzVeLOTrainingVersatile2022,lange_discovering_2023, goldie_can_2024, goldie_how_2025}), hyperparameter tuning (e.g., \citep{parker-holder_provably_2021, tani_evolutionary_2021, vincent_improved_2023}), and drug discovery \citep{towers_adios_2025}. ES has also enabled the development of neural network architectures that are unsuitable for backpropagation, such as activation-free models that exploit floating point rounding error as an implicit nonlinearity~\citep{foerster2017nonlinear}. Here, we consider how to apply ES at a scale beyond the small networks and population sizes of prior work. For example,~\citet{salimans2017} use a maximum population size of 1440, whereas we use over a million.

While low-rank structures have been used in prior evolutionary algorithms, they have been applied to different ends, with different trade-offs, relative to EGGROLL. \citet{NEURIPS2019_88bade49} use a low-rank search space found via principal component analysis, which provides a better search direction to more efficiently use small populations. \citet{garbus2025lowrank} optimise a low-rank factorisation instead of the full dense matrix with neuroevolution, achieving similar computational gains to EGGROLL but is limited to the low-rank structure regardless of population size.

\vspace{-0.2cm}
\subsection{Evolution Strategies for LLMs}

Although gradient backpropagation is typically used for LLM training and fine-tuning, prior work explores ES variants for fine-tuning. In particular, \citet{zhang2024revisiting}'s two-point zeroth-order gradient estimator, which can be viewed as an ES-inspired method using a single perturbation direction and two function queries per update, is used by \citet{MeZO} for memory-efficient LLM fine-tuning. \citet{Yu_2025_ICCV} extend this approach by projecting perturbations to a low-rank subspace, improving convergence. \citet{10.1109/TASLP.2024.3477330} perform ES directly on LoRA matrices. These works focus on supervised fine-tuning and report performance comparable to full fine-tuning, but do not address whether pretraining is possible with two-point zeroth-order methods; we find that large population sizes are necessary for pretraining, indicating such methods are unsuitable here.

Recent work also explores ES in the context of LLM reasoning. \citet{korotyshova2025essaevolutionarystrategiesscalable} first train LoRA adapters using supervised fine-tuning (SFT) before decomposing them into fixed SVD bases alongside singular values that are trained using CMA-ES. They achieve comparable performance to GRPO~\citep{shao2024} in significantly less wall-clock time on maths reasoning benchmarks. \citet{qiu2025evolutionstrategiesscalellm} directly use ES to optimise all LLM parameters for reasoning, with stronger performance than GRPO on the countdown reasoning task. However, both of these approaches use relatively small population sizes, on the order of a hundred unique perturbations per update, and instead collect hundreds of rollouts per perturbation to efficiently use GPUs. By contrast, our approach allows all generations to use different perturbations, such that our maximum population size per update is orders of magnitude larger (equal to the maximum inference batch size), without compromising token generation throughput.

    \section{EGGROLL}\label{sec:eggroll}
\vspace{-0.1cm}
We now introduce EGGROLL (\cref{alg:eggroll}). A practical issue with using a low-rank matrix approximation is that its distribution and score function have no analytic solution except for degenerate cases, so in \cref{sec:low_rank_es} we derive the EGGROLL approximate score function from the limiting high-rank Gaussian. \cref{sec:eggroll_hardware} describes how to efficiently implement EGGROLL on modern hardware.
\vspace{-0.1cm}
\subsection{ Low-Rank Evolution Strategies}\label{sec:low_rank_es}
\begin{wrapfigure}{r}{0.46\textwidth}
\vspace{-1.5cm}
\begin{minipage}{0.45\textwidth}
\begin{algorithm}[H]
\caption{EGGROLL$(r,\alpha,\sigma,T_\textrm{max},N_\textrm{workers})$}  \label{alg:eggroll}
\footnotesize
\begin{algorithmic}
\STATE \textbf{initialise} $M$ and workers with known random seeds $\varsigma$ 
\FOR{ $T_\textrm{max}$ timesteps}
\FOR{each worker $i\in\{1,\dots N_\textrm{workers}\}$ in parallel}
\STATE $A_i\sim p(A_i), B_i\sim p(B_i)$ 
\STATE $E_i\gets\frac{1}{\sqrt{r}} A_i B_i^\top$
\STATE $f_i\gets f(W=M+\sigma E_i)$ 
\ENDFOR
\STATE workers share scalar fitness $f_i$ with other workers
\FOR{each worker $i\in\{1,\dots N_\textrm{workers}\}$ in parallel}
\STATE reconstruct $E_j$ for $j\in\{1,\dots N_\textrm{workers}\}$ from $\varsigma$  
\STATE $M\gets M+\alpha\frac{1}{N_\textrm{Workers}} \sum_{j=1}^{N_\textrm{Workers}} E_jf_j$
\ENDFOR
\ENDFOR
\end{algorithmic}
\end{algorithm}
\end{minipage}
\vspace{-0.4cm}
\end{wrapfigure}
Recall the Gaussian matrix ES update from \cref{eq:ES_update}. Our goal is to introduce a tractable approximation to generating full-rank matrices by using low-rank matrices $A B^\top$ as our search matrices instead. Let $p(A)$ and $p(B)$ denote the distribution of $A\in\mathbb{R}^{m\times r}$ and $B\in\mathbb{R}^{n\times r}$.
\begin{assumption}[I.I.D. Sampling]\label{ass:iid}
Assume all elements $a_{i,j}\in A$ and $b_{i,j}\in B$ are continuous, identically and independently distributed random variables according to some zero-mean, symmetric, absolutely continuous distribution $p_0(\cdot)$ with finite fourth-order moments and unit variance.
\end{assumption}
This assumption is easily satisfied for most perturbation distributions used by ES, including members from the set of generalised Gaussian distributions like Laplace, normal, and uniform distributions. We then form a low-rank search matrix: $E= \frac{1}{\sqrt{r}}A B^\top$. The $\frac{1}{\sqrt{r}}$ scaling ensures the variance of $E$ remains bounded for all $r$. We denote the induced distribution of $E$ as $P(E)$. $E= \frac{1}{\sqrt{r}}A B^\top$ maps to the manifold $\mathbb{M}^r\subset\mathbb{R}^{m\times n}$ of rank-$r$ matrices. Hence, the density $p(E)$ is defined with respect to a unit volume on the manifold and cannot be defined with respect to the standard unit volume in Euclidean space. For the corresponding score function, gradients with respect to $\log p(E)$ are not defined over the usual Euclidean space. Instead, we use an approximation $\hat{S}(E):\mathbb{R}^{m\times n}\rightarrow \mathbb{R}^{m\times n}$ for the score function, yielding our low-rank update:
 \begin{align}
     \hat{g}_\textrm{LR}=-\frac{1}{\sigma}\mathbb{E}_{E\sim p(E)}\left[  \hat{S}(E) \cdot f(W=M+\sigma E) \right].\label{eq:approx_lr_grad}
 \end{align}
In our experiments, analysis and \cref{alg:eggroll},  we use a Gaussian approximate score function:
\begin{align}
    \hat{S}(E)= -E,\label{eq:gaussian_approx}
\end{align}
which is the score function for the Gaussian distribution $\mathcal{N}(0,I_m,I_n)$. This choice is motivated by two theoretical insights from \cref{sec:analysis}. The matrix $A B^\top $ can be decomposed as a sum of independent, zero-mean vector outer products. Under Assumption \ref{ass:iid}, the central limit theorem applies to this sum of variables, proving that $P(E)$ converges in distribution to a Gaussian $\mathcal{N}(0,I_m,I_n)$ as rank $r$ increases, recovering the approximate Gaussian score in the limit. Secondly, we investigate the convergence of ES and EGGROLL as the number of parameters grows, proving both updates converge to a linearised form that is consistent with the EGGROLL update using the Gaussian approximate score function. 

EGGROLL is not wedded to any particular score function approximator and we derive and explore a set of mean-field approximators in Appendix~\ref{app:mean-field} as alternatives. However, our experiments show that the Gaussian approximator has the best overall performance on the tasks we consider. To optimise the ES objective using the EGGROLL update, we adapt the parallelised evolutionary strategies algorithm from \citet{salimans2017}. We make a Monte Carlo estimate of the expectation in \cref{eq:approx_lr_grad} with $N_\textrm{workers}$ samples to optimise the mean matrix parameters $M$ using (approximate) stochastic gradient ascent. This yields the Gaussian EGGROLL update:
\begin{mybox}
\textbf{EGGROLL UPDATE:} For each worker $i$ (in parallel), sample $A_{i,t}\sim p(A_{i,t}), B_{i,t}\sim p(B_{i,t})$ and  form a low-rank perturbation $E_{i,t}=\frac{1}{\sqrt{r}}A_{i,t} B_{i,t}^\top$. Update matrix parameters using: 
 \begin{align}
     &M_{t+1}\leftarrow M_t +\frac{\alpha_t}{N_\textrm{workers}}\sum_{i=1}^{N_\textrm{workers}}
      E_{i,t} f(W=M_t+\sigma E_{i,t}).\label{eq:eggroll_update}
 \end{align}
\end{mybox}
Here we absorb the constant $\frac{1}{\sigma}$ into the tunable learning rate $\alpha_t$. As each random matrix $E_{i,t}$ in \cref{eq:eggroll_update} has rank $r$ almost surely and the matrix is updated using a sum of $N_\textrm{worker}$ such matrices, the overall EGGROLL matrix parameter update has rank $\min(Nr,m,n)$ almost surely, i.e., the overall parameter update is not restricted to be low-rank. For all experiments in \cref{sec:experiments}, $Nr> \min(m,n)$, i.e., EGGROLL parameter updates are full-rank.
\vspace{-0.2cm}
\subsection{Hardware-Efficient Implementation}\label{sec:eggroll_hardware}

A key reason to use EGGROLL over standard ES is that large populations can be simulated in parallel on a GPU thanks to the low-rank perturbations. For the sake of exposition, we write equations from the perspective of a single worker, $i$, and explain in text how this corresponds to batched GPU operations. Consider the task of computing a batched forward pass over inputs $u_i \in \mathbb{R}^{d_{in}}$ for a linear layer with mean parameter $M \in \mathbb{R}^{d_{out} \times d_{in}}$. The standard forward pass is just a regular matrix multiplication, $u_iM^T$, since $M$ is constant across all threads. In contrast, na{\"i}vely applying ES by trying to compute $u_i(M + \sigma E_i)^T$ becomes a batched matrix multiplication, which is inefficient on GPUs since every element of $M + \sigma E_i$ is only used in a single multiplication, yielding poor arithmetic intensity.

However, with EGGROLL we know that $u_i(M + \sigma E_i)^T = u_iM^T + \frac{\sigma}{\sqrt{r}}(u_iB_i)A_i^T$, which improves arithmetic intensity since it preserves the efficient general matrix multiplication used in batched inference while adding some additional cheap work per perturbation. In this context, the bulk of compute is spent on the efficient calculation of $u_iM^T$ using regular matrix multiplication. Meanwhile, when $r=1$, $u_iB_i$ simply becomes an inexpensive batch of $N$ vector-vector dot products of length $d_{in}$ to get a batch of $N$ scalars, which is then processed by a batched scalar-vector multiplication when multiplying by $A_i^T$. This decomposition is key to efficient batched LoRA inference, such as those used by vLLM~\citep{kwon2023efficient}, which is why EGGROLL achieves the same speeds as batched LoRA inference systems. The batched LoRA inference enables high arithmetic intensity, enabling us to saturate compute with many unique perturbations per input. Note that this is impossible with na{\"i}ve ES because each perturbation requires a separate matrix-vector multiplication, setting an upper bound of 1 for arithmetic intensity regardless of population size; see Appendix~\ref{sec:arithmetic_intensity_appendix} for a full derivation. We additionally optimise the update by not explicitly materialising the individual $E_i$ in the computation of $\sum_{i=1}^{N}E_if_i$, the key term in the Gaussian approximate score function. In particular, when the rank is 1, we reconstruct $A \in \mathbb{R}^{N \times d_{out}}$ and $B \in \mathbb{R}^{N \times d_{in}}$ and calculate the expression as $(\text{diag}(f)A)^TB$, a simple matrix multiplication.

    \section{Analysis}
\label{sec:analysis}
\emph{Proofs for all theorems can be found in Appendices~\ref{sec:notation} to~\ref{app:asymptotic_rank}}. 

In this section, we investigate the theoretical properties of the ES and EGGROLL updates. In \cref{sec:high_dim_es}, we study the convergence properties of the general Gaussian ES update as the parameter dimension $d\rightarrow \infty$, obtaining the conditions required for convergence to a linearised form. We then extend this analysis to the EGGROLL update in \cref{sec:high_dim_egg}. Finally, in \cref{sec:rank} we provide an analysis investigating the effect that increasing the rank of the EGGROLL approximation, proving convergence to the true ES update in the limit.

\subsection{High-Dimensional Gaussian ES}\label{sec:high_dim_es}
We first analyse the general ES update under Gaussian perturbations from \cref{eq:standard_normal}:
\begin{align}
    \nabla_\mu J(\theta)=\frac{1}{\sigma_d}\mathbb{E}_{v\sim \mathcal{N}(0,I_d)}\left[v\cdot f(\mu+\sigma_d v)\right],
\end{align}
where $v\in \mathbb{R}^d$. In high dimensions, the Gaussian annulus theorem \citep{vershynin2018hdp,wegner2024gaussrnd} proves that the probability mass of standard Gaussian distributions concentrates in thin shells of radius $\sqrt{d}$, which place probability mass further from the origin as dimension $d$ increases. To counter this, we let $\sigma_d$ depend on $d$ and analyse the \emph{critical decay rate} of $\sigma_d$ that yields convergence of the ES updates.  We make the following mild regularity assumptions:
\begin{assumption}[Locally Continuous Fitness]\label{ass:continuous}
   With probability 1 with respect to the random initialisation of $\mu$, assume there exists a ball $B_\rho(\mu)\coloneqq  \{x'\vert \lVert x'-\mu\rVert<\rho\}$ of fixed radius $\rho>0$ where $f(x)$ is $C^1$-continuous for all $x\in B_\rho(\mu)$. Within this ball, let $\nabla f(x)$ be $\alpha$-H\"{o}lder continuous, i.e., $\lVert \nabla f(x)-\nabla f(y)\rVert\le L\lVert x-y\rVert^\alpha$ for all $x,y\in B_\rho(\mu)$, $\alpha\in(0,1]$ and $L=\mathcal{O}(1)$.
\end{assumption}
   Assumption~\ref{ass:continuous} \emph{does not restrict the fitness to be globally continuous}; with probability one with respect to the initialisation distribution there must exist an arbitrarily small $C^1$-continuous ball around $\mu$. In particular, discontinuities, kinks, and non-differentiable regions may exist in the domain, provided they are not encountered with nonzero probability in the local region explored by the algorithm. $\alpha$-H\"{o}lder is the weakest simple, dimension-robust assumption that guarantees vanishing local gradient variation under Gaussian perturbations; it is weaker than Lipschitz continuity, which is recovered with $\alpha=1$. 
  
\begin{assumption}[Global Polynomial Growth] \label{ass:polynomial}
   Assume that there exists some constant $0<C<\infty$  that is $\mathcal{O}(1)$ in $d$ and finite polynomial degree $p\ge0$ such that $\lvert f(\mu+\sigma_d v)\rvert \le C(1+\lVert \mu+\sigma_d v\rVert^p)$ and $\lVert\nabla f(\mu+\sigma_d v)\rVert \le C(1+\lVert \mu+\sigma_d v\rVert^p)$ almost surely under  $v\sim\mathcal N(0,I_d)$. 
\end{assumption}
Unlike Assumption~\ref{ass:continuous}, this is a \emph{global} assumption. Again, discontinuities can exist. The assumption is weaker than boundedness, is satisfied by essentially all fitness functions used in ES, and ensures that both the objective and its gradient are integrable under Gaussian perturbations; objectives violating this condition typically exhibit super-polynomial growth and derivative growth, which leads to ill-defined or highly unstable ES updates. Moreover, if the condition is not satisfied almost surely, then the function and its gradients are undefined in regions that have nonzero Gaussian measure. 
\begin{assumption}[Bounded Derivative]\label{ass:bounded_derivative}
 With probability 1 with respect to the random initialisation of $\mu$, assume that $\lVert \mu\rVert=\mathcal{O}(1)$ and $\lVert \nabla f(\mu)\rVert=\mathcal{O}(1)$, i.e. $\lVert \mu\rVert$ and $\lVert \nabla f(\mu)\rVert$ do not grow with increasing $d$.  
\end{assumption}
This assumption is standard in high-dimensional analysis proving convergence to linearity, as proving convergence to $\nabla f(\mu)$ becomes meaningless if $\lVert \nabla f(\mu)\rVert\rightarrow \infty$. Moreover, the ES update as a whole can diverge if Assumption~\ref{ass:bounded_derivative} is not satisfied. It can be ensured by scaling, typically by scaling networks parameters by $d^{-\frac{1}{2}}$ or using an appropriate scaled initialisation, commonly Gaussian initialisation $\mu\sim \mathcal{N}\left(0,\frac{1}{d} I_d\right)$. This is precisely the scaling employed in the neural tangent kernel (NTK) regime \citep{Jacot18,Lee19,Chizat19}, where it guarantees dimension-independent gradients and stable training dynamics.

 These assumptions encompass essentially all objectives encountered in modern machine learning, including networks with finitely many ReLU activations, max- and hinge-based losses, and other piecewise-smooth or discontinuous models. Our first theorem proves convergence of a Gaussian ES update to a linearised form, that is to the local first-order derivative $\nabla f(\mu)$,  with a tight convergence rate for any function satisfying these assumptions:

\begin{theorem}[Convergence to Linearity]\label{proof:convergence_to_lin}
Let Assumptions~\ref{ass:continuous},~\ref{ass:polynomial}, and~\ref{ass:bounded_derivative} hold and $\sigma_d=o\left(d^{-\frac{1}{2}}\right)$. Then: $\left\lVert \nabla_\mu J(\theta)-\nabla f(\mu)\right\rVert=\Theta\left(\left(\sigma_d \sqrt{d}\right)^\alpha\right)=o(1)$, almost surely with respect to the distribution over $\mu$. \vspace{-0.1cm}
 \end{theorem}
To understand the effect that breaching the $\sigma_d=o\left(d^{-\frac{1}{2}}\right)$ rate has on the convergence of Gaussian ES, we study the space of functions that can be represented by cubic polynomials of the form:
\begin{align}
    f(x)=a^\top x+\frac{1}{2}x^\top B x+\frac{1}{6}C(x,x,x),\label{eq:cubic}
\end{align}
where $a\in\mathbb{R}^d$, $B\in\mathbb{R}^{d\times d}$ is a symmetric matrix and  $C(x,x,x)=\sum_{i,j,k} c_{i,j,k} x_ix_j x_k$ denotes a symmetric 3-linear map represented by the symmetric 3-tensor $C\in\mathbb{R}^{d\times d\times d}$, which generalises  cubic equations of the form $f(x)=ax+bx^2+cx^3$ to vector-valued $x$. These are non-pathological, well-behaved, analytic $C^\infty$-continuous functions, and include a rich subclass of convex optimisation problems, for instance, cubic perturbations of strictly convex quadratics. Moreover, any convex $C^3$-continuous objective admits a local third-order Taylor expansion of this form around a minimiser.
\begin{theorem}[Exact Divergence for Cubic Objectives]\label{proof:exact_cubic_divergence}
Let $f(x)$ denote the cubic polynomial in \cref{eq:cubic}. Assume $\lVert a\rVert=\mathcal{O}(1)$,$\lVert B\rVert=\mathcal{O}(1)$, $\lVert C\rVert=\mathcal{O}(1)$ where $\lVert\cdot \rVert$ denotes operator norm for $i$-tensor $T(x_1,\dots x_i)$: $\lVert T\rVert\coloneq\sup_{\lVert x_1\rVert=\dots =\lVert x_i\rVert=1}\lvert T(x_1,\dots x_i)\rvert$. Let Assumption~\ref{ass:bounded_derivative} hold, then:
\begin{align}
    \nabla_\mu J(\theta)
    =
    \nabla f(\mu)
    + \frac{\sigma_d^2}{2}
    \mathbb{E}_{v\sim\mathcal{N}(0,I_d)}
    \left[
         C(v,v,\cdot)
    \right].\label{eq:update}
    \end{align}
    Moreover:
    \begin{align}
    \left\lVert
    \frac{\sigma_d^2}{2}\mathbb{E}_{v\sim\mathcal{N}(0,I_d)}
    \left[
        C(v,v,\cdot)
    \right]
    \right\rVert
    &=
    \Theta(\sigma_d^2d),\label{eq:remainder_rate}\\
    \left\lVert
    \nabla_\mu J(\theta)-\nabla f(\mu)
    \right\rVert&=
    \Theta(\sigma_d^2 d).\label{eq:cubic_convergence}
\end{align}\vspace{-0.4cm}
\end{theorem}
Together, \cref{proof:convergence_to_lin,proof:exact_cubic_divergence} prove Gaussian ES has a \emph{critical convergence rate} of $\sigma_d=o\left(d^{-\frac{1}{2}}\right)$ in high dimensions, and operates in three regimes:

\paragraph{Regime I (Convergence to Linearity):} For $\sigma_d=o\left(d^{-\frac{1}{2}}\right)$, ES converges to a linearised form, recovering a local first-order gradient update $\nabla f(\mu)$. This result is \emph{analogous to neural tangent kernel} (NTK) type theorems, which prove that neural networks linearise in high dimensions \citep{Jacot18} and results from the concentration of the population distribution as $d\rightarrow \infty$, but applies to a more general set of objectives including discontinuous architectures. Moreover, \cref{proof:convergence_to_lin} proves that the $(\sigma_d \sqrt{d})^\alpha$ rate at which Gaussian ES converges is tight and cannot in general be improved upon without strengthening continuity or introducing specific structure into the objective to ensure the H\"{o}lder constant $L$ decays with $d$; for the class of cubic functions we consider in \cref{proof:exact_cubic_divergence}, the faster $\sigma_d^2 d$ convergence rate found in \cref{eq:cubic_convergence} is possible due to the $C^\infty$-continuity of this function class, which means the converge rate is governed by third order derivative terms. 
\paragraph{Regime II (Critical):} For $\sigma_d\asymp d^{-\frac{1}{2}}$, Gaussian ES converges to a nonlinear limiting update that may retain higher-order derivative terms when they exist; for our cubic example, \cref{eq:remainder_rate} proves that at this critical rate, the second-order term associated with the matrix $B$ vanishes due to symmetry and the third-order term associated with the tensor $C$ remains:
\begin{align}
    \left\lVert \frac{\sigma_d^2}{2}\mathbb{E}_{v\sim\mathcal{N}(0,I_d)}
    \left[
        C(v,v,\cdot)
    \right]\right\rVert=\Theta(1).
\end{align}
As the polynomial form is representative of general Taylor expansions, this implies that the limiting high dimensional update retains third-order derivatives (and higher order odd derivatives) as $d\rightarrow \infty$.

\paragraph{Regime III (Divergence):} For $d^{-\frac{1}{2}}=o\left(\sigma_d\right)$, \cref{proof:exact_cubic_divergence} shows that there exist smooth cubic objectives with bounded coefficients for which: 
\begin{align}
    \left\lVert
    \nabla_\mu J(\theta)
    \right\rVert&=
    \Theta(\sigma_d^2 d)\rightarrow \infty.
\end{align}
In particular, divergence occurs whenever the cubic tensor has a non-vanishing Gaussian contraction (equivalently, non-zero partial trace), i.e. in non-degenerate cases; only in the exceptional trace-free case does the cubic contribution vanish.

In practice, $\sigma_d$ is often absorbed into the ES update stepsize, and its scale is adjusted automatically as part of the hyperparameter regime to ensure stability. 

\subsection{High-Dimensional EGGROLL}\label{sec:high_dim_egg}
We now extend our high-dimensional analysis to study the EGGROLL update using the Gaussian approximate score function $\hat{g}_\textrm{LR}$ from \cref{eq:gaussian_approx}. Taking $r$ as fixed, we consider the Gaussian matrix ES setting outlined in \cref{sec:matrix_es}. We take $x=\textrm{Vec}(W)$ where $W\in\mathbb{R}^{m\times n}$ and analyse the effect of increasing the total number of matrix parameters $d=mn$. Recall the true ES Gaussian matrix update is:
\begin{align}
    \nabla_M J(\theta)=\frac{1}{\sigma}\mathbb{E}_{E\sim P(E)}\left[E \cdot f(W=M+\sigma E)\right],
\end{align}
where $M$ is the set of mean matrix parameters associated with the matrix $W$ and 
 $P(E)$ is a zero-mean standard normal $p(E)=\mathcal{N}(0,I_m,I_n)$.

Two key differences between full-rank Gaussian ES and EGGROLL are that $\hat{g}_\textrm{LR}$ is an approximation to a true gradient and $P(E)$ may have heavier tails than a Gaussian. To account for these differences, we require a slightly stricter local continuity control assumption:
\begin{assumption}[EGGROLL Locally Continuous Fitness]\label{ass:egg_continuous}
   With probability 1 with respect to the random initialisation of $\mu$, assume there exists a ball $B_\rho(\mu)\coloneqq  \{x'\vert \lVert x'-\mu\rVert<\rho\}$ of fixed radius $\rho>0$ where $f(x)$ is $C^2$-continuous for all $x\in B_\rho(\mu)$ and $\lVert \nabla^2 f(\mu)\rVert$ be polynomial bounded in $d$. Within this ball, let $\nabla^2 f(x)$ be Lipschitz continuous, i.e. $\lVert \nabla^2 f(x)-\nabla^2 f(y)\rVert\le L_d\lVert x-y\rVert$ for all $x,y\in B_\rho(\mu)$.
\end{assumption}
 This assumption still permits discontinuous objectives. We also assume that $p_0(\cdot)$ generates sub-Gaussian elements with uniform tail control:
\begin{assumption}[Sub-Gaussian Tails]
\label{ass:convergence_sampling}
    In addition to Assumption~\ref{ass:iid}, assume that $p_0(\cdot)$ generates variables that have sub-Gaussian tails, i.e. for $x_i\sim p_0(x_i)$:
    \begin{align}
    \mathbb{P}(\lvert x_i \rvert >t)\le 2\exp(-C t^2),
    \end{align}
    for some $0\le C<\infty$ that does not depend on $d$.
\end{assumption}
We discuss sub-Gaussian variables and their properties in \cref{app:egg_lin}
 The assumption is trivially satisfied for Gaussian distributions $a\sim\mathcal N(0,I_m)$ and $b\sim\mathcal N(0,I_n)$, and holds more generally, for example for bounded distributions, uniform distributions and generalised Gaussian distributions with shape parameter greater than two. This flexibility is particularly relevant for the models in \cref{sec:egg}, where heavier-shouldered distributions may be preferred over the Gaussian.

 \begin{theorem}[EGGROLL Convergence to Linearity]\label{proof:egg_convergence_to_lin} Let $W\in\mathbb{R}^{m\times n}$, $d=mn$ and $x=\textrm{Vec}(W)$.
Let Assumptions~\ref{ass:polynomial},~\ref{ass:bounded_derivative},~\ref{ass:egg_continuous} and~\ref{ass:convergence_sampling}  hold, $\sigma_d=o(d^{-1/2})$, and $L_d(\sigma_d d)^2=o(1)$. Then there exists some $K>0$ such that:
\begin{align}
    \lVert  \hat{g}_\textrm{LR}-\nabla_W f(W=M)\rVert_F=\mathcal{O}\left(L_d(\sigma_d d)^2\right)+\mathcal{O}\left(\frac{\sqrt{d}}{\sigma_d^2}  \exp\left(-K\frac{\rho}{\sqrt{d}\sigma_d}\right) \right)=o(1),\label{eq:matrix_linearisation_rate}
\end{align}
and
\begin{align}
    \lVert \hat{g}_{LR}-\nabla_M J(\theta) \rVert_F=\mathcal{O}\left(\sigma_d \sqrt d\cdot\left(1+L_d\sigma_dd^\frac{3}{2}\right)\right)=o(1).\label{eq:convergence_to_matrix_es}
\end{align}
 almost surely with respect to the distribution over $\mu$. \vspace{-0.2cm}
 \end{theorem}
Our theory explains the success of EGGROLL in high dimensions with rank as small as $r=1$; \cref{eq:convergence_to_matrix_es} proves EGGROLL converges to the true update matrix ES update $\nabla_M J(\theta)$ as $d\rightarrow \infty $ regardless of $r$. In addition, \cref{eq:matrix_linearisation_rate} proves that under the same conditions, the EGGROLL update also linearises like the true Gaussian ES update analysed in \cref{sec:high_dim_es}, recovering a local first-order derivative as $d\rightarrow \infty$. For high-dimensional neural networks, standard parametrisations place training in the NTK regime, in which the network behaves approximately linearly in its parameters and gradient descent converges to a global minimum \citep{Jacot18,Lee19,Chizat19}. Recent results show that the spectral norm of the Hessian decays polynomially with width, and that higher-order derivatives governing the variation of the Hessian also vanish \citep{Liu2020}. Consequently, the Lipschitz constant $L_d=o(1)$, typically at rate $d^{-\frac{1}{2}}$ or $d^{-1}$ depending on the network architecture. Substituting these rates into our upper bound in \cref{eq:matrix_linearisation_rate} yields convergence rates of $\mathcal{O}(\sigma_d^2 d^{\frac{3}{2}})$ or $\mathcal{O}(\sigma_d^2 d)$ respectively.

\subsection{Rank Analysis}\label{sec:rank}
We now analyse how fast the low-rank update from \cref{eq:approx_lr_grad} with Gaussian score approximation  converges to the true Gaussian ES matrix gradient in \cref{eq:ES_update} as the rank of the update $r$ increases. We make notation explicit in $r$ in this subsection, for example writing $E^r=\frac{1}{\sqrt{r}}A^r{B^r}^\top $. We introduce the following formal regularity assumption for the fitness function:
\begin{assumption}[Bounded Fitness]\label{ass:boundedness}
     Assume that $f(W)$ is bounded, that is $\sup_{W}\left\lvert f(W)\right\rvert <\infty $.
\end{assumption}
 Our key theoretical result characterises the error rate between the Gaussian score approximator in the low-rank update $\hat{g}_\textrm{LR}^r$ from \cref{eq:approx_lr_grad} and the true gradient using the matrix Frobenius norm:
 \begin{theorem}[EGGROLL Rank Convergence]\label{proof:approximator_error}
Let Assumptions~\ref{ass:iid} and~\ref{ass:boundedness} hold, then:
\begin{align}
    \lVert \hat{g}_\textnormal{\textrm{LR}}^r-\nabla_\mu J(\theta)\rVert_F=\mathcal{O}\left(r^{-1}\right).\label{eq:lr_convergence_rate}
\end{align}
\end{theorem}
\begin{wrapfigure}{R}{0.5\textwidth}\vspace{-0.5cm}
    \centering
    \includegraphics[width=0.5\textwidth]{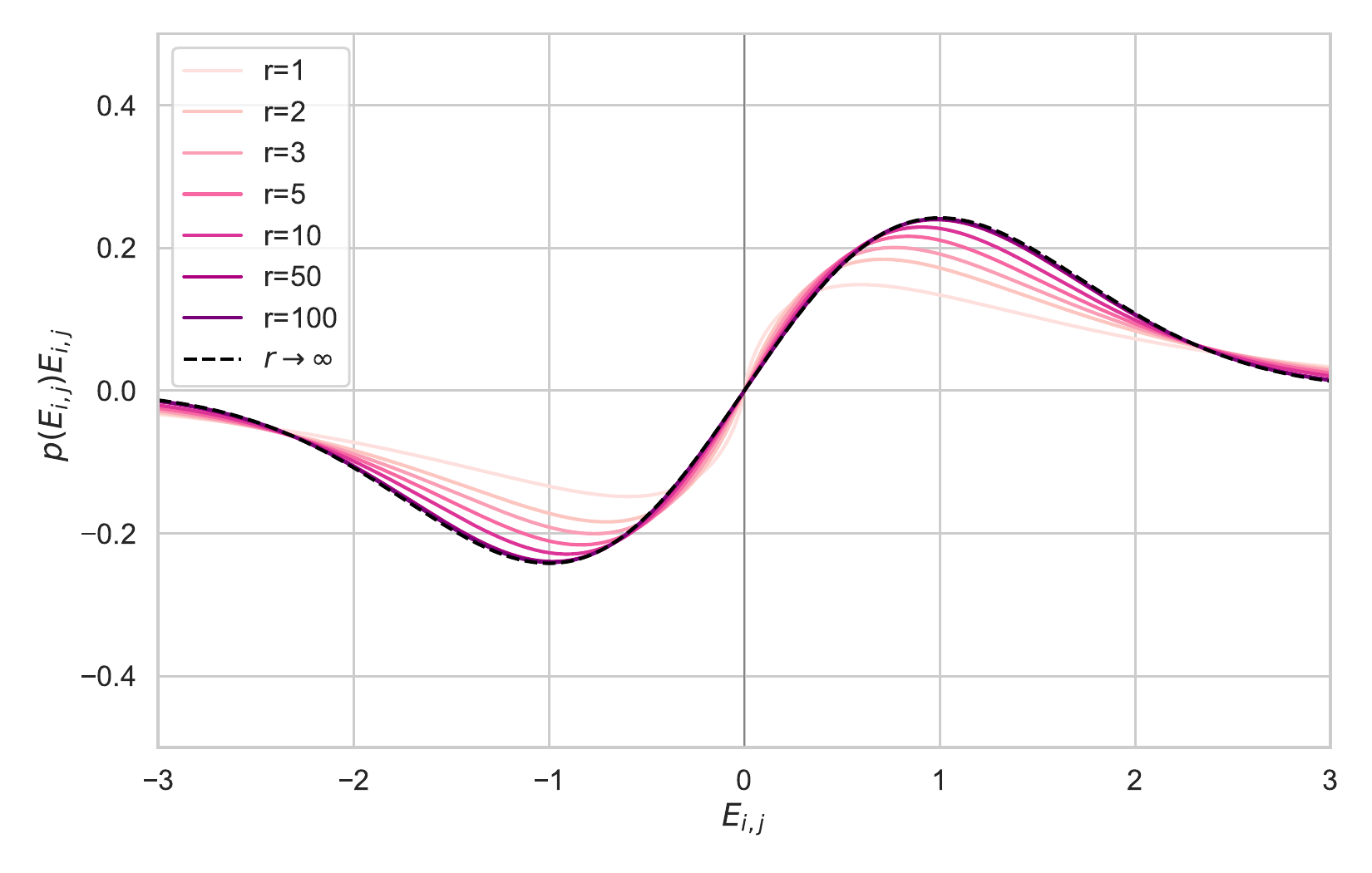}
    \vspace{-0.6cm}
    \caption{Plot of Marginal Score Multiplied by Density for Increasing $r$}
    \label{fig:score_convergence}
    \vspace{-0.3cm}
\end{wrapfigure}
The convergence rate in \cref{eq:lr_convergence_rate} is faster than the typical $\mathcal{O}\left(r^{-\frac{1}{2}}\right)$ rate dictated by the general parametric central limit theorem. Our analysis shows that this is due to the symmetry in our problem under \cref{ass:iid}. To obtain our results, we make an Edgeworth expansion \citep{Bhattacharya76} of the distribution $P(E^r)$, which expands $P(E^r)$ as the limiting Gaussian distribution plus a sum of decaying terms that are controlled by the 3rd order and higher cumulants of $P(E^r)$. Each $i$th order cumulant term is multiplied by a factor that decays at rate $\mathcal{O}\left(r^{-\frac{i-2}{2}}\right)$. For symmetric zero-mean distributions, all odd cumulants are zero (for the same reason that all odd moments of a symmetric distribution are zero). Hence, the rate of convergence to the limiting distribution is controlled by the 4th order term, which has rate $\mathcal{O}\left(r^{-1}\right)$. 

Although the full distribution $P(E^r)$ has no general closed-form solution, the distribution over marginals $P(E_{i,j})$ is more amenable to analysis. We derive the density of the marginal distribution $P(E_{i,j})$ for generalised Gaussian distributed $a_{i,j}$ and $b_{i,j}$ in \cref{app:mean-field}. To illustrate the fast convergence rate, we plot the negative density $\times$ score function $p(E_{i,j})E_{i,j}$ for the marginal density $p(E_{i,j})$ in \cref{fig:score_convergence} using Gaussian distributed $a_{i,j}$ and $b_{i,j}$ (see \cref{proof:marginal_score_gauss} for a derivation). The figure shows that $p(E_{i,j})E_{i,j}$ quickly converges to the limiting function $\frac{E_{i,j}}{\sqrt{2\pi}}\exp\left(-\frac{{E_{i,j}}^2}{2}\right)$, recovering the Gaussian form from the true Gaussian ES update. Even at $r=1$, the function is not a poor approximation. After $r=10$, the function has nearly converged and after $r=50$, the function is visually indistinguishable from the limit, providing evidence for the hypothesis that the low-rank approximation is accurate even for very low-rank regimes $r\ll\min(m,n)$.

    \section{Experiments}
In the following section we showcase the effectiveness of EGGROLL in a variety of tasks that position it as a strong alternative to back-propagation for the end-to-end training of foundation models.
\label{sec:experiments}
\subsection{Pure Integer Language Model Pretraining}
\label{sec:egg}

To demonstrate the potential of EGGROLL as a general optimisation method, we apply it to language model pretraining. Since EGGROLL does not rely on gradients, we explicitly design a language model architecture to be efficient and hardware-friendly at inference time. To highlight EGGROLL's flexibility, we train a nonlinear recurrent neural network (RNN) in pure integer datatypes with no explicit activation functions, relying only on the implicit nonlinearity of clipping in int8 operations. We call the resulting language model EGG, the Evolved Generative GRU, an EGGROLL-friendly architecture with all weights in int8. See Appendix~\ref{sec:egg_appendix} for more details on the architecture and motivation behind EGG.

We train an EGG model with 6 layers and hidden dimension 256 (6L-256D) to do character-level prediction on the minipile dataset~\citep{kaddour2023minipile}. We update parameters after 100 tokens for each population member, applying truncated ES by keeping the hidden state and only resetting at document boundaries. We plot the test loss in~\cref{fig:pretrain_loss} over training steps across a range of population sizes with a fixed data batch size of 16 sequences per step, where the best test loss is 3.40 bits/byte. With a sufficiently large population size, EGG outperforms a dense 6L-256D Transformer trained with backprop SGD using the same data batch size. Note that larger population sizes require more parallel compute for the same amount of data; our largest population size of $2^{20}=1048576$ requires around 180 times more GPU-hours than the backprop baseline, demonstrating the potential for compute-only scaling in limited data regimes using EGGROLL.

Moreover, our largest population size of $2^{20}$ is three orders of magnitude larger than the largest experiment done by~\citet{salimans2017} while only requiring a single GPU to train, highlighting EGGROLL's computational efficiency. We note that large population sizes are critical for pretraining; a population size of 2, analogous to MeZO~\citep{MeZO}, significantly underperforms larger population sizes despite having access to the same data batch. We conduct more ablations in Appendix~\ref{sec:egg_ablation}, analysing the tradeoff between population size and data batch size.

\subsection{Reinforcement Learning Tasks}\label{ssec:rl_exp}
\begin{figure*}[t]
  \centering
  \begin{subfigure}[t]{0.51\textwidth}
    \centering
    \includegraphics[width=\linewidth]{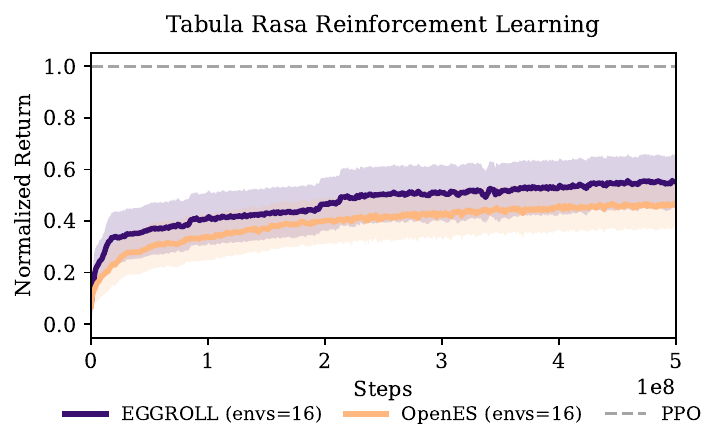}
    \phantomsubcaption
    \label{fig:rl_combined}
  \end{subfigure}
  \hfill
  \begin{subfigure}[t]{0.47\textwidth}
    \centering
    \includegraphics[width=\linewidth]{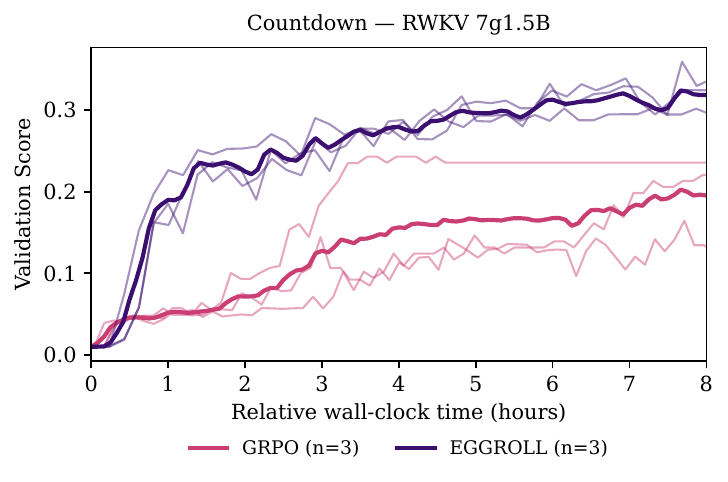}
    \phantomsubcaption
    \label{fig:eggroll_countdown_1}
  \end{subfigure}
  \hfill
  \vspace{-0.5cm}
  \caption{(a) Comparison of reinforcement learning returns normalised by PPO performance across 16 environments for 10 seeds. The shaded region is the standard error of the mean.(b) Validation score of 3 seeds of EGGROLL v.s. 3 seeds of GRPO in countdown task with an RWKV 7g1.5B model on a single GPU. EGGROLL allows 1024 parallel generations per GPU (618 updates) whereas GRPO only 64 (915 updates).}
  \vspace{-0.4cm}
\end{figure*}

To verify that low-rank perturbations do not change the optimisation behavior of ES in standard control settings, we benchmark EGGROLL against OpenES \citep{salimans2017} across 16 tabula rasa environments spanning Navix, Craftax, Brax, Kinetix, and Jumanji. We use a fixed 3-layer MLP policy (256 hidden units) and perform per-environment hyperparameter optimisation for each method before evaluating the selected configuration over 10 random seeds, reporting mean performance (normalised by PPO) and uncertainty. Overall, EGGROLL is competitive with OpenES on 7/16 environments, underperforms on 2/16, and outperforms on 7/16, while often delivering substantial wall-clock improvements due to its batched low-rank structure (full environment list, learning curves, timing comparisons, and complete HPO ranges/settings are provided in Appendix \ref{app:rl}). Figure \ref{fig:rl_combined} shows the averaged normalised return across the 16 environments with 10 seeds per environment. We additionally report MARL results in~\cref{app:marl}.

\subsection{Foundation Model Fine-tuning}
\begin{figure*}[t]
  \begin{subfigure}[t]{0.49\textwidth}
    \centering
    \includegraphics[width=\linewidth]{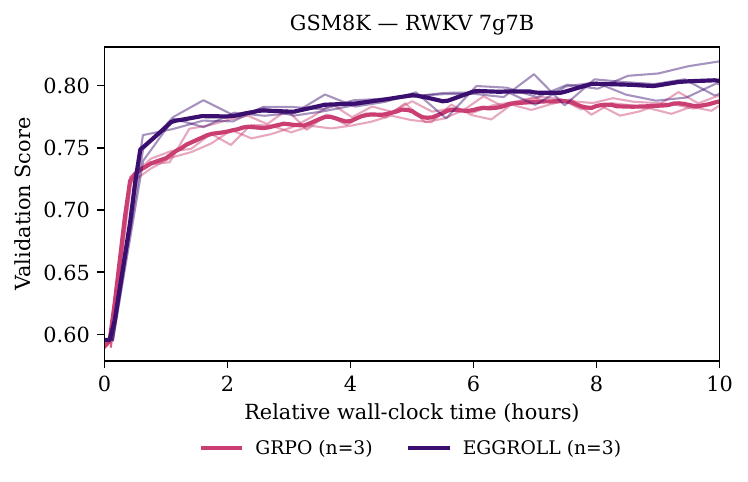}
    \phantomsubcaption
    \label{fig:eggroll_gsm8k}
  \end{subfigure}
  \hfill
  \begin{subfigure}[t]{0.49\textwidth}
    \centering
    \includegraphics[width=\linewidth]{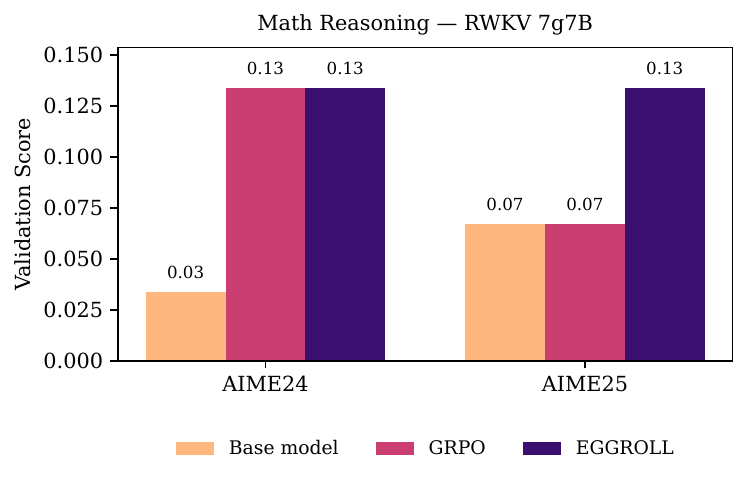}
    \phantomsubcaption
    \label{fig:eggroll_hard_reasoning}
  \end{subfigure}
      \vspace{-0.5cm}
  \caption{(a) Comparison of the validation score of 3 seeds of EGGROLL v.s. 3 seeds of GRPO in GSM8K task with an RWKV 7g7B model on 8 GPUs. EGGROLL allows 8192 parallel generations (1024 per GPU with 260 updates) whereas  GRPO only 256 (32 per GPU with 340 updates). (b) Performance of our finetuned RWKV 7G 7 billion model on hard reasoning tasks using 128 GPUs for 12 hours. The model was trained using the DeepScaleR dataset and the best checkpoint was chosen by evaluating on AIME24.}
  \vspace{-0.4cm}
\end{figure*}
We apply EGGROLL to finetune an RWKV-7 \citep{peng2025rwkv7gooseexpressivedynamic} LLM on two reasoning tasks: countdown~\citep{gandhi2024streamsearchsoslearning} and GSM8K~\citep{cobbe2021gsm8k}. RWKV is a recurrent model that is better suited to parallelisation than transformers because any memory otherwise spent on the KV cache is used to evaluate population members. Figure \ref{fig:eggroll_countdown_1} shows that EGGROLL fine-tuning on an RWKV-7 1.5B model converges to a higher validation accuracy of 35\% (vs.\ 23\%) given the same hardware and wall-clock time in the countdown task. Similarly, Figure \ref{fig:eggroll_gsm8k} shows that EGGROLL outperforms GRPO on GSM8K fine-tuning. Our scoring function draws parallels to the group relative advantage of GRPO. In particular, to score a set of noise directions, $E \equiv \{E_1, \hdots, E_n\}$, we first compute their accuracies, $\{s_{1, q_i}, \hdots, s_{n, q_i}\}$, on $|q|=m$ questions, creating a matrix of scores $S \in \mathbb{R}^{m\times n}$. We then compute the normalised score per question, with the main difference that we use the global variance $\bar{\sigma}$, and average over all the questions to compute a score for the noise direction $E_{i}$:
\begin{equation}
    \bar{s}_{i} = \frac{1}{m}\sum_{j=1}^m z_{i, q_j} =\frac{1}{m}\sum_{j=1}^m \frac{s_{i, j} - \mu_{q_j}}{\bar{\sigma}}.
\end{equation}
This scoring function weights all questions within the same batch the same across population members. 
We use this recipe to train a 14 billion parameter RWKV 7 model on the DeepScaleR dataset and evaluate in more challenging maths reasoning tasks. In this regime, GRPO is infeasible due to the extra memory used by the Adam optimiser \cite{kingma2014adam}. Using a thinking budget of 5000 tokens for training and evaluation, our fine-tuned 14B model improves from 13\% to 30\% accuracy on AIME24, from 7\% to 33\% accuracy on AIME25 and from 11\% to 13\% accuracy on HMMT25 after training on 32 GPUs for 12 hours (Figure \ref{fig:eggroll_hard_reasoning_14B}). On 7B models, we outperform GRPO using 128 GPUs for 24 hours (Figure \ref{fig:eggroll_hard_reasoning}). 

In \cref{sec:vllm-exps}, we achieve similar performance to GRPO when fine-tuning Qwen Transformer models, and additionally demonstrate that EGGROLL can directly optimise for pass@k, a known limitation of GRPO~\citep{yue2025limit-of-rlvr}. Beyond language models, we also fine-tune a finance world model into an agent for high-frequency trading that directly optimises for PnL; see ~\cref{sec:eggroll_trading} for more details.

\subsection{Fine-tuning Integer Quantised LLMs}

\begin{figure*}[t]
  \begin{subfigure}[t]{0.49\textwidth}
    \centering
    \includegraphics[width=\linewidth]{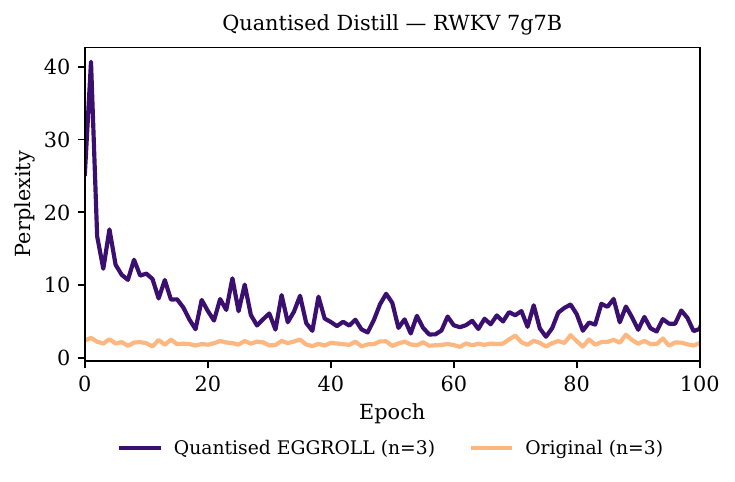}
    \phantomsubcaption
    \label{fig:eggroll_quantised_distillation}
  \end{subfigure}
  \hfill
  \begin{subfigure}[t]{0.49\textwidth}
    \centering
    \includegraphics[width=\linewidth]{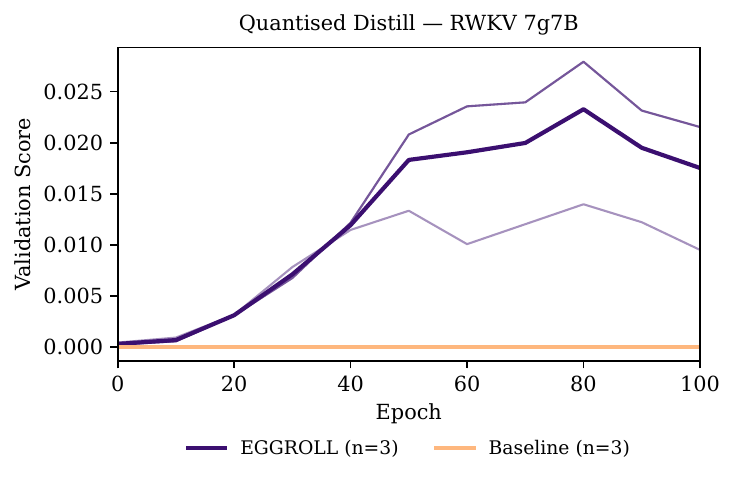}
    \phantomsubcaption
    \label{fig:distillation_validation}
  \end{subfigure}
      \vspace{-0.5cm}
  \caption{(a)  Average per token perplexity (during training) of 3 seeds of a quantised (int8) RWKV 7G 7 billion parameter model on distillation from the non quantised model using examples from GSM8K. (b) Validation score on unseen examples of GSM8K of 3 seeds of a quantised RWKV 7G 7 billion parameter model. Initially the model is unable to solve any problems, but progressively it is capable of solving more problems. The baseline here indicates the validation score of a quantised model without any further training.}
  \vspace{-0.4cm}
\end{figure*}

We follow the same procedure as \citet{jacob2017quantizationtrainingneuralnetworks} to quantise the RWKV-7 family of models by dividing by the maximum \emph{per-channel} value on each weight matrix and mapping into the int8 range of $[-127, 127]$. We then apply EGGROLL with Adam to do model distillation from the original, non-quantised RWKV-7, into the resulting int8 quantised model using examples from GSM8K. See Appendix \ref{sec:finetuning_appendix} for full details about the specifics of quantisation and fine-tuning. The distillation is done by matching the distributions between the quantised and non-quantised models on teacher forced examples (with solutions) from the GSM8K dataset. 
More specifically, the fitness for a given set of parameters, $\mu_i$, is computed as follows:
\begin{equation}
    f_{\mu_i}(x_{1:T}) = \sum_{t=1}^T \text{KL}\left(p_t||q_t(\cdot; \mu_i)\right),
\end{equation}
where $x_{1:T}$ is a subsequence of tokens taken from the solutions of GSM8K and $\text{KL}\left(p_t||q_t(\cdot; \mu_i)\right)$ is the Kullback-Leibler divergence between the distribution of the non-quantised model, $p_t$, and the distribution of the quantised model $q_t$ over the vocabulary at token $t$. 
Figure \ref{fig:eggroll_quantised_distillation} shows the average per token perplexity of $3$ seeds of a quantised RWKV 7G 7 billion parameter model compared to that of the original non-quantised model over the same sequence, as a baseline. Progressively, the quantised model recovers the capability to solve a subset of the GSM8K dataset (Figure \ref{fig:distillation_validation}).

    \section{Conclusion}
We introduce EGGROLL, a powerful method for black-box optimisation that scales evolutionary strategies to billion-parameter models and beyond using low-rank search matrices. Our experiments demonstrate that EGGROLL is effective with a rank of 1, giving substantial computational and memory savings for negligible decrease in performance when compared to the full-rank perturbations. Empirically, EGGROLL delivers large speedups over naïve ES in tabula rasa and multi-agent RL, and can power end-to-end training pipelines for foundation models. Our theoretical analysis shows that the EGGROLL update converges towards the Gaussian ES update with increasing rank $r$ and parameter dimension $d=mn$, and we provide a rigorous study of general ES at high dimensions, deriving necessary and sufficient conditions for convergence and linearisation.

Looking forward, we can use EGGROLL for other problems beyond the reach of modern first-order gradient-based techniques. In particular, EGGROLL can enable the training of large scale end-to-end neurosymbolic systems~\citep{sarker2021neurosymbolicartificialintelligencecurrent} with non-differentiable components. For instance, we can train neural networks that interface with symbolic modules for specific functions, like memory or calculations. We can also optimise end-to-end systems of language models, training them to be aware of inference-time harnesses and interactions with other agents in complex systems.

    \subsection*{Acknowledgements}

Compute for this project is graciously provided by the Isambard-AI National AI Research Resource, under the projects ``FLAIR 2025 Moonshot Projects'' and ``Robustness via Self-Play RL.'' Some experiments also used compute generously given by JASMIN, the UK’s collaborative data analysis environment (\url{https://www.jasmin.ac.uk}).

Bidipta Sarkar is supported by the Clarendon Fund Scholarship in partnership with a Department of Engineering Science Studentship for his Oxford DPhil.
Mattie Fellows is funded by a generous grant from the UKRI Engineering and Physical Sciences Research Council EP/Y028481/1.
Juan Agustin Duque is supported by the St-Pierre-Larochelle Scholarship at the University of Montreal and by Aaron Courville's CIFAR AI Chair in Representations that Generalize Systematically.
Jarek Liesen and Theo Wolf are supported by the EPSRC Centre for Doctoral Training in Autonomous Intelligent Machines \& Systems EP/Y035070/1. Jarek Liesen is also supported by Sony Interactive Entertainment Europe Ltd.
Uljad Berdica is supported by the EPSRC Centre for Doctoral Training in Autonomous Intelligent Machines \& Systems EP/S024050/1 and the Rhodes Scholarship.
Lukas Seier is supported by the Intelligent Earth CDT with funding from the UKRI grant number EP/Y030907/1.
Alexander D. Goldie is funded by the EPSRC Centre for Doctoral Training in Autonomous Intelligent Machines and Systems EP/S024050/1.
Jakob Nicolaus Foerster is partially funded by the UKRI grant EP/Y028481/1 (originally selected for funding by the ERC). Jakob Nicolaus Foerster is also supported by the JPMC Research Award and the Amazon Research Award.

We thank Andreas Kirsch for discovering an emergent log-linear scaling law for EGG loss with respect to int8 OPs in \href{https://x.com/blackhc/status/1992772994486440106}{this tweet} along with other community members for their comments and recommendations during the first arXiv release of this work.

	\bibliography{Hyperscale_ES}
	\bibliographystyle{rlc}
	
	\newpage
    \onecolumn
\newpage
\appendixwithtoc

\setcounter{theorem}{0}
\setcounter{proposition}{0}
\newpage
\section{Notation}\label{sec:notation}\label{sec:notaion}
In our proofs, we use the integral notation $\int$ to denote the integral over the corresponding $\mathbb{R}^d$ space, for example, for a matrix $E\in\mathbb{R}^{m\times n}$, $\int f(E) dE=\int_{\mathbb{R}^{m\times n}}f(E) dE$ and for a vector  $E\in\mathbb{R}^{mn}$, $\int f(v) dv=\int_{\mathbb{R}^{mn}}f(v) dv$. For $f:\mathbb{R}^d\rightarrow\mathbb{R}$, we use $\nabla f(x)$ to denote the derivative of $f(\cdot)$ evaluated at $x$. 
For a vector $v\in\mathbb{R}^{mn}$, we define the mat operator as:
\begin{align}
    \textrm{mat}(v)=\begin{bmatrix}
v_1 & v_{m+1} & \dots &v_{(n-1)m+1}\\
v_2 & v_{m+2} &\dots &v_{(n-1)m+2} \\
\vdots & \vdots & \ddots & \vdots\\
v_m & v_{2m} & \cdots& v_{mn}
\end{bmatrix},
\end{align}
so $\textrm{mat}(\textrm{vec}(M))=M$. We will use the fact that the Frobenius norm becomes the $\ell_2$ norm in vector space:
\begin{align}
    \lVert M \rVert_F=\sqrt{\sum_{i,j} {m_{i,j}}^2}=\sqrt{\sum_{k}{\textrm{vec}(M)_{k}}^2 }=\lVert\textrm{vec}(M) \rVert.\label{eq:frobenius_l2}
\end{align}
Our proofs make use of Fourier analysis. For a vector-valued function $f(v):\mathbb{R}^d\rightarrow \mathbb{R}$, we define the Fourier transform as:
\begin{align}
    \tilde{f}(\omega)=\mathcal{F}[f](\omega)\coloneqq \int f(v) \exp(-i\omega^\top v)dv,
\end{align}
and the inverse Fourier transform as:
\begin{align}
    f(v)=\mathcal{F}^{-1}[\tilde{f}](v)\coloneqq \frac{1}{(2\pi)^d} \int \tilde{f}(\omega) \exp(i\omega^\top v)d\omega,
\end{align}

\section{ES Matrix Gradient Deviations}\label{app:es_matrix}

Let $\mu_M=\textrm{vec}(M)\in\mathbb{R}^{mn}$ be the vector of mean parameters associated with the matrix $M$. Let $v_M\in \mathbb{R}^{mn}$ denote the corresponding search vector associated with $\mu_M$. As each element of $v$ is generated independently from a standard normal $\mathcal{N}(0,1)$, the search vector $v_M$ is generated from the standard multivariate norm: $v_M\sim \mathcal{N}(0,I_{mn})$. From \cref{eq:standard_normal}, the update for $\mu_M$ is: 
\begin{align}
    \sigma\nabla_{\mu_M}J(\theta)&=\mathbb{E}_{v_M\sim \mathcal{N}(0,I_{mn})}\left[v_M \cdot f(W=\textrm{mat}(\mu_M)+\sigma \textrm{mat}(v_M))\right],\\
    &=\mathbb{E}_{v_M\sim \mathcal{N}(0,I_{mn})}\left[\textrm{vec}(\textrm{mat}(v_M)) \cdot f(W=\textrm{mat}(\mu_M)+\sigma \textrm{mat}(v_M))\right],\\
    &=\mathbb{E}_{E\sim \mathcal{N}(0,I_{m},I_n)}\left[\textrm{vec}(E) \cdot f(W=M+\sigma E)\right],
\end{align}
where $E=\textrm{mat}(v_M)$ and we have used the fact that sampling $v_M\sim\mathcal{N}(0,I_{mn})$ is equivalent to sampling $E\sim \mathcal{N}(0,I_m,I_n)$ and applying $v_M=\textrm{vec}(E)$. Now 
\begin{align}
    \nabla_M J(\theta)&=\textrm{mat}(\nabla_{\mu_M} J(\theta)),\\
    &=\frac{1}{\sigma}\mathbb{E}_{E\sim \mathcal{N}(0,I_{m},I_n)}\left[\textrm{mat}(\textrm{vec}(E)) \cdot f(W=M+\sigma E)\right],\\
    &=\frac{1}{\sigma}\mathbb{E}_{E\sim \mathcal{N}(0,I_{m},I_n)}\left[E \cdot f(W=M+\sigma E)\right],\\
    &=-\frac{1}{\sigma}\mathbb{E}_{E\sim \mathcal{N}(0,I_{m},I_n)}\left[\nabla_E \log p(E) \cdot f(W=M+\sigma E)\right].
\end{align}
\newpage

\section{High-Dimensional Analysis}\label{app:high_dim}

\subsection{High-Dimensional Gaussian ES and Convergence}\label{app:es_lin}

We use insights from the Gaussian annulus theorem when investigating the convergence properties of high-dimensional ES: our proof relies on the fact that all probability mass converges to the interior of the ball $B_\epsilon(\mu)\coloneqq \{x'\vert \lVert x'-\mu\rVert < \epsilon\} $ where $\epsilon=\frac{\rho}{2}$ in the limit $d\rightarrow \infty$, where $\rho$ is the radius of the local ball from \cref{ass:continuous}, meaning we only need to consider the smooth region around $\mu $ in this limit. Our first result proves that the mass outside of the ball for any polynomially bounded function tends to zero at an exponential rate.

\begin{lemma}[Polynomial Tail Bounds]\label{proof:tail_bound}
Let $g(x)$ be polynomial bounded as:
\begin{align}
    \lVert g(\mu+\sigma_d v)\rVert\le C\lVert v \rVert^q(1+\lVert \mu+\sigma_d v\rVert^p),
\end{align}
for some finite polynomial of orders $p$ and $q$ and constant $C>0$. Let $A_d\coloneqq \{\Vert \sigma_d v\rVert \ge \epsilon\}$ denote the event that a mutation lies outside the a local ball of radius $\epsilon$ around $\mu$.
Assume $\sigma_d=o(d^{-1/2})$. Then for some constant $K>0$:
\begin{align}
   \left\lVert \mathbb{E}_{v\sim \mathcal{N}(0,I_d)}
   \left[g(\mu+\sigma_d v)\mathds{1}(A_d) \right]\right\rVert
   =
   \mathcal{O}\left(d^\frac{q}{2}
  \exp\left(-K\left(\frac{\epsilon}{\sigma_d}\right)^2\right)
   \right),
\end{align}
and in particular the right-hand side is $o(1)$ as $d\to\infty$.

\begin{proof}
We start by bounding the integrand using the polynomial bound. Denote
$\mathbb{P}(A_d)\coloneqq
\mathbb{E}_{v\sim \mathcal{N}(0,I_d)}[\mathds{1}(A_d)]$.
Then, by Jensen's inequality in the first line, polynomial boundedness in the second and $\lVert a+b\rVert^p \leq 2^{p-1}(\lVert a \rVert^p + \lVert b \rVert^p)$ in the third:
\begin{align}
    \left\lVert \mathbb{E}_{v\sim \mathcal{N}(0,I_d)}
    \left[g(\mu+\sigma_d v)\mathds{1}(A_d) \right]\right\rVert
    &\le
    \mathbb{E}_{v\sim \mathcal{N}(0,I_d)}
    \left[\lVert g(\mu+\sigma_d v)\rVert \mathds{1}(A_d)\right],\\
    &\leq
      C \, \mathbb{E}_{v\sim \mathcal{N}(0,I_d)}\left[\lVert v\rVert^q (1 + \lVert \mu + \sigma_d v\rVert^p )\mathds{1}(A_d)\right], \\
    &\leq C \, \mathbb{E}_{v\sim \mathcal{N}(0,I_d)}
    \left[ \lVert v\rVert^q(1 + 2^{p-1}\lVert \mu \rVert^p) \mathds{1}(A_d) + 2^{p-1}\sigma_d^p  \lVert v\rVert^{p+q} \mathds{1}(A_d) \right], \\
    &= C' \mathbb{E}_{v\sim \mathcal{N}(0,I_d)} \left[ \lVert v\rVert^{q} \mathds{1}(A_d) \right] + C'' \sigma_d^p \mathbb{E}_{v\sim \mathcal{N}(0,I_d)} \left[ \lVert v\rVert^{p+q} \mathds{1}(A_d) \right] .
\end{align}
where $C' = C(1+2^{p-1}\lVert \mu \rVert^p)$ and $C'' = C2^{p-1}$ are constants independent of $d$. Applying the Cauchy--Schwarz inequality to the second expectation gives:
\begin{align}
    \mathbb{E}_{v\sim \mathcal{N}(0,I_d)}
    [\lVert v\rVert^{p+q} \mathds{1}(A_d)]
    &\le
    \sqrt{
    \mathbb{E}_{v\sim \mathcal{N}(0,I_d)}[\lVert v\rVert^{2(p+q)}]
    }
    \cdot
    \sqrt{\mathbb{P}(A_d)}.
\end{align}
Now, the variable $\left\lVert v\right\rVert$ is $\chi_d$-distributed. Using the formula for the $i$-th central moment of $\left\lVert v\right\rVert$ about the origin \citep[Chapter 11.3]{forbes2011statistical} yields:
       \begin{align}
           \mathbb{E}_{v\sim\mathcal{N}(0,I_d)}\left[ \left\lVert v\right\rVert^{i}\right]=2^\frac{i}{2} \frac{\Gamma\left(\frac{1}{2}(d+i)\right)}{\Gamma\left(\frac{1}{2}d\right)}.\label{eq:expected_chi}
       \end{align}
      Applying the identity $\frac{\Gamma(z+a)}{\Gamma(z+b)}\sim z^{a-b}$ \citep[Eq. 5.11.12]{dlmf_gamma}:
       \begin{align}
           \mathbb{E}_{v\sim\mathcal{N}(0,I_d)}\left[ \left\lVert v\right\rVert^{i}\right]&\sim 2^\frac{i}{2} \left(\frac{d}{2}\right)^\frac{i}{2}=d^\frac{i}{2},\label{eq:stirling}
       \end{align}
       where $\sim$ denotes asymptotic equivalence. For $i=2(p+q)$, this yields the bound:
       \begin{align}
    \mathbb{E}_{v\sim \mathcal{N}(0,I_d)}[\lVert v\rVert^{2(p+q)}]
    =
    \mathcal{O}(d^{p+q}),
\end{align}
hence:
\begin{align}
    \left\lVert \mathbb{E}_{v\sim \mathcal{N}(0,I_d)}
    \left[g(\mu+\sigma_d v)\mathds{1}(A_d) \right]\right\rVert
    &\le   C'd^\frac{q}{2}\sqrt{\mathbb{P}(A_d)}+C'' \sigma_d^p d^\frac{p+q}{2}\sqrt{\mathbb{P}(A_d)},\\
    &=   (C'+C'' \sigma_d^p d^\frac{p}{2})d^\frac{q}{2}\sqrt{\mathbb{P}(A_d)},\\
    \label{eq:triangle_bound_ad}
\end{align}
We use the Gaussian concentration inequality for the Euclidean norm
\citep[Theorem 3.1.1]{vershynin2018hdp}, which states that for
$x\sim\mathcal N(0,I_d)$ there exists an absolute constant $K>0$ such that
for all $t\ge0$,
\begin{align}
    \mathbb{P}\left(\left\lvert\lVert x\rVert-\sqrt d\right\rvert\ge t\right)
    \le
    2\exp(-Kt^2).
\end{align}
In our setting, we need to bound:
\begin{align}
    \mathbb{P}(A_d)
    &=
    \mathbb{P}(\lVert\sigma_d v\rVert\ge \epsilon)
    =
    \mathbb{P}\left(\lVert v\rVert\ge \frac{\epsilon}{\sigma_d}\right) = \mathbb{P}\left(\lVert v\rVert - \sqrt{d} \ge \frac{\epsilon}{\sigma_d}-\sqrt{d}\right) .
\end{align}
Setting $t=\frac{\epsilon}{\sigma_d}-\sqrt{d}$, the assumption $\sqrt{d}\sigma_d = o(1)$ implies for sufficiently large $d$ that $\sqrt{d}\sigma_d \leq \epsilon$ and therefore $t \geq 0$, so we can apply the concentration bound to obtain:
\begin{align}
    \mathbb{P}(A_d)
    &= \mathbb{P}\left(\lVert v\rVert - \sqrt{d} \ge t\right) \leq \mathbb{P}\left(\Big\lvert\lVert v\rVert - \sqrt{d}\Big\rvert \ge t\right),\label{eq:concentration_ineq} \\
    &= \mathcal{O}\left(\exp\left(-K\left(\frac{\epsilon}{\sigma_d}-\sqrt{d}\right)^2\right)\right) = \mathcal{O}\left(\exp\left(-K\left(\frac{\epsilon}{\sigma_d}\right)^2\left(1-\frac{\sigma_d\sqrt{d}}{\epsilon}\right)^2\right)\right).
\end{align}
Now, as $\sqrt{d} \sigma_d=o(1)$, it follows $\frac{\sigma_d\sqrt{d}}{\epsilon}=o(1)$, yielding:
\begin{align}
    \mathbb{P}\left(A_d\right)
    =&\mathcal{O}\left(\exp\left(-K\left(\frac{\epsilon}{\sigma_d}\right)^2\right)\right),\\
    \implies \sqrt{\mathbb{P}\left(A_d\right)}
    =&\mathcal{O}\left(\exp\left(-\frac{K}{2}\left(\frac{\epsilon}{\sigma_d}\right)^2\right)\right),\\
    \implies d^\frac{q}{2} \sqrt{\mathbb{P}\left(A_d\right)}
    =&\mathcal{O}\left(d^\frac{q}{2}\exp\left(-\frac{K}{2}\left(\frac{\epsilon}{\sigma_d}\right)^2\right)\right),
\end{align}
 Applying these results to \cref{eq:triangle_bound_ad} , along with $\sigma_d^p d^\frac{p}{2} = \mathcal{O}(d^\frac{-p}{2})d^\frac{p}{2} = \mathcal{O}(1)$, yields our desired result:
\begin{align}
    \left\lVert \mathbb{E}_{v\sim \mathcal{N}(0,I_d)}
    \left[g(\mu+\sigma_d v)\mathds{1}(A_d) \right]\right\rVert\le& C' \mathcal{O}\left(d^\frac{q}{2}\exp\left(-\frac{K}{2}\left(\frac{\epsilon}{\sigma_d}\right)^2\right)\right)\\
    &\quad+C''\mathcal{O}(1) \mathcal{O}\left(d^\frac{q}{2}\exp\left(-\frac{K}{2}\left(\frac{\epsilon}{\sigma_d}\right)^2\right)\right),\\
    =&\mathcal{O}\left(d^\frac{q}{2}\exp\left(-K\left(\frac{\epsilon}{\sigma_d}\right)^2\right)\right). 
\end{align}
where we have absorbed the factor of $\frac{1}{2}$ into the constant $K$.
\end{proof}
\end{lemma}
Our proof in Lemma~\ref{proof:tail_bound} reveals the necessity of the condition $\sigma_d\sqrt d=o(1)$ for convergence as we can only apply the Gaussian concentration inequality in \cref{eq:concentration_ineq} for $\sigma_d\sqrt d=o(1)$; this is a direct consequence of the Gaussian annulus theorem, as for slower rates $1=o(\sigma_d\sqrt d)$, the Gaussian probability mass will exit any local ball around $\mu$ and flood the tail, meaning that the tail probability will grow with increasing $d$. Having bounded the tail, convergence to linearity follows by proving convergence within the ball, which allows us to exploit the local $C^1$ smoothness of $f(x)$:

\begin{theorem}[Convergence to Linearity]\label{app_proof:convergence_to_lin}
Let Assumptions~\ref{ass:continuous},~\ref{ass:polynomial} and~\ref{ass:bounded_derivative} hold and $\sigma_d=o\left(d^{-\frac{1}{2}}\right)$. Then:
\begin{align}
\left\lVert \nabla_\mu J(\theta)-\nabla f(\mu)\right\rVert=\Theta\left(\left(\sigma_d \sqrt{d}\right)^\alpha\right)=o(1),
\end{align}
 almost surely with respect to the distribution over $\mu$. 

\begin{proof}
We start with the definition of the ES update: 
 \begin{align}
   \nabla_\mu J(\theta)=  \frac{1}{\sigma_d}\mathbb{E}_{v\sim \mathcal{N}(0,I_d)}\left[v\cdot f(\mu+\sigma_d v)\right].
 \end{align}
Now let $\epsilon=\frac{\rho}{2}$ where $\rho$ is the radius of the ball from Assumption~\ref{ass:continuous}. Consider the hinge function: 
\begin{align}
    \phi(x)=\begin{cases}
        1,\quad &\lVert x\rVert\le\epsilon,\\
        2-\frac{\lVert x\rVert}{\epsilon},\quad &\epsilon<\lVert x\rVert <2\epsilon,\\
        0,\quad &\lVert x\rVert\ge 2\epsilon,
    \end{cases}
\end{align}
which interpolates between $1$ and $0$ in the region $\epsilon<\lVert x\rVert <2\epsilon$. Our first goal is to use $\phi(x)$ to generate a function $\tilde{f}(x)$ that is absolutely continuous and has integrable derivatives outside of $B_\rho(\mu)$ to allow us to apply Stein's lemma \citep{Stein1972}. We define $\tilde{f}(x)$ as:
\begin{align}
    \tilde{f}(x)=f(x)\cdot\phi(x-\mu)
\end{align}
Consider the closed ball $B_\epsilon(\mu)\coloneqq \{x'\vert \lVert x'-\mu\rVert \le \epsilon\} $. We note that within the ball $f(\mu+\sigma_d v)$ remains unchanged:  
\begin{align}
    \tilde{f}(\mu+\sigma_d v)=\begin{cases}
        f(\mu+\sigma_d v),\quad & \lVert \sigma_d v\rVert \le \epsilon,\\
        f(\mu+\sigma_d v)\cdot\left (2-\frac{\lVert \sigma_d v\rVert}{\epsilon}\right),\quad &\epsilon<\lVert \sigma_d v\rVert <2\epsilon,\\
        0,\quad &\lVert \sigma_d v\rVert\ge 2\epsilon.
    \end{cases}\label{eq:smoothed_fitness}
\end{align}
The derivative of the function with respect to $v$ is: 
\begin{align}
   \nabla_v \tilde{f}(\mu+\sigma_d v)=\begin{cases}
        \sigma_d\nabla f(\mu+\sigma_d v),\quad & \lVert \sigma_d v\rVert \le \epsilon,\\
       \sigma_d\nabla  f(\mu+\sigma_d v)\cdot \left (2-\frac{\lVert \sigma_d v\rVert}{\epsilon}\right)-\frac{\sigma_dv}{\epsilon\lVert v\rVert}\cdot f(\mu+\sigma_d v),\quad &\epsilon<\lVert \sigma_d v\rVert <2\epsilon,\\
        0,\quad &\lVert \sigma_d v\rVert\ge 2\epsilon.
    \end{cases}\label{eq:smoothed_derivatives}
\end{align}
where the gradient fails to exist only on the sets $\lVert \sigma_d v\rVert \in\{\epsilon,2\epsilon\}$, which have Lebesgue measure zero.
We start by using this function to decompose $J(\mu)$ into a smoothed part and a remainder: 
 \begin{align}
   \nabla_\mu J(\theta)=  \underbrace{\frac{1}{\sigma_d}\mathbb{E}_{v\sim \mathcal{N}(0,I_d)}\left[v\cdot \tilde{f}(\mu+\sigma_d v) \right]}_{\coloneqq \nabla_\mu \tilde{J}(\mu)}+\underbrace{\frac{1}{\sigma_d}\mathbb{E}_{v\sim \mathcal{N}(0,I_d)}\left[v\cdot (f(\mu+\sigma_d v)-\tilde{f}(\mu+\sigma_d v)) \right]}_{\coloneqq \Delta(\mu)},
 \end{align}
 Hence:
 \begin{align}
 \left\lVert \nabla_\mu J(\theta)-\nabla f(\mu)\right\rVert\le \left\lVert \nabla_\mu \tilde{J}(\mu)-\nabla f(\mu)\right\rVert+\left\lVert \Delta(\mu)\right\rVert.\label{eq:smooth_decomp}
 \end{align}
Consider the smoothed part:
\begin{align}
    \nabla_\mu \tilde{J}(\mu)\coloneqq\frac{1}{\sigma_d}\mathbb{E}_{v\sim \mathcal{N}(0,I_d)}\left[v\cdot \tilde{f}(\mu+\sigma_d v) \right].
\end{align} 
Our goal is to apply Stein’s lemma \citep{Stein1972} in its multivariate form \citep[Lemma~1]{Liu94}. The assumptions of \citep[Lemma~1]{Liu94} require that the partial derivatives $\partial_{v_i} \tilde{f}(\mu+\sigma_d v) $ are absolutely continuous almost everywhere and:
\begin{align}
    \mathbb{E}_{v\sim \mathcal{N}(0,I_d)}\left[\lvert \partial_{v_i} \tilde{f}(\mu+\sigma_d v) \rvert\right]<\infty.
\end{align}
These two conditions are satisfied by construction. Indeed, under Assumption~\ref{ass:continuous}, $f(\cdot)$ is $C^1$ continuous on $B_\rho(\mu)$, hence from \cref{eq:smoothed_fitness}, $\tilde{f}(\cdot)$ coincides with a compactly supported, piecewise $C^1$ function whose gradient (\cref{eq:smoothed_derivatives}) exists almost everywhere. Moreover, under Assumption~\ref{ass:polynomial}. both $f(\mu+\sigma_d v)$ and $\nabla f(\mu+\sigma_d v)$ are polynomially bounded, and since $\nabla \tilde{f}(\mu+\sigma_d v)$ is supported on $\lVert \sigma_d v\rVert \le 2\epsilon$, it follows that: 
\begin{align}
    \mathbb{E}_{v\sim \mathcal{N}(0,I_d)}\left[\lVert \nabla \tilde{f}(\mu+\sigma_d v) \rVert\right]<\infty.
\end{align}
Applying \citep[Lemma~1]{Liu94}:
\begin{align}
   \frac{1}{\sigma_d}\mathbb{E}_{v\sim \mathcal{N}(0,I_d)}\left[v\cdot f(\mu+\sigma_d v)\right]&=
   \frac{1}{\sigma_d} \mathbb{E}_{v\sim \mathcal{N}(0,I_d)}\left[\nabla_v \tilde{f}(\mu+\sigma_d v)\right]\label{eq:steins},\\
   &=
\mathbb{E}_{v\sim \mathcal{N}(0,I_d)}\left[\nabla \tilde{f}(\mu+\sigma_d v)\right],\\
    \implies \lVert \nabla_\mu \tilde{J}(\mu)-\nabla f(\mu)\rVert&=\left\lVert\mathbb{E}_{v\sim \mathcal{N}(0,I_d)}\left[\nabla \tilde{f}(\mu+\sigma_d v)-\nabla f(\mu)\right]\right\rVert\\
    &\le\mathbb{E}_{v\sim \mathcal{N}(0,I_d)}\left[\left\lVert\nabla \tilde{f}(\mu+\sigma_d v)-\nabla f(\mu)\right\rVert\right].
\end{align}
 Let $\{\mu+\sigma_d v \in B_{\epsilon}(\mu)\}=\{\Vert \sigma_d v\rVert \le\epsilon\}$ denote the event that a mutation lies within the ball $ B_{\epsilon}(\mu)$. We now split the integral into two regions, the first within the ball and the second outside:
\begin{align}
    \mathbb{E}_{v\sim \mathcal{N}(0,I_d)}\left[\left\lVert\nabla f(\mu+\sigma_d v)-\nabla f(\mu)\right\rVert\right]=&\underbrace{\mathbb{E}_{v\sim \mathcal{N}(0,I_d)}\left[\left\lVert\nabla \tilde{f}(\mu+\sigma_d v)-\nabla f(\mu)\right\rVert\mathds{1}(\lVert \sigma_d v\rVert \le \epsilon) \right]}_{\coloneqq I_\textrm{loc}}\\
    &+\underbrace{\mathbb{E}_{v\sim \mathcal{N}(0,I_d)}\left[\left\lVert\nabla \tilde{f}(\mu+\sigma_d v)-\nabla f(\mu)\right\rVert\mathds{1}(\lVert \sigma_d v\rVert > \epsilon) \right]}_{\coloneqq I_\textrm{tail}}.\label{eq:loc_tail_split}
\end{align}
Consider the region inside the ball, $I_\textrm{loc}$. From \cref{eq:smoothed_derivatives}, $\nabla \tilde{f}(\mu+\sigma_d v)=\nabla f(\mu+\sigma_d v)$ within this region. Using the local $\alpha$-H\"{o}lder continuity from Assumption~\ref{ass:continuous}:
    \begin{align}
            I_\textrm{loc}=&   \mathbb{E}_{v\sim \mathcal{N}(0,I_d)}\left[\left\lVert\nabla f(\mu+\sigma_d v)-\nabla f(\mu)\right\rVert\mathds{1}(\lVert \sigma_d v\rVert \le\epsilon) \right] ,\\
    \le & L\mathbb{E}_{v\sim \mathcal{N}(0,I_d)}\left[\left\lVert\sigma_d v\right\rVert^\alpha\mathds{1}(\lVert \sigma_d v\rVert \le\epsilon)\right],\\
    \le&{\sigma_d}^\alpha L \mathbb{E}_{v\sim \mathcal{N}(0,I_d)}\left[\left\lVert v\right\rVert^\alpha\right].
    \end{align}
    Now, applying the identity  $\mathbb{E}_{v\sim\mathcal{N}(0,I_d)}\left[ \left\lVert v\right\rVert^{i}\right]\sim d^\frac{i}{2}$, from \cref{eq:stirling}:
    \begin{align}
            I_\textrm{loc}=\mathcal{O}\left(\left(\sigma_d \sqrt{d}\right)^\alpha\right).
    \end{align}
We now bound the tail region outside the ball: 
\begin{align}
    I_\textrm{tail}&=\mathbb{E}_{v\sim \mathcal{N}(0,I_d)}\left[\left\lVert\nabla \tilde{f}(\mu+\sigma_d v)-\nabla f(\mu)\right\rVert\mathds{1}(\lVert \sigma_d v\rVert > \epsilon) \right],\\
    &\le \mathbb{E}_{v\sim \mathcal{N}(0,I_d)}\left[\left\lVert\nabla \tilde{f}(\mu+\sigma_d v)-\nabla f(\mu)\right\rVert\mathds{1}(\lVert \sigma_d v\rVert \ge \epsilon) \right].
\end{align}
Now, as $\lVert \nabla f(\mu)\rVert = \mathcal{O}(1)$ from Assumption~\ref{ass:bounded_derivative} and we have established that $\lVert\nabla \tilde{f}(\mu+\sigma_d v)\rVert$ is polynomial bounded under Assumption~\ref{ass:polynomial} when applying Stein's lemma, it follows that $\left\lVert\nabla \tilde{f}(\mu+\sigma_d v)-\nabla f(\mu)\right\rVert$ is also polynomial bounded, that is there exists some constant $C>0$ and finite polynomial order $p $ such that:
\begin{align}
    \left\lVert\nabla \tilde{f}(\mu+\sigma_d v)-\nabla f(\mu)\right\rVert\le C(1+\lVert \mu+\sigma_d v\rVert^p).
\end{align}
Applying Lemma~\ref{proof:tail_bound}, it follows:
\begin{align}
        I_\textrm{tail}= \mathcal{O}\left(
  \exp\left(-K\left(\frac{\epsilon}{\sigma_d}\right)^2\right)
   \right),
    \end{align}
    for some constant $K>0$. Together, this yields:
    \begin{align}
        \lVert \nabla_\mu \tilde{J}(\mu)-\nabla f(\mu)\rVert&=  I_\textrm{loc}+I_\textrm{tail},\\
        &=\mathcal{O}\left(\left(\sigma_d \sqrt{d}\right)^\alpha\right)+\mathcal{O}\left(
  \exp\left(-K\left(\frac{\epsilon}{\sigma_d}\right)^2\right)\right).
    \end{align}
    As $\exp(-x)=o\left(x^{-a}\right)$ for any $a > 0$, we take $a = \alpha/2$ to obtain a weakened bound matching the first term:
    \begin{align}
  \exp\left(-K\left(\frac{\epsilon}{\sigma_d}\right)^2\right) = o\left( \left(\frac{\sigma_d}{\epsilon}\right)^{\alpha}\right) 
=o\left(\left(\sigma_d \sqrt{d}\right)^\alpha\right).
    \end{align}
 This yields the upper bound:
 \begin{align}
     \lVert \nabla_\mu \tilde{J}(\mu)-\nabla f(\mu)\rVert &=  \mathcal{O}\left(\left(\sigma_d \sqrt{d}\right)^\alpha\right). \label{eq:smoothed_bound}
 \end{align}
 Returning to \cref{eq:smooth_decomp}, we must bound the remainder term:
 \begin{align}
     \left\lVert \Delta(\mu)\right\rVert&=\left\lVert \frac{1}{\sigma_d}\mathbb{E}_{v\sim \mathcal{N}(0,I_d)}\left[v\cdot (f(\mu+\sigma_d v)-\tilde{f}(\mu+\sigma_d v)) \right]\right\rVert,\\
     &\le  \frac{1}{\sigma_d}\mathbb{E}_{v\sim \mathcal{N}(0,I_d)}\left[\lVert v\rVert\cdot\left\lvert (f(\mu+\sigma_d v)-\tilde{f}(\mu+\sigma_d v))\right\rvert \right].
 \end{align}
 Again, from Assumption~\ref{ass:polynomial}, it follows that $\left\lvert (f(\mu+\sigma_d v)-\tilde{f}(\mu+\sigma_d v))\right\rvert$ is polynomially bounded, that is there exists some constant $C'>0$ and finite polynomial order $p' $ such that:
 \begin{align}
     \left\lvert (f(\mu+\sigma_d v)-\tilde{f}(\mu+\sigma_d v))\right\rvert\le C'(1+\lVert \mu+\sigma_d v\rVert^p).
 \end{align}
 Applying Lemma~\ref{proof:tail_bound} with $q=1$:
\begin{align}
    \left\lVert \Delta(\mu)\right\rVert= \mathcal{O}\left(\frac{d^\frac{1}{2}}{\sigma_d}
  \exp\left(-K\left(\frac{\epsilon}{\sigma_d}\right)^2\right)
   \right).
\end{align}
Now, as $\exp(-x)=o\left(x^{-1}\right)$ for $x\rightarrow \infty$, it follows:
\begin{align}
      \exp\left(-K\left(\frac{\epsilon}{\sigma_d}\right)^2\right)&=o\left(\sigma_d^2\right),\\
      \implies\left\lVert \Delta(\mu)\right\rVert&= \mathcal{O}\left( \frac{d^\frac{1}{2}}{\sigma_d}
  \exp\left(-K\left(\frac{\epsilon}{\sigma_d}\right)^2\right)\right),\\
  &=o(\sigma_d\sqrt{d} ),\\
  &=o\left( (\sigma_d\sqrt{d})^\alpha\right),\label{eq:remainder_bound}
\end{align}
where the final line follows from the fact $\sqrt{d}\sigma_d=o(1)$.
Assembling our bounds using Ineq.~\ref{eq:smooth_decomp} yields our desired result:
\begin{align}
    \left\lVert \nabla_\mu J(\theta)-\nabla f(\mu)\right\rVert&\le \underbrace{\left\lVert \nabla_\mu \tilde{J}(\mu)-\nabla f(\mu)\right\rVert}_{=O\left( (\sigma_d\sqrt{d})^\alpha\right),\ \textrm{Eq.}~\ref{eq:smoothed_bound}}+\underbrace{\left\lVert \Delta(\mu)\right\rVert}_{=o(\left( (\sigma_d\sqrt{d})^\alpha\right),\ \textrm{Eq.}~\ref{eq:remainder_bound}}
    =O\left( (\sigma_d\sqrt{d})^\alpha\right).
\end{align}
 
  We now show that the bound is tight. Consider the function $f(x)=\frac{L}{2} \sum_{i=1}^d x_i \lvert x_i\rvert +a^\top x$ where $\lVert a\rVert=\mathcal{O}(1)$. Taking partial derivatives:
  \begin{align}
      \partial_{i} f(x)=L\lvert x_i\rvert+a_i,\label{eq:example_partials}
  \end{align}
  hence:
  \begin{align}
      \lVert \nabla f(x)-\nabla f(y)\rVert=L\sqrt{\sum_{i=1}^d (\lvert x_i\rvert- \lvert y_i\rvert)^2}
  \end{align}
Applying the reverse triangle inequality $\lvert \lvert x_i\rvert- \lvert y_i\rvert\rvert\le \lvert x_i-  y_i\rvert \implies(\lvert x_i\rvert- \lvert y_i\rvert)^2\le (  x_i -  y_i)^2 $:
\begin{align}
      \lVert \nabla f(x)-\nabla f(y)\rVert\le L\sqrt{\sum_{i=1}^d (x_i- y_i)^2} =L\lVert x-y\rVert.
  \end{align}
  We have thus shown that $f(x)$ is $C^1$-continuous and its gradient has Lipschitz constant $L$, i.e. $\alpha=1$ with H\"{o}lder constant $L$. It is also bounded by a polynomial of order 2. Without loss of generality, we take a deterministic initialisation $\mu=0$ to simplify algebra,  yielding;
  \begin{align}
      \nabla f(\mu)=a\implies \lVert \nabla f(\mu)\rVert =\lVert  a\rVert=\mathcal{O}(1).
  \end{align}
  $f(x)$ thus satisfies Assumptions~\ref{ass:continuous},~\ref{ass:polynomial} and~\ref{ass:bounded_derivative}. Using $f(x)$ as the fitness:
  \begin{align}
      \nabla_\mu J(\theta)-\underbrace{\nabla f(\mu)}_{=a}&= \frac{1}{\sigma_d}\mathbb{E}_{v\sim \mathcal{N}(0,I_d)}\left[v\cdot f(\sigma_d v)\right]-a,\\
      &=\mathbb{E}_{v\sim \mathcal{N}(0,I_d)}\left[\nabla f(\sigma_d v)\right]-a;
  \end{align}
  Taking expectations element-wise and using \cref{eq:example_partials}:
   \begin{align}
       [\nabla_\mu J(\theta)-\nabla f(\mu)]_i&= \mathbb{E}_{v \sim \mathcal{N}(0,I_d) }\left[\partial_i f(\sigma_d v)\right]-a_i,\\
       &=\sigma_d L \mathbb{E}_{v_i \sim \mathcal{N}(0,1) }[\lvert v_i\rvert].
   \end{align}
   Applying \cref{eq:stirling}:
   \begin{align}
       \mathbb{E}_{v_i \sim \mathcal{N}(0,1) }[\lvert v_i\rvert]=\sqrt{2}\frac{\Gamma(1)}{\Gamma(\frac{1}{2})}=\sqrt{\frac{2}{\pi}},
   \end{align}
   Hence:
   \begin{align}
        \lVert \nabla_\mu J(\theta)-\nabla f(\mu)\rVert=\sigma_d\sqrt{d}\cdot L \sqrt{\frac{2}{\pi}},
   \end{align}
   thereby attaining the upper bound rate of $\sigma_d \sqrt{d}$.
\end{proof}
\end{theorem}

\subsection{Critical Convergence Rate}\label{app:critical}
To show that our rate is critical, we investigate the space of functions that can be represented by cubic polynomials of the form:
\begin{align}
    f(x)=a^\top x+\frac{1}{2}x^\top B x+\frac{1}{6}C[x,x,x],\label{eq_app:cubic}
\end{align}
where $a\in\mathbb{R}^d$, $B\in\mathbb{R}^{d\times d}$ is a symmetric matrix and  $C[x,x,x]=\sum_{i,j,k} c_{i,j,k} x_ix_j x_k$ denotes a symmetric 3-linear map represented by the 3-tensor $C\in\mathbb{R}^{d\times d\times d}$.

Since our theory depends on analysing the local stability of a smooth ball for a fitness function, stability over this class is necessary for convergence on more general objectives. We show that once $\sigma_d$ decays slower than the critical rate, divergence already occurs within this subclass, establishing the sharpness of the rate.

\begin{theorem}[Exact divergence for cubic objectives]
Let $f(x)$ denote the cubic polynomial in \cref{eq_app:cubic}. Assume $\lVert a\rVert=\mathcal{O}(1)$,$\lVert B\rVert=\mathcal{O}(1)$, $\lVert C\rVert=\mathcal{O}(1)$ where $\lVert\cdot \rVert$ denotes operator norm for $i$-tensor $T(x_1,\dots x_i)$: $\lVert T\rVert\coloneq\sup_{\lVert x_1\rVert=\dots =\lVert x_i\rVert=1}\lvert T(x_1,\dots x_i)\rvert$. Let Assumption~\ref{ass:bounded_derivative} hold, then:
\begin{align}
    \nabla_\mu J(\theta)
    =
    \nabla f(\mu)
    + \frac{\sigma_d^2}{2}
    \mathbb{E}_{v\sim\mathcal{N}(0,I_d)}
    \left[
         C(v,v,\cdot)
    \right].
    \end{align}
    Moreover:
    \begin{align}
    \left\lVert
    \frac{\sigma_d^2}{2}\mathbb{E}_{v\sim\mathcal{N}(0,I_d)}
    \left[
        C(v,v,\cdot)
    \right]
    \right\rVert
    &=
    \Theta(\sigma_d^2d),\\
    \left\lVert
    \nabla_\mu J(\theta)-\nabla f(\mu)
    \right\rVert&=
    \Theta(\sigma_d^2 d).
\end{align}
\begin{proof}
We start by taking derivatives of $f(x)$:
\begin{align}
    \nabla f(x)
    =
    a + Bx + \frac{1}{2} C(x,x,\cdot).
\end{align}

Substituting this into the definition of $\nabla_\mu J(\theta)$ and using \cref{eq:steins}:
\begin{align}
    \nabla_\mu J(\theta)&=\frac{1}{\sigma_d}\mathbb{E}_{v\sim\mathcal{N}(0,I_d)}
    \left[vf(\mu+\sigma_d v)\right],
    \\
    &=\mathbb{E}_{v\sim\mathcal{N}(0,I_d)}
    \left[\nabla f(\mu+\sigma_d v)\right],\\
    &=\mathbb{E}_{v\sim\mathcal{N}(0,I_d)}
    \left[a + B(\mu+\sigma_d v) + \frac{1}{2} C(\mu+\sigma_d v,\mu+\sigma_d v,\cdot)\right],\\
    &=a + B\mu+\sigma_d B\underbrace{\mathbb{E}_{v\sim\mathcal{N}(0,I_d)}[v]}_{=0} + \frac{1}{2} \mathbb{E}_{v\sim\mathcal{N}(0,I_d)}
    \left[C(\mu+\sigma_d v,\mu+\sigma_d v,\cdot)\right],\\
    &=a + B\mu + \frac{1}{2} \mathbb{E}_{v\sim\mathcal{N}(0,I_d)}
  \left[C(\mu,\mu,\cdot)+\sigma_dC( v,\mu,\cdot)+\sigma_d C(\mu ,v,\cdot)+\sigma_d ^2C(v , v,\cdot)\right],\\
   &=\underbrace{a + B\mu + \frac{1}{2}C(\mu,\mu,\cdot)}_{=\nabla f(\mu)}+\frac{1}{2} \mathbb{E}_{v\sim\mathcal{N}(0,I_d)}
  \left[2\sigma_dC(v,\mu,\cdot)+\sigma_d ^2C(v , v,\cdot)\right],\\
  &=\nabla f(\mu)+\sigma_d\mathbb{E}_{v\sim\mathcal{N}(0,I_d)}
  \left[C(v,\mu,\cdot)\right]+\frac{\sigma_d ^2}{2}\mathbb{E}_{v\sim\mathcal{N}(0,I_d)}
  \left[C(v , v,\cdot)\right],
\end{align}
where we have used the fact $C(v,\mu,\cdot)=C(\mu,v,\cdot)$ by definition of the symmetry of C. As $C( v,\mu,\cdot)$ is linear in $v$, its expectation under zero-mean $\mathcal{N}(0,I_d)$ is zero, hence:
\begin{align}
    \nabla_\mu J(\theta)&=\nabla f(\mu)+\frac{\sigma_d ^2}{2}\mathbb{E}_{v\sim\mathcal{N}(0,I_d)}
  \left[C(v , v,\cdot)\right],
\end{align}
proving our first result. Now, it follows that $\lVert C(v,v,\cdot)\rVert \le \lVert C\rVert \lVert v\rVert^2$ and as $\lVert C\rVert=\mathcal{O}(1)$:
\begin{align}
    \lVert \mathbb{E}_{v\sim\mathcal{N}(0,I_d)}
  \left[C(v , v,\cdot)\right]\rVert\le  &\lVert C\rVert\mathbb{E}_{v\sim\mathcal{N}(0,I_d)}\left[\lVert v \rVert^2\right],\\
  =&\mathcal{O}(\mathbb{E}_{v\sim\mathcal{N}(0,I_d)}\left[\lVert v \rVert^2\right])
\end{align}
Now as $v$ is unit Gaussian: $\mathbb{E}_{v\sim\mathcal{N}(0,I_d)}\left[\lVert v \rVert^2\right]=d$, hence:
\begin{align}
    \lVert \mathbb{E}_{v\sim\mathcal{N}(0,I_d)}
  \left[C(v , v,\cdot)\right]\rVert=\mathcal{O}(d).
\end{align}
       
       We now show that the bound is tight. Consider the function $f(x)=u^\top x\lVert x\rVert^{2}$ for  $u^\top=\frac{1}{\sqrt{d}}[1,\dots 1]$. The factor of $\frac{1}{\sqrt{d}}$ ensures that the gradient of the function $\nabla_x f(x)=\mathcal{O}(1)$. We can write $\lVert x\rVert^{2}$ as the tensor contraction:
       \begin{align}
           \lVert x\rVert^{2}={I_d}(x,x),
       \end{align}
       where $I_d$ is the identity matrix and: 
       \begin{align}
           u^\top (x)=u(x),
       \end{align}
       hence we write $f(x)$ as a tensor contraction as:
       \begin{align}
           f(x)= C(x,x,x),
       \end{align}
       where $C\coloneqq \operatorname{Sym}( u\otimes I_d)$.  Using this function: 
       \begin{align}
       C(v , v,\cdot)&=\nabla_v (u^\top v\lVert v\rVert^{2})\\
       &=u\lVert v\rVert^{2}+2v u^\top v,\\
           \implies\mathbb{E}_{v\sim\mathcal{N}(0,I_d)}
  \left[C(v , v,\cdot)\right]&=u\underbrace{\mathbb{E}_{v\sim\mathcal{N}(0,I_d)}
  \left[\lVert v\rVert^2\right]}_{=d}+2\underbrace{\mathbb{E}_{v\sim\mathcal{N}(0,I_d)}
  \left[ v v^\top\right]}_{=I_d}u,\\
  &=u(d+2),
       \end{align}
hence $\lVert \mathbb{E}_{v\sim\mathcal{N}(0,I_d)}
  \left[C(v , v,\cdot)\right]\rVert=d+2$, achieving the upper bound rate of $\mathcal{O}(d)$ which implies:
  \begin{align}
      \lVert \mathbb{E}_{v\sim\mathcal{N}(0,I_d)}
  \left[C(v , v,\cdot)\right]\rVert=\Theta(d).
  \end{align}
  Our final result follows immediately:
  \begin{align}
      \lVert \nabla_\mu J(\theta)-\nabla f(\mu)\rVert=\frac{(\sigma_d)^2}{2}\lVert \mathbb{E}_{v\sim\mathcal{N}(0,I_d)}
  \left[C(v , v,\cdot)\right]\rVert=\Theta(\sigma_d^2 d). \qquad\qedhere
  \end{align}
\end{proof}

\end{theorem}

\subsection{EGGROLL Linearisation}\label{app:egg_lin}

We now study the effect of EGGROLL in high dimensions. We introduce the notation $v=\textrm{vec}(E)$ to denote the vectorisation of the low-rank matrix perturbation $E=\frac{1}{\sqrt{r}}A B^\top$ and work in vector space. The EGGROLL vector update $v$ can thus be written  as sum of independent variables:
        \begin{align}
            v= \sum_{i=1}^r \frac{1}{\sqrt{r}} u_i
        \end{align}
with:
\begin{align}
    u_i=\textrm{vec}\left(a_i b_i^\top  \right),
\end{align}
where recall $a_i$ and $b_i$ are the $i$th column vectors of $A$ and $B$. We write $\mu=\textrm{vec}(M)$. Using \cref{eq:frobenius_l2}, we can convert between results in vector space and matrix space as:
\begin{align}
    \left\lVert  \textnormal{\textrm{vec}}(\hat{g}_\textrm{LR})-\nabla f(\mu)\right\rVert= \left\lVert  \hat{g}_\textrm{LR}-\nabla_W f(W=M)\right\rVert_F,\\
     \left\lVert  \textnormal{\textrm{vec}}(\hat{g}_\textrm{LR})-\nabla_\mu J(\theta)\right\rVert= \left\lVert  \hat{g}_\textrm{LR}-\nabla_M J(\theta)\right\rVert_F.
\end{align}

To extend our analysis, we need to ensure that all polynomial moments of $P(v)$ are finite and grow at most polynomially in the dimension $d=mn$. In particular, such tail bounds are sufficient to dominate polynomial error terms in our analysis. To introduce sub-Gaussian variables, we follow the exposition of \citet{vershynin2018hdp} and results therein. A random variable $x_i\in\mathbb{R}$ is sub-Gaussian if there exists some finite constant $C>0$ such that for all $t>0$:
\begin{align}
    \mathbb{P}(\lvert x_i \rvert >t)\le 2\exp(-C t^2),
\end{align}
meaning their their tails decay like Gaussians. This is equivalent to any of the following three properties holding \citep[2.6.1]{vershynin2018hdp}:
There exist constants $C_1,C_2,C_3>0$ that differ at most by an absolute constant factor such that:
\begin{align}
    \left(\mathbb{E}[\lvert x_i\rvert^p]\right)^\frac{1}{p}&\le C_1 \sqrt{p},\quad \forall p\ge1,\\
    \mathbb{E}\left[\exp\left(\frac{x_i^2}{C_2^2}\right)\right]&\le 2,
\end{align}
and if $\mathbb{E}[x_i]=0$:
\begin{align}
    \mathbb{E}\left[\exp(\lambda x_i)\right]\le \exp(C_3^2 \lambda^2),\quad \forall \lambda\in\mathbb{R}.\label{eq:exp_sub}
\end{align}
A random vector $x\in\mathbb{R}^d$ is sub-Gaussian if all one-dimensional marginals of $x$ are sub-Gaussian, i.e. $x^\top u$ is sub-Gaussian for all $u\in\mathbb{R}^d$. The sub-Gaussian norm is defined as:
\begin{align}
\lVert x\rVert_{\psi_2}\coloneqq\inf_K\left\{K\Bigg\vert \mathbb{E}\left[\exp\left(u^\top (x-\mathbb{E}[x])\right)\right]\le \exp\left(\frac{K^2\lVert u\rVert^2}{2}\right),\quad \forall u\in\mathbb{R}^d\right\}.
\end{align}
which returns the smallest universal sub-Gaussian constant for all marginals. 

A key property of sub-Gaussian vectors that we use in our proofs is the sub-Gaussian concentration inequality for the Euclidean norm
\citep[Theorem 3.1.1]{vershynin2018hdp}, which states that for if $x$ is a sub-Gaussian vector with $\mathbb{E}[x_i^2]=1$ and $K=\lVert x\rVert_{\psi_2}$, there exists an absolute constant $C>0$ such that
for all $t\ge0$,
\begin{align}
    \mathbb{P}\left(\left\lvert\lVert x\rVert-\sqrt m\right\rvert\ge t\right)
    \le
    2\exp\left(-\frac{C{t}^2}{K^4}\right).\label{eq:sub_Gauss_ineq}
\end{align}
We also use a weaker form of control, that replaces the Gaussian-like tail decay with an exponential decay, but all other properties are defined similarly. In this paper, we use the definition that a variable $x$ is known as sub-exponential if there exists a $K>0$ such that for all $t\ge0$:
        \begin{align}
          \mathbb{P}\left(\lvert x\rvert \ge t\right)\le 2 \exp\left(-\frac{t}{ K}\right).
        \end{align}

Our first result derives a bound on the expected value of the norms $\lVert a\rVert^i$ and $\lVert b\rVert^i$:

\begin{lemma} Let Assumption~\ref{ass:convergence_sampling} hold. Let $P(a)$ denote the distribution over columns of $A$ and $P(b)$ denote the distribution over columns of $B$. Then: 
\begin{align}
    \mathbb{E}_{a\sim P(a)}[\lVert a\rVert^i]=\mathcal{O}(m^\frac{i}{2}),\quad \mathbb{E}_{b\sim P(b)}[\lVert b\rVert^i]=\mathcal{O}(n^\frac{i}{2}).
\end{align}
 \begin{proof}\label{proof:expectation_growth}
        It suffices to prove $\mathbb{E}_{a\sim P(a)}[\lVert a\rVert^i]=\mathcal{O}(m^\frac{i}{2})$ as $\mathbb{E}_{b\sim P(b)}[\lVert b\rVert^i]=\mathcal{O}(n^\frac{i}{2})$ follows automatically from the same assumptions. We start by using the `layer cake' representation of the expectation \citet[Theorem 1.13]{Lieb10}:
        \begin{align}
            \mathbb{E}_{a\sim P(a)}[\lVert a\rVert^i]=i\int^\infty_0 t^{i-1} \mathbb{P}(\lVert a\rVert>t) dt.
        \end{align}
        Let $t_m= C\sqrt{m}$ for any $C>1$. We split the integral into two regions: 
        \begin{align}
            i\int^\infty_0 t^{i-1} \mathbb{P}(\lVert a\rVert>t) dt=\int_0^{t_m}it^{i-1} \mathbb{P}(\lVert a\rVert>t) dt+\int_{t_m}^\infty it^{i-1} \mathbb{P}(\lVert a\rVert>t) dt.
        \end{align}
        For the first integral:
        \begin{align}
            \int_0^{t_m}it^{i-1} \mathbb{P}(\lVert a\rVert>t) dt&\le \int_0^{t_m} it^{i-1} dt,\\
           &=(t_m)^i,\\
           &=C^i m^\frac{i}{2}.
        \end{align}
        For the second integral, we wish to bound $\mathbb{P}(\lVert a\rVert>t)$ for the region $t\ge t_m=C\sqrt m$. Setting $t'=t-\sqrt{m}>0$, the assumption $C > 1$ implies $t' \geq 0$ in this region, hence
        \begin{align}
            \mathbb{P}(\lVert a\rVert>t)=\mathbb{P}(\lVert a\rVert-\sqrt m>t') \leq \mathbb{P}(\lvert \lVert a\rVert-\sqrt m\rvert>t').
        \end{align}
       We bound this using the sub-Gaussian concentration inequality from \cref{eq:sub_Gauss_ineq}. Under Assumption~\ref{ass:convergence_sampling}, $a$ is a sub-Gaussian vector with $\lVert x\rVert_{\psi_2}\le \infty$, hence there exists an absolute constant $C'>0$ such that
for all $t'\ge0$,
\begin{align}
    \mathbb{P}\left(\left\lvert\lVert a\rVert-\sqrt m\right\rvert\ge t'\right)
    \le
    2\exp\left(-C'{t'}^2\right).
\end{align}
 This implies:
\begin{align}
    \mathbb{P}\left(\lVert a\rVert\ge t\right)\le 2\exp\left(-C'(t-\sqrt{m})^2\right),
\end{align}
for all $t\ge t_m$. Substituting yields:
\begin{align}
    \int_{t_m}^\infty it^{i-1} \mathbb{P}(\lVert a\rVert>t) dt\le \int_{t_m}^\infty i t^{i-1} \exp\left(-C'(t-\sqrt{m})^2\right)dt.
\end{align}
Let $x=t-\sqrt{m}\implies dt=dx$: 
\begin{align}
    \int_{t_m}^\infty it^{i-1} \mathbb{P}(\lVert a\rVert>t) dt\le \int_{\sqrt{m}(C-1)}^\infty i (x+\sqrt{m})^{i-1} \exp\left(-C'x^2\right)dx.
\end{align}
Now, $\sqrt{m}\le \frac{x}{C-1} $ for all $x\ge \sqrt{m}(C-1)$, hence:
\begin{align}
    \int_{t_m}^\infty it^{i-1} \mathbb{P}(\lVert a\rVert>t) dt&\le \int_{\sqrt{m}(C-1)}^\infty i x^{i-1} \left(1+\frac{1}{C-1}\right)^{i-1} \exp\left(-C'x^2\right)dx,\\
    &= i\left(1+\frac{1}{C-1}\right)^{i-1}\int_{\sqrt{m}(C-1)}^\infty  x^{i-1}  \exp\left(-C'x^2\right)dx,\\
    &\le i\left(1+\frac{1}{C-1}\right)^{i-1}\int_{0}^\infty  x^{i-1}  \exp\left(-C'x^2\right)dx,\\
    &\le i\left(1+\frac{1}{C-1}\right)^{i-1}\frac{1}{2}(C')^{-i/2}\Gamma\!\left(\frac{i}{2}\right),\\
    &=\mathcal{O}(1).
\end{align}
Combining the two bounds yields:
\begin{align}
     \mathbb{E}_{a\sim P(a)}[\lVert a\rVert^i]=\mathcal{O}(m^\frac{i}{2}),
\end{align}
as required.
\end{proof}
\end{lemma}

Using this result, we now bound the whole vector $v=\frac{1}{\sqrt{r}}\sum_{i=1}^r \textrm{vec}(a_i b_i^\top)$

\begin{lemma}\label{proof:low_rank_chi}
   Let $i\ge1$. Under Assumption~\ref{ass:convergence_sampling}:
    \begin{align}
        \mathbb{E}_{v\sim P(v)}\left[\lVert v\rVert^i \right]=\mathcal{O}\left( (rmn)^\frac{i}{2}\right)
    \end{align}
       \begin{proof}
      For any vectors $a, b$:
        \begin{align}
            \lVert \textrm{vec}(a b^\top)\rVert &= \sqrt{\sum_{j=1}^m\sum_{k=1}^n(a_j b_k)^2} = \sqrt{\sum_{j=1}^m{a_j}^2\sum_{k=1}^n {b_k}^2} =\lVert a\rVert \lVert b\rVert,\\
            \implies\lVert \textrm{vec}(a b^\top)\rVert^i&= \lVert a\rVert^i\lVert b\rVert^i \,.
        \end{align}
        Applying Lemma~\ref{proof:expectation_growth} under Assumption~\ref{ass:convergence_sampling} for each summand of $v=\frac{1}{\sqrt{r}}\sum_{l=1}^r \textrm{vec}(a_l b_l^\top)$:
        \begin{align}
            \mathbb{E}_{v\sim P(v)}\left[\lVert \textrm{vec}(a_l b_l^\top)\rVert^i\right]&=\mathbb{E}_{a_l\sim P(a_l)}\left[\lVert a_l\rVert^i \right]\mathbb{E}_{b_l\sim p(b_l)}\left[\lVert b_l\rVert^i \right],\\
            &=\mathcal{O}\left((mn)^\frac{i}{2}\right).\label{eq:egg_norm_i}
        \end{align}
            Applying the triangle inequality: 
            \begin{align}
                 \mathbb{E}_{v\sim P(v)}\left[\lVert v\rVert^i \right]&=\mathbb{E}_{v\sim P(v)}\left[\left\lVert \frac{1}{\sqrt{r}} \sum_{l=1}^r\textrm{vec}(a_l b_l^\top)\right\rVert^i \right],\\
                 &\le \mathbb{E}_{v\sim P(v)}\left[ \left( \frac{1}{\sqrt{r}} \sum_{l=1}^r\left\lVert\textrm{vec}(a_l b_l^\top)\right\rVert\right)^i \right],\\
                 &=r^{\frac{i}{2}}\mathbb{E}_{v\sim P(v)}\left[\left( \frac{1}{r}\sum_{l=1}^r\left\lVert\textrm{vec}(a_l b_l^\top)\right\rVert\right)^i \right].
            \end{align}
            Now, as $i\ge 1$, we can apply Jensen's inequality: 
            \begin{align}
                \left(\frac{1}{r}\sum_{l=1}^r\left\lVert\textrm{vec}(a_l b_l^\top)\right\rVert\right)^i \le \frac{1}{r} \sum_{l=1}^r\lVert \textrm{vec}(a_l b_l^\top) \rVert^i,
            \end{align}
            yielding: 
            \begin{align}
                \mathbb{E}_{v\sim P(v)}\left[\lVert v\rVert^i \right]&\le r^{\left(\frac{i}{2} - 1\right)} \sum_{l=1}^r\mathbb{E}_{v\sim P(v)}\left[\lVert \textrm{vec}(a_l b_l^\top)\rVert^i\right] =r^\frac{i}{2}\mathcal{O}\left((mn)^\frac{i}{2}\right) =\mathcal{O}\left(\left(rmn\right)^\frac{i}{2}\right). \qedhere
            \end{align}
    \end{proof}
\end{lemma}
Our proof borrows techniques used to prove linearisation of the ES update in \cref{app:es_lin} by bounding the tail probability of any polynomial under the low-rank distribution outside of the ball $B_\rho(\mu)$. To apply the concentration inequality that would generalise Lemma~\ref{proof:tail_bound}, we show that $v$ has an exponentially decaying tail:

\begin{lemma}[Exponential Tail Bound]\label{proof:exp_tail_bound}
    Let $r<\infty$ and Assumption~\ref{ass:convergence_sampling} hold. Then all elements of $v$ are sub-exponential and for $\sqrt{d}\sigma_d =o(1)$ there exists some constant $C>0$ such that:
    \begin{align}
        \mathbb{P}(\lVert \sigma_d v \rVert\ge \rho)\le 2d \exp\left(-C\frac{\rho}{\sqrt{d}\sigma_d}\right).
    \end{align}
    \begin{proof}
        In matrix form:
        \begin{align}
            E=\frac{1}{\sqrt{r}}\sum_{i=1}^r a_i b_i^\top.
        \end{align}
        The elements of $E$ are thus: 
        \begin{align}
            E_{j,k}=\frac{1}{\sqrt{r}}\sum_{i=1}^r {a_i}_j{b_i}_k.
        \end{align}
        As ${a_i}_j$ and ${b_i}_k$ are independent sub-Gaussian random variables with zero mean, it follows from \citet[Lemma 2.8.6]{vershynin2018hdp} that their product ${a_i}_j {b_i}_k$ is a zero-mean sub-exponential variable with a uniform norm $\lVert {a_i}_j {b_i}_k\rVert_{\psi_1}<\infty$. Finally, a finite sum of sub-exponential variables is sub-exponential \citep[Eq. (2.18)]{Wainwright_2019} with a uniform norm, so all elements of $E$ and hence $v=\textrm{vec}(E)$ are sub-exponential and zero-mean  with a uniform $\psi_1$-norm  $K<\infty$.
        
        We now bound $\mathbb{P}(\lVert \sigma_d v \rVert\ge \rho)=\mathbb{P}(\lVert v \rVert\ge \frac{\rho}{\sigma_d})$. For the vector $v$, it follows for $t\ge 0$:
        \begin{align}
            \lVert v\rVert\ge t \implies \max_j\lvert v_j\rvert \ge \frac{t}{\sqrt{d}}.
        \end{align}
        This is easily proven via the contrapositive: if $\max_j\lvert v_j\rvert < \frac{t}{\sqrt{d}}$ then
        \begin{align}
            \lVert v\rVert^2=\sum_{j=1}^d v_j^2 <d \frac{t^2}{d}=t^2,
        \end{align}
        implying $\lVert v\rVert < t$. This means for $t\ge 0$:
        \begin{align}
            \mathbb{P}(\lVert v \rVert\ge t)&\le \mathbb{P}\left(\max_j\lvert v_j\rvert \ge \frac{t}{\sqrt{d}}\right),\\
            &\le \sum_{j=1}^d \mathbb{P}\left(\lvert v_j\rvert \ge \frac{t}{\sqrt{d}}\right).\label{eq:max_exp_bound}
        \end{align}
        
        As $v_{j}$ is a sub-exponential variable with finite uniform sub-exponential norm, by definition \citep[Proposition 2.8.1]{vershynin2018hdp} there exists a finite $K$ such that for all $j$:
        \begin{align}
          \mathbb{P}\left(\lvert v_{j}\rvert \ge \frac{t}{\sqrt{d}}\right)\le 2 \exp\left(-\frac{t}{\sqrt d K}\right).
        \end{align}
        Applying to \cref{eq:max_exp_bound} yields:
        \begin{align}
            \mathbb{P}(\lVert v \rVert\ge t)\le2d \exp\left(-\frac{t}{\sqrt d K}\right).
        \end{align}
        Now, using $t=\frac{\rho}{\sigma_d}$ and $C=\frac{1}{K}$ yields:
        \begin{align}
            \mathbb{P}(\lVert \sigma_d v \rVert\ge \rho)&\le 2d \exp\left(-C\frac{\rho}{\sqrt{d}\sigma_d}\right). \qedhere
        \end{align}
    \end{proof}
\end{lemma}

We now use these results to assemble into our key polynomial tail bound:

\begin{lemma}[EGGROLL Polynomial Tail Bounds]\label{proof:egg_tail_bound}
Let Assumption~\ref{ass:convergence_sampling} hold. Let $g(x)$ be polynomial bounded as:
\begin{align}
    \lVert g(x)\rVert\le C(1+\lVert x\rVert^p),
\end{align}
for some finite polynomial of order $p$ and constant $C>0$. Consider the ball $B_{\rho}(\mu)\coloneqq \{ x'\vert \lVert x'- \mu\rVert< \rho\}$. Let $\{\mu+\sigma_d v \in B_{\rho}(\mu)\}=\{\Vert \sigma_d v\rVert < \rho\}$ denote the event that a mutation lies outside the ball.
Assume $\sigma_d=o(d^{-1/2})$. Then for some constant $K>0$ independent of $d$:
\begin{align}
   \left\lVert \mathbb{E}_{v\sim P(v)}
   \left[g(\mu+\sigma_d v)\mathds{1}(A_d) \right]\right\rVert
   =
  \mathcal{O}\left(\sqrt{d} \exp\left(-K\frac{\rho}{\sqrt{d}\sigma_d}\right)\right),
\end{align}
and in particular the right-hand side is $o(1)$ as $d\to\infty$.

\begin{proof}
Let $$A_d\coloneqq \{\mu+\sigma_d v\in B_{\rho}(\mu)\}$$ and denote
$\mathbb{P}(A_d)\coloneqq
\mathbb{E}_{v\sim P(v)}[\mathds{1}(A_d)]$. Our proof proceeds as in Lemma~\ref{proof:tail_bound} to obtain:
\begin{align}
    \left\lVert \mathbb{E}_{v\sim P(v)}
    \left[g(\mu+\sigma_d v)\mathds{1}(A_d) \right]\right\rVert
    &\le C' \mathbb{P}(A_d) + C'' \sigma_d^p \mathbb{E}_{v\sim P(v)} \left[ \lVert v\rVert^p \mathds{1}(A_d) \right] .
\end{align}
where $C' = C(1+2^{p-1}\lVert \mu \rVert^p)$ and $C'' = C2^{p-1}$ are constant in $d$. Applying the Cauchy--Schwarz inequality to the second expectation gives:
\begin{align}
    \mathbb{E}_{v\sim P(v)}
    [\lVert v\rVert^p \mathds{1}(A_d)]
    &\le
    \sqrt{
    \mathbb{E}_{v\sim P(v)}[\lVert v\rVert^{2p}]
    }
    \cdot
    \sqrt{\mathbb{P}(A_d)}.
\end{align}
Applying Lemma~\ref{proof:low_rank_chi} with fixed $r$ and $d=mn$:
\begin{align}
     \sqrt{\mathbb{E}_{v\sim P(v)}[\lVert v\rVert^{2p}]}=\mathcal{O}\left(d^\frac{p}{2}\right).
\end{align}
Now, $\mathbb{P}(A_d)=\mathbb{P}(\lVert \sigma_d v \rVert\ge \rho)$. From Lemma~\ref{proof:exp_tail_bound}, there exists some $K>0$ such that:
\begin{align}
    \mathbb{P}(\lVert \sigma_d v \rVert\ge \rho)\le 2d\exp\left(-K\frac{\rho}{\sqrt{d}\sigma_d}\right),\\
    \implies \sqrt{\mathbb{P}(A_d)}=\mathcal{O}\left(\sqrt{d} \exp\left(-K\frac{\rho}{\sqrt{d}\sigma_d}\right)\right),
\end{align}
where we have absorbed the factor of $\frac{1}{2}$ into $K$, hence:
\begin{align}
    \mathbb{E}_{v\sim P(v)}
    [\lVert v\rVert^p \mathds{1}(A_d)]
    &=\mathcal{O}\left(d^{\frac{p+1}{2}}\exp\left(-K\frac{\rho}{\sqrt{d}\sigma_d}\right) \right).
\end{align}
Now, as $\sqrt d\sigma_d=o(1) $, $\sigma^p_d d^\frac{p}{2}=o(1)$, hence:
\begin{align}
    \sigma_d^p \mathbb{E}_{v\sim P(v)} \left[ \lVert v\rVert^p \mathds{1}(A_d) \right]=\mathcal{O}\left(\sqrt d\exp\left(-K\frac{\rho}{\sqrt{d}\sigma_d}\right) \right).
\end{align}
Applying our bounds yields our desired result:
\begin{align}
    \left\lVert \mathbb{E}_{v\sim P(v)}
    \left[g(\mu+\sigma_d v)\mathds{1}(A_d) \right]\right\rVert=\mathcal{O}\left(\sqrt d\exp\left(-K\frac{\rho}{\sqrt{d}\sigma_d}\right) \right)=o(1).
\end{align}
where the $o(1)$ bound follows from the fact that the exponential factor dominates $\sqrt{d}$ and $\sqrt{d}\sigma_d=o(1)$.
\end{proof}
\end{lemma}

\begin{theorem}[EGGROLL Convergence to Linearity]\label{app_proof:EGG_convergence_to_lin}
Let Assumptions~\ref{ass:polynomial},~\ref{ass:bounded_derivative},~\ref{ass:egg_continuous} and~\ref{ass:convergence_sampling}  hold and $\sigma_d=o(d^{-1/2})$ and $L_d(\sigma_d d)^2=o(1)$. Then there exists some $K>0$ such that:
\begin{align}
\left\lVert  \textnormal{\textrm{vec}}(\hat{g}_\textrm{LR})-\nabla f(\mu)\right\rVert&=\mathcal{O}\left(L_d(\sigma_d d)^2\right)+\mathcal{O}\left(\frac{\sqrt{d}}{\sigma_d^2}  \exp\left(-K\frac{\rho}{\sqrt{d}\sigma_d}\right) \right)=o(1),\\
\lVert \textnormal{\textrm{vec}}(\hat{g}_\textrm{LR})-\nabla_\mu J(\theta) \rVert&=\mathcal{O}\left(\sigma_d \sqrt d\cdot\left(1+L_d\sigma_dd^\frac{3}{2}\right)\right)=o(1).
\end{align}
 almost surely with respect to the distribution over $\mu$. 

\begin{proof}
We start with the definition of the vectorised EGGROLL update: 
 \begin{align}
   \textnormal{\textrm{vec}}(\hat{g}_\textrm{LR})-\nabla f(\mu)&=  \frac{1}{\sigma_d}\mathbb{E}_{v\sim P(v)}\left[v\cdot f(\mu+\sigma_d v)\right]-\nabla f(\mu),\\
   &=  \frac{1}{\sigma_d}\mathbb{E}_{v\sim P(v)}\left[v\cdot f(\mu+\sigma_d v)\right]-\frac{1}{\sigma_d}\underbrace{\mathbb{E}_{v\sim P(v)}[v]}_{=0}\cdot f(\mu)-\underbrace{\mathbb{E}_{v\sim P(v)}[vv^\top]}_{I_d}\nabla f(\mu)\\
   &\quad+\frac{1}{2\sigma_d}\underbrace{\mathbb{E}_{v\sim P(v)}[{\sigma_d}^2v v^\top\nabla^2 f(\mu)v ]}_{=0},\\
   &=  \frac{1}{\sigma_d}\mathbb{E}_{v\sim P(v)}\Bigg[v\cdot \Bigg(\underbrace{f(\mu+\sigma_d v)-f(\mu)-\sigma_d v^\top\nabla f(\mu)+\frac{{\sigma_d}^2}{2}v^\top\nabla^2 f(\mu)v}_{\coloneqq T_d(v)}\Bigg)\Bigg],\\
   &=  \frac{1}{\sigma_d}\mathbb{E}_{v\sim P(v)}\left[v\cdot T_d(v)\right],
 \end{align}
 where we have used the fact that the expectation of an odd function under a symmetric, zero mean distribution is always zero, and $P(v)$ satisfies this under Assumption~\ref{ass:convergence_sampling}, hence $\mathbb{E}_{v\sim P(v)}[v v^\top\nabla^2 f(\mu)v]=0$, and $\mathbb{E}_{v\sim P(v)}[vv^\top]=I_d$ from Lemma~\ref{proof:covariance_form}. Consider the  ball $B_{\rho}(\mu)\coloneqq \{ x'\vert \lVert x' - \mu\rVert< \rho\}$. 
We now split the integral into two regions, the first within the ball and the second outside:
\begin{align}
    \frac{1}{\sigma_d}\mathbb{E}_{v\sim P(v)}\left[v\cdot (f(\mu+\sigma_d v)-f(\mu))\right]=&\underbrace{ \frac{1}{\sigma_d}\mathbb{E}_{v\sim P(v)}\left[v\cdot T_d(v)\mathds{1}(\lVert \sigma_d v\rVert < \rho) \right]}_{\coloneqq I_\textrm{loc}}\\
    &+\underbrace{ \frac{1}{\sigma_d}\mathbb{E}_{v\sim P(v)}\left[v\cdot T_d(v)\mathds{1}(\lVert \sigma_d v\rVert \ge \rho) \right]}_{\coloneqq I_\textrm{tail}}.
\end{align}

Consider the region inside the ball:
\begin{align}
   \lVert I_\textrm{loc}\rVert&=\frac{1}{\sigma_d}\left\lVert \mathbb{E}_{v\sim P(v)}\left[v\cdot T_d(v)\mathds{1}(\lVert \sigma_d v\rVert < \rho) \right]\right\rVert,\\
   &\le\frac{1}{\sigma_d} \mathbb{E}_{v\sim P(v)}\left[\lVert v\rVert \left\lvert T_d(v)\right\rvert\mathds{1}(\lVert \sigma_d v\rVert < \rho) \right].\label{eq:local_egg}
\end{align}
Within this region, $f(\mu+\sigma_d v)$ is $C^2$ continuous under Assumption~\ref{ass:egg_continuous}. We can thus write $f(\mu+\sigma_d v)$ using a first-order Taylor expansion about $\mu$ with a Hessian (second order derivative) remainder within the ball:
\begin{align}
  f(\mu+\sigma_d v)&=f(\mu)+\sigma_d\nabla f(\mu)^\top v+{\sigma_d}^2 v^\top \left( \int_0^1 (t-1) \nabla^2 f(\mu+t\sigma_dv)dt \right) v,\\
  \implies T_d(v)&={\sigma_d}^2 v^\top \left( \int_0^1 (t-1) \nabla^2 f(\mu+t\sigma_dv)dt \right) v+\frac{{\sigma_d}^2}{2}v^\top\nabla^2 f(\mu) v,\\
  &={\sigma_d}^2 v^\top \left( \int_0^1 (t-1) (\nabla^2 f(\mu+t\sigma_dv)-\nabla^2 f(\mu))dt \right) v.
\end{align}
Applying the Lipschitz bound on the Hessian from Assumption~\ref{ass:egg_continuous}:
\begin{align}
    \left\lvert T_d(v)\right\rvert&\le{\sigma_d}^2 \lVert v\rVert^2 \left\lVert \int_0^1 (t-1) (\nabla^2 f(\mu+t\sigma_dv)-\nabla^2 f(\mu))dt \right\rVert,\\
    &\le{\sigma_d}^2 \lVert v\rVert^2  \int_0^1 (t-1) \left\lVert\nabla^2 f(\mu+t\sigma_dv)-\nabla^2 f(\mu)\right\rVert dt ,\\
    &\le{\sigma_d}^2 \lVert v\rVert^2  \int_0^1 (t-1) L_d\left\lVert t\sigma_dv\right\rVert dt ,\\
        &={\sigma_d}^3 \lVert v\rVert^3 L_d \left\lvert \int_0^1 (t-1)  t dt \right\rvert,\\
        &=\frac{ L_d}{6}  {\sigma_d}^3 \lVert v\rVert^3.
\end{align}
Using this to bound \cref{eq:local_egg}:
\begin{align}
    \lVert I_\textrm{loc}\rVert\le& \frac{ L_d}{6}  {\sigma_d}^2\mathbb{E}_{v\sim P(v)}\left[ \lVert v\rVert^4 \mathds{1}(\lVert \sigma_d v\rVert < \rho)\right],\\
    \le &\frac{ L_d}{6}  {\sigma_d}^2\mathbb{E}_{v\sim P(v)}\left[ \lVert v\rVert^4\right].
\end{align}
Now, (for fixed $r$) we apply the identity  $\mathbb{E}_{v\sim P(v)}\left[ \lVert v\rVert^4\right]=\mathcal{O}\left( (mn)^2\right)$ with $mn=d$ from Lemma~\ref{proof:low_rank_chi}:
\begin{align}
    \lVert I_\textrm{loc}\rVert=\mathcal{O}(L_d (\sigma_d d)^2).
\end{align}
   
We now bound the tail region outside the ball: 
\begin{align}
    I_\textrm{tail}&= \frac{1}{\sigma_d}\mathbb{E}_{v\sim P(v)}\left[v\cdot T_d(v)\mathds{1}(\lVert \sigma_d v\rVert \ge \rho) \right],\\
    &\le \frac{1}{\sigma_d}\mathbb{E}_{v\sim P(v)}\left[\lVert v\rVert\lvert T_d(v)\rvert\mathds{1}(\lVert \sigma_d v\rVert \ge \rho) \right],\\
     &=\frac{1}{\sigma_d^2}\mathbb{E}_{v\sim P(v)}\left[\lVert \sigma_d v\rVert\lvert T_d(v)\rvert\mathds{1}(\lVert \sigma_d v\rVert \ge \rho) \right].
\end{align}
Now under Assumptions~\ref{ass:polynomial},~\ref{ass:bounded_derivative} and~\ref{ass:egg_continuous}, $f(\mu+\sigma_d v)$ is polynomial bounded, $\lVert \nabla f(\mu)\rVert =\mathcal{O}(1)$  and $\lVert \nabla^2 f(\mu)\rVert $ is polynomial bounded hence there exists some finite constant $C>0$ and finite polynomial order $p$ such that:
\begin{align}
    \lVert\sigma_d v\rVert\lvert T_d(v)\rvert\le C(1+\lVert \mu+\sigma_d  v\rVert^p).
\end{align}
We thus apply Lemma~\ref{proof:egg_tail_bound}:
\begin{align}
    \frac{1}{\sigma_d^2}\mathbb{E}_{v\sim P(v)}\left[\lVert \sigma_d v\rVert\lvert T_d(v)\rvert\mathds{1}(\lVert \sigma_d v\rVert \ge \rho) \right]&=\mathcal{O}\left(\frac{\sqrt{d}}{\sigma_d^2}  \exp\left(-K\frac{\rho}{\sqrt{d}\sigma_d}\right) \right),\\
    &=\mathcal{O}\left(\frac{d\sqrt{d}}{d\sigma_d^2}  \exp\left(-K\frac{\rho}{\sqrt{d}\sigma_d}\right) \right).
\end{align}
Now, as $\sigma_d\sqrt{d}=o(1)$, the exponential term dominates the prefactor $\frac{d\sqrt{d}}{d\sigma_d^2}$, we conclude:
\begin{align}
    \frac{1}{\sigma_d^2}\mathbb{E}_{v\sim P(v)}\left[\lVert \sigma_d v\rVert\lvert T_d(v)\rvert\mathds{1}(\lVert \sigma_d v\rVert \ge \rho) \right]&=o(1) 
\end{align}
Our final result follows from:
\begin{align}
    \lVert \textnormal{\textrm{vec}}(\hat{g}_\textrm{LR})-\nabla_\mu J(\theta) \rVert&=\lVert \textnormal{\textrm{vec}}(\hat{g}_\textrm{LR})-\nabla f(\mu)+\nabla f(\mu)-\nabla_\mu J(\theta) \rVert,\\
    &\le\lVert \textnormal{\textrm{vec}}(\hat{g}_\textrm{LR})-\nabla f(\mu)\rVert+\lVert\nabla f(\mu)-\nabla_\mu J(\theta) \rVert.
\end{align}
We have already shown $\lVert \textnormal{\textrm{vec}}(\hat{g}_\textrm{LR})-\nabla f(\mu)\rVert=o(1)$ and under the assumptions for this theorem, \cref{app_proof:convergence_to_lin} holds and so $\lVert\nabla f(\mu)-\nabla_\mu J(\theta) \rVert= o(1)$. \qedhere
\end{proof}
\end{theorem}

\newpage

\section{Asymptotic Rank Analysis}\label{app:asymptotic_rank}
For convenience, we work with random vectors in our analysis. We analyse the vector $v^r=\textrm{vec}(E^r)$, which is the vectorisation of the low-rank matrix $E^r$. We denote $v=\textrm{vec}(E)$, which is the vectorisation of the full rank matrix $E$. Note $v\sim \mathcal{N}(0,I_d)$ which we denote as $P(v)$. We write $v^r$ as a standardised sum of $r$ independent, zero-mean random vectors. Let 
\begin{align}
    u_i=\textrm{vec}\left(a_i b_i^\top  \right),\label{eq:u_def}
\end{align}
where recall $a_i$ and $b_i$ are the $i$th column vectors of $A$ and $B$ so:
\begin{align}
    v^r=\frac{1}{\sqrt{r}} \sum_{i=1}^r u_i.
\end{align}
 Denoting the covariance matrix of $p(u)$ as $\Sigma_u$, the central limit theorem proves that the distribution of $v^r$ converges in distribution to a zero-mean Gaussian $\mathcal{N}(0,\Sigma_r)$. In \cref{proof:covariance_form}, we derive the covariance matrix for $\Sigma_u$, which we prove is the identity. Our analysis uses an Edgeworth expansion \citep{Bhattacharya76} to characterise precisely the rate at which $P(v^r) $ converges to the limiting Gaussian distribution. In \cref{proof:edgeworth}, we make an Edgeworth expansion of $P(v^r)$ to show that it is dominated by $\mathcal{O}\left(r^{-1}\right)$ terms and higher. These are then used to prove \cref{proof:distance_to_Gauss}, which allows us to bound the integral of the remainder of the Edgeworth expansion, thereby characterising how fast $P(v^r)$ converges to the limiting Gaussian distribution.

\begin{lemma} \label{proof:covariance_form} Let Assumption~\ref{ass:iid} hold and $u_i$ be defined in \cref{eq:u_def}. Then the variable $u_i$ has identity covariance matrix: 
    \begin{align}
        \Sigma_u\coloneqq\mathbb{E}_{u_i\sim p(u_i)}[u_i u_i^\top]=I_d,
    \end{align}
has finite $4$th-order absolute moments:
\begin{align}
    \mathbb{E}_{u_i\sim p(u_i)}\left[\lVert u_i\rVert^4\right]<\infty,
\end{align}
and the vector $v^r=\textrm{vec}(E^r)$ is zero-mean and has identity covariance matrix:
\begin{align}
    \Sigma_v\coloneq\mathbb{E}_{v^r\sim P(v^r)}[v^r {v^r}^\top]=I_d
\end{align}
    \begin{proof}
        Under the vec operator, the vector $u_i$ can be written element wise as:
        \begin{align}
            u_i=[a_1b_1, a_2b_1,\dots a_mb_1, a_1b_2,\dots a_mb_n]^\top.
        \end{align}
        We note that all elements in the vector $u_i$ have zero mean, and so the covariance matrix is the expectation of the outer product:
        \begin{align}
            \Sigma_u=\mathbb{E}_{u_i\sim p(u_i)}\left[u_i {u_i}^\top\right].
        \end{align}
        The diagonal elements of $\Sigma_u$ are:
        \begin{align}
            \mathbb{E}_{a_i,b_j}\left[(a_i b_j)^2\right]&=\mathbb{E}_{a_i}\left[a_i^2 \right]\mathbb{E}_{b_j}\left[b_j^2\right]=1.\label{eq:diag}
        \end{align}
        As all elements of $a$, $b$ and $\epsilon$ are zero-mean, off-diagonal elements are zero:
        \begin{align}
            \mathbb{E}_{a_i,b_j,a_k,b_l}&\left[a_ib_j a_kb_l\right]
            =0\quad i\ne k\ \textrm{or } j\ne l.\label{eq:off_diag}
        \end{align}
       Using \cref{eq:diag,eq:off_diag}, our first result follows:
       \begin{align}
           \Sigma_u= I_{d}.
       \end{align}
        Now, as $u_i$ is a vector of elements which are sums and products of variables which all have finite 4th order moments from Assumption~\ref{ass:iid}, it immediately follows that $u$ has finite $4$th order absolute moments. 
        
        For our final result, we can write $v^r$ as sum of independent variables:
        \begin{align}
            v^r= \sum_{i=1}^r \left(r^{-\frac{1}{2}} u_i\right)=\sum_{i=1}^r x_i,
        \end{align}
        where $x_i\coloneqq \frac{1}{\sqrt{r}} u_i$. As $v^r$ is a sum of zero-mean vectors, it is also zero-mean. We use the fact that the covariance of $r$ i.i.d. random variables is equal to the sum of the individual covariances, hence
        \begin{align}
            \mathbb{E}_{v^r}[v^r v^r] = &r \mathbb{E}_{x_i}[x_i x_i^\top],\\
            =& r \mathbb{E}_{u_i}\left[\frac{1}{r}u_i u_i^\top\right],\\
            =& \mathbb{E}_{u_i}\left[u_i u_i^\top\right],\\
            =& I_d,
        \end{align}
        as required.
    \end{proof}
\end{lemma}
Using \cref{proof:covariance_form}, we see the asymptotic Gaussian density of $v^r$ is a standard normal:
\begin{align}
    g(v^r)&=\frac{1}{\sqrt{(2\pi)^d}}\exp\left(-\frac{\lVert v^r\rVert^2}{2} \right).\label{eq:limiting_gauss}
\end{align}
which is the density of $P(v)$, where recall $v=\textrm{vec}(E)$, is the vectorisation of the full rank matrix $E$.

Although  $P(v^r)$ does not have a density in the usual sense for low-rank $r$, we can still approximate it with a distribution $\hat{p}(v^r)$ by making a Taylor series expansion of its characteristic function, which always exists regardless of whether $P(v^r)$ has a well-defined density or not. We now derive the 4th order Edgeworth expansion for $P(v^r)$. Our proof reveals that 3rd order cumulants control all terms in the expansion that decay at rate $\mathcal{O}\left(r^{-\frac{1}{2}}\right)$. As 3rd order cumulants are all zero due to symmetry in Assumption~\ref{ass:iid}, the overall decay rate is controlled by $\mathcal{O}\left(r^{-1}\right)$ terms associated with 4th order cumulants. It is for this reason that we obtain a faster convergence rate than the standard central limit theorem. 

\begin{lemma}\label{proof:edgeworth}
Let Assumption~\ref{ass:iid} hold and let $v^r=\textrm{vec}(E^r)$ and $u_i$ be defined in \cref{eq:u_def}. 
Let $g(v^r)$ denote the limiting Gaussian density in \cref{eq:limiting_gauss}. 
Then, the 2nd order Edgeworth expansion of $v^r$ is a distribution $\hat{P}(v^r)$ defined by the approximate density:
\begin{align}
    \hat{p}(v^r)
    &=
    g(v^r)
    +
    \frac{1}{4!r}
    g(v^r)
    \sum_{i,j,k,l}\kappa^4_{i,j,k,l} H_{i,j,k,l}(v^r),
\end{align}
where:
\begin{align}
    H_{i,j,k,l}(v^r)
    \coloneqq
    \exp\!\left(\frac{\lVert v^r\rVert^2}{2}\right)
    \frac{\partial^4}{\partial v^r_i\partial v^r_j\partial v^r_k\partial v^r_l}
    \exp\!\left(-\frac{\lVert v^r\rVert^2}{2}\right)
\end{align}
is a 4th order Hermite polynomial associated with $g(v^r)$
\citep{Laplace1811,Hall92,Withers00}.

\begin{proof}
We denote the characteristic function of $P(u_i)$ as:
\begin{align}
    \varphi_{U}(\omega)
    =
    \int \exp\left(-i\omega^\top u\right)dP(u),
\end{align}
and the characteristic function of $P(v^r)$ as:
\begin{align}
    \varphi_{r}(\omega)
    =
    \int \exp\left(-i\omega^\top u\right)dP(v^r).
\end{align}
Recall $v^r=\frac{1}{\sqrt{r}}\sum_{i=1}^r u_i$ is the sum of $r$ i.i.d. copies of $\frac{1}{\sqrt{r}}u_i$. 
Using the scaling property of the Fourier transform, the characteristic function of $\frac{1}{\sqrt{r}}u_i$ is
$\varphi_{U}\!\left(\frac{\omega}{\sqrt{r}}\right)$.
The distribution of a sum of $r$ independent random variables is given by the $r$-fold convolution of the individual distributions.
As convolution in the spatial domain corresponds to multiplication in the frequency domain, the characteristic function of $v^r$ is \citep{Bhattacharya76}:
\begin{align}
    \varphi_{r}(\omega)
    =
    \left(\varphi_{U}\!\left(\frac{\omega}{\sqrt{r}}\right)\right)^r.
\end{align}
Taking logarithms yields the log-characteristic function:
\begin{align}
    \log \varphi_{r}(\omega)
    &=
    r\log\!\left(\varphi_{U}\!\left(\frac{\omega}{\sqrt{r}}\right)\right),
    \\
    &=
    rK_U\!\left(\frac{\omega}{\sqrt{r}}\right),
\end{align}
where $K_U(\omega)\coloneqq\log\varphi_{U}(\omega)$.
The cumulants are defined by
\begin{align}
    \kappa^{(n)}_{i_1,\dots,i_n}
    \coloneqq
    i^{-n}
    \left.
    \frac{\partial^n K_U(\omega)}
    {\partial\omega_{i_1}\cdots\partial\omega_{i_n}}
    \right|_{\omega=0}.
\end{align}

The Edgeworth expansion proceeds by a Taylor expansion of
$rK_U\!\left(\frac{\omega}{\sqrt{r}}\right)$ about $\omega=0$.
A 4th order expansion yields:
\begin{align}
    rK_U\!\left(\frac{\omega}{\sqrt{r}}\right)
    &\approx
    rK_U(0)
    +
    \sqrt{r}
    \sum_i \omega_i \kappa^1_i
    +
    \frac{1}{2!}
    \sum_{i,j}\omega_i\omega_j\kappa^2_{i,j}
    \\
    &\quad+
    \frac{1}{3!\sqrt{r}}
    \sum_{i,j,k}\omega_i\omega_j\omega_k\kappa^3_{i,j,k}
    +
    \frac{1}{4!r}
    \sum_{i,j,k,l}\omega_i\omega_j\omega_k\omega_l\kappa^4_{i,j,k,l},
\end{align}
where $K_U(0)=0$.
Under Assumption~\ref{ass:ggd}, $u_i$ is symmetric, hence all odd-order cumulants vanish:
$\kappa^1=\kappa^3=0$.
The second-order cumulant satisfies
\begin{align}
    \sum_{i,j}\omega_i\omega_j\kappa^2_{i,j}
    =
    -\omega^\top\Sigma_u\omega,
\end{align}
and from \cref{proof:covariance_form} we have $\Sigma_u=I$.
Substituting yields:
\begin{align}
    rK_U\!\left(\frac{\omega}{\sqrt{r}}\right)
    &\approx
    -\frac{\lVert\omega\rVert^2}{2}
    +
    \frac{1}{4!r}
    \sum_{i,j,k,l}\omega_i\omega_j\omega_k\omega_l\kappa^4_{i,j,k,l}.
\end{align}

Exponentiating and expanding the exponential to first-order in $1/r$ gives:
\begin{align}
    \varphi_r(\omega)
    &=
    \exp\!\left(rK_U\!\left(\frac{\omega}{\sqrt{r}}\right)\right),
    \\
    &\approx
    \exp\!\left(-\frac{\lVert\omega\rVert^2}{2}\right)
    \left(
    1+
    \frac{1}{4!r}
    \sum_{i,j,k,l}\omega_i\omega_j\omega_k\omega_l\kappa^4_{i,j,k,l}
    \right).
\end{align}

Taking the inverse Fourier transform (with the convention
$\mathcal{F}^{-1}(f)(v)=(2\pi)^{-d}\int e^{i\omega^\top v}f(\omega)d\omega$)
yields:
\begin{align}
    \hat{p}(v^r)
    &=
    g(v^r)
    +
    \frac{1}{4!r}
    \sum_{i,j,k,l}\kappa^4_{i,j,k,l}
    \frac{\partial^4}{\partial v^r_i\partial v^r_j\partial v^r_k\partial v^r_l}
    g(v^r),
\end{align}
and using the identity
\(
H_{i,j,k,l}(v^r)
=
g(v^r)^{-1}
\frac{\partial^4}{\partial v^r_i\partial v^r_j\partial v^r_k\partial v^r_l}
g(v^r)
\),
we recover the stated Edgeworth density.
\end{proof}
\end{lemma}

We now apply key results from \citet{Bhattacharya76} to bound the difference in expectation between the low-rank distribution and the Edgeworth approximation as well as the difference in expectation between the true ES Gaussian distribution and the Edgeworth approximation.

\begin{lemma} \label{proof:distance_to_Gauss} Let $f(v)\coloneqq f(M=\mu+\sigma \textrm{mat}(v)) $, let $P(v)=\mathcal{N}(0,I_{d})$, $P(v^r)$ be the distribution of $v^r$ and $\hat{P}(v^r)$ be the 2nd order Edgeworth expansion of $P(v^r)$.  Let Assumptions~\ref{ass:iid} and~\ref{ass:boundedness} hold and let $v^r=\textrm{vec}(E^r)$ and $u_i$ be defined in \cref{eq:u_def}. 
    Then:
    \begin{align}
       \left\lVert \mathbb{E}_{v^r\sim P(v^r)}\left[v^r \cdot  f(v^r)\right]-\mathbb{E}_{v^r\sim \hat{P}(v^r)}\left[v^r \cdot  f(v^r)\right]\right\rVert &=\mathcal{O}\left(r^{-1}\right),\\
    \left\lVert \mathbb{E}_{v\sim P(v)}\left[v \cdot  f(v)\right]-\mathbb{E}_{v\sim \hat{P}(v)}\left[v \cdot  f(v)\right]\right\rVert &=\mathcal{O}\left(r^{-1}\right).
\end{align}
\begin{proof}
  From \cref{proof:edgeworth}, we have shown that the Edgeworth expansion for $P(v^r)$ is controlled by 4th order cumulants and higher, that is;
  \begin{align}
    \hat{p}(v^r)&=g(v^r)+\frac{1}{4!r}g(v^r)\sum_{i,j,k,l}\kappa^4_{i,j,k,l} H_{i,j,k,l}(v^r).\label{eq:edgeworth}
  \end{align}
  We show that the three assumptions needed to apply \citet[Theorem 20.1]{Bhattacharya76} to obtain our result using \cref{eq:edgeworth} hold. Firstly, the boundedness assumption of the integrand holds:
    \begin{align}
        \sup_{v^r} \frac{\lVert f(v^r) v^r\rVert}{1+\rVert v^r\rVert}\le \sup_{v^r}\lvert  f(v^r)\rvert<\infty.
    \end{align}
     Secondly, the sampling regularity assumption that $u_i$ (as defined in \cref{eq:u_def}) is zero-mean i.i.d. (satisfied under Assumption~\ref{ass:iid}) with finite 4th order moments (satisfied from \cref{proof:covariance_form}) holds. Let  $\varphi_U(\omega)$ denote the characteristic function of $p(u)$, then the final assumption we need to verify is the Cramer condition: $\limsup_{\lVert \omega\rVert\rightarrow \infty} \varphi_U(\omega)<1$, which is satisfied from the Riemann-Lebesgue lemma \citet[Theorem~8.22]{Folland1999} because $p_0(\cdot)$ is absolutely continuous under Assumption~\ref{ass:iid} and hence $\lvert \varphi_U(\omega)\rvert\rightarrow 0$ as $\lVert \omega \rVert\rightarrow 0$. Our first result thus follows from applying \citet[Theorem 20.1]{Bhattacharya76}: 
    \begin{align}
         \left\lVert \mathbb{E}_{v^r\sim P(v^r)}\left[v^r \cdot  f(v^r)\right]-\mathbb{E}_{v^r\sim \hat{P}(v^r)}\left[v^r \cdot  f(v^r)\right]\right\rVert &=\mathcal{O}\left(r^{-1}\right).
    \end{align}

We now derive our second result.
\begin{align}
    \mathbb{E}_{v\sim \hat{P}(v)}\left[v \cdot  f(v)\right]=\int v \cdot  f(v) g(v)\left(1+\frac{1}{4! r} \sum_{i,j,k,l}\kappa^4_{i,j,k,l} H_{i,j,k,l}(v)\right)dv,\\
    =\mathbb{E}_{v\sim P(v)}\left[v \cdot  f(v)\right]-\int v \cdot  f(v) g(v)\frac{1}{4! r} \sum_{i,j,k,l}\kappa^4_{i,j,k,l} H_{i,j,k,l}(v)dv,
\end{align}
hence
\begin{align}
    \lVert \mathbb{E}_{v\sim P(v)}\left[v \cdot  f(v)\right]-\mathbb{E}_{v\sim \hat{P}(v)}\left[v \cdot  f(v)\right]\rVert&=\frac{1}{r}\left\lVert \int v \cdot  f(v) \frac{1}{4! } \sum_{i,j,k,l}\kappa^4_{i,j,k,l} H_{i,j,k,l}(v) g(v) dv\right\rVert,\\
    &\le \frac{1}{r} \int  \left\lVert v\right \rVert \cdot \lvert f(v)\rvert \frac{1}{4! } \sum_{i,j,k,l}\lvert \kappa^4_{i,j,k,l} H_{i,j,k,l}(v) \rvert g(v) dv.
\end{align}
Now by definition, $H_{i,j,k,l}(v)$ is a 4th order Hermite polynomial and under Assumption~\ref{ass:boundedness}, $\lvert f(v)\rvert$ is bounded, hence $\left\lVert v\right \rVert \cdot \lvert f(v)\rvert \frac{1}{4! r} \sum_{i,j,k,l}\lvert \kappa^4_{i,j,k,l} H_{i,j,k,l}(v) \rvert $ has polynomial growth of order 5 and is bounded by:
\begin{align}
    \left\lVert v\right \rVert \cdot \lvert f(v)\rvert \frac{1}{4! } \sum_{i,j,k,l}\lvert \kappa^4_{i,j,k,l} H_{i,j,k,l}(v) \rvert\le C(1+\lVert v\rVert^5)
\end{align}
for some finite $C>0$. As the expectation of a finite order polynomial under $\mathcal{N}(0,I_d)$ is bounded, it thus follows: 
\begin{align}
      \lVert \mathbb{E}_{v\sim P(v)}\left[v \cdot  f(v)\right]-\mathbb{E}_{v\sim \hat{P}(v)}\left[v \cdot  f(v)\right]\rVert
    &\le \frac{1}{r} \int  C(1+\lVert v\rVert^5)g(v) dv=\mathcal{O}\left(r^{-1}\right),
\end{align}
as required.

\end{proof}
\end{lemma}
Using \cref{proof:distance_to_Gauss}, we have all ingredients needed derive our main about the convergence result, which follows after some simple algebra on the norm: 
 \begin{theorem}\label{proof_app:approximator_error}
Let Assumptions~\ref{ass:iid} and~\ref{ass:boundedness} hold, then:
\begin{align}
    \lVert\nabla_\mu J(\theta)-\hat{g}_\textnormal{\textrm{LR}}^r\rVert_F=\mathcal{O}\left(r^{-1}\right).
\end{align}
\begin{proof}
We start by converting the Frobenius norm to vector form using \cref{eq:frobenius_l2}: 
\begin{align}
    \lVert \nabla_\mu J(\theta)-g_\textnormal{\textrm{LR}}^r\rVert_F&=\left\lVert \frac{1}{\sigma}(\textrm{vec}\left(\mathbb{E}_{E}\left[E\cdot f(W=M+\sigma E)\right]\right)-\textrm{vec}\left(\mathbb{E}_{E^r}\left[E^r \cdot f(W=M+\sigma E^r)\right]\right))\right\rVert,\\
     &=\left\lVert\frac{1}{\sigma}( \mathbb{E}_{E}\left[\textrm{vec}(E) f(W=M+\sigma E)\right]-\mathbb{E}_{E^r}\left[\textrm{vec}(E^r) f(W=M+\sigma E^r)\right])\right\rVert,\\
     &=\left\lVert\frac{1}{\sigma}( \mathbb{E}_{v}\left[v f(v)\right]-\mathbb{E}_{v^r}\left[v^r f(v^r)\right])\right\rVert,\\
    \end{align}
    where $f(v)\coloneqq f(M=\mu+\sigma \textrm{mat}(v)) $ and $v=\textrm{vec}(E)$ is the vectorisation of variable $E$, which is distributed as $v\sim P(v)\coloneqq\mathcal{N}(0,I_{d}) $. Let $\hat{P}(v)$ be the distribution for the 2nd order Edgeworth expansion, which we derived in \cref{proof:edgeworth}. Since $\hat{P}(v^r)$ and $\hat{P}(v)$ are identified as the same Edgeworth-expanded distribution on $\mathbb{R}^d$, we may equivalently write:
    \begin{align}
        \mathbb{E}_{v^r\sim \hat{P}(v^r)}\left[v^r f(v^r)\right]= \mathbb{E}_{v\sim \hat{P}(v)}\left[v^r f(v)\right],
    \end{align}
    hence:
    \begin{align}
        \mathbb{E}_{v}\left[v f(v)\right]-\mathbb{E}_{v^r}\left[v^r f(v^r)\right]&=\mathbb{E}_{v}\left[v f(v)\right]-\mathbb{E}_{v\sim \hat{P}(v)}\left[v f(v)\right]+\mathbb{E}_{v^r\sim \hat{P}(v^r)}\left[v^r f(v^r)\right]-\mathbb{E}_{v^r}\left[v^r f(v^r)\right],\\
        \implies  \lVert\nabla_\mu J(\theta)-\hat{g}_\textnormal{\textrm{LR}}^r\rVert_F&\le\frac{1}{\sigma} \left\lVert \mathbb{E}_{v}\left[v f(v)\right]-\mathbb{E}_{v\sim \hat{P}(v)}\left[v f(v)\right]\right\rVert\\
        &\quad+\frac{1}{\sigma}\left\lVert\mathbb{E}_{v^r\sim \hat{P}(v^r)}\left[v^r f(v^r)\right]-\mathbb{E}_{v^r}\left[v^r f(v^r)\right]\right\rVert.
    \end{align}
    Applying \cref{proof:distance_to_Gauss} to each bound yields our desired result:
    \begin{align}
        \lVert\nabla_\mu J(\theta)-\hat{g}_\textnormal{\textrm{LR}}^r\rVert_F&=\mathcal{O}\left(r^{-1}\right).
    \end{align}
\end{proof}
\end{theorem}

\subsection{Mean Field Score Function Approximator}\label{app:mean-field}

We will use $n$th order Bessel functions of the second kind $K_n(z)$ \citep{Basset1892,Macdonald1899,Watson1922}, which are conveniently represented by the integral equations:
\begin{align}
    K_n(z)=\int_0^\infty \exp(-z \cosh\theta)\cosh(n\theta)d\theta.
\end{align}
Bessel functions are the solutions to systems of differential equations that occur naturally in phenomena where there is strong radial symmetry, typically involving the propagation of spherical waves from  points like the ripples formed from water droplets \citep{Whitham99}. For our setting, Bessel functions describe the probability density of the product of rotationally invariant random variables, whose solution is analogous to the interference pattern of two spherical wave propagators. 

Using the representation, we find the derivative of the zeroth order function takes the recursive form:
\begin{align}
    \frac{d K_0(z) }{dz}=-\int_0^\infty \exp(-z \cosh\theta)\cosh(\theta)d\theta=-K_1(z).\label{eq:bessel_derivative}
\end{align}
More generally, the derivative of the $n$th order Bessel function is \citet[Section 3.71, Eq. 4]{Watson1922}:
\begin{align}
    \frac{dK_n(z)}{dz}=\frac{n}{z}K_n(z)-K_{n+1}(z).\label{eq:derivative_general}
\end{align}

\subsection{Derivation of Mean-field Approximators}
To derive a mean-field approximation, we assume that the elements of $A$ and $B$ are drawn independently from the set of generalised Gaussian distributions (GGDs):
\begin{assumption}\label{ass:ggd}
    Assume each element $a_{i,j}\sim \mathcal{GG}(s,p)$ and $b_{i,j}\sim \mathcal{GG}(s,p)$ of $A$ and $B$ is independently distributed according to the zero-mean generalised Gaussian distribution $\mathcal{GG}(s,p)$ with density:
    \begin{align}
        \mathcal{GG}(x\vert s,p)=\frac{p}{2s \Gamma\left(\frac{1}{p}\right)} \exp\left(-\left\lvert\frac{x}{s}\right\rvert^p\right),\label{eq:ggd}
    \end{align}
    where $0<s<\infty$ is the scale parameter, $p>0$ the shape parameter and $\Gamma(\cdot)$ is the gamma function.
\end{assumption}
We observe common distributions emerge from the set of GGDs including the Laplace for $p=1$, the Gaussian for $p=2$ and the uniform over $[-s,+s]$ in the limit $p\rightarrow\infty$.

If we make the assumption that all elements of $E$ are independent (this is true as $r$ grows) then we can write $p(E)\approx \hat{p}(E)\coloneqq\prod_{i=1}^{m}\prod_{j=1}^np(E_{i,j})$ as the product of the marginal distributions. Under this approximation, the score function can be defined element-wise as:
\begin{align}
    [\nabla_E \log p(E)]_{i,j}\approx \hat{S}(E_{i,j})\coloneqq\partial_{E_{i,j}}
    \log p(E_{i,j}).
\end{align}
Using this approximation we apply the score function $\hat{S}(\cdot)$ element-wise to the  matrix $E$:
\begin{align}
      g_\textrm{LR}&\approx\hat{g}_\textrm{MF}\coloneqq-\frac{1}{\sigma}\mathbb{E}_{E\sim p(E)}\left[   f(W=M+\sigma E ) \cdot \hat{S}\odot(E)\right].
\end{align}
For $r=1$, $\hat{S}(\cdot)$ has a convenient analytic form for all members of the set of GGDs:
\begin{theorem}
 Let Assumption~\ref{ass:ggd} hold  and $r=1$. Then the distribution over marginals $p(E_{i,j})$ is:
 \begin{align}
     p(E_{i,j})= \frac{p}{\left(s \Gamma\left(\frac{1}{p}\right)\right)^2}K_0\left(\frac{2\lvert E_{i,j}\rvert^\frac{p}{2}}{s^p}\right),\label{eq:product_density}
 \end{align}
 where $K_0\left(\cdot\right)$ is the zeroth-order modified Bessel function of the second kind and the marginal score function is defined element-wise as:
 \begin{align}
    \hat{S}(E_{i,j}) =-\frac{  K_1\left(\frac{2\lvert E_{i,j}\rvert^\frac{p}{2}}{s^p}\right) }{K_0\left(\frac{2\lvert E_{i,j}\rvert^\frac{p}{2}}{s^p}\right)}\cdot \frac{p\lvert E_{i,j}\rvert^{\frac{p}{2}-1}\textnormal{\textrm{sign}}(E_{i,j})}{s^p}.
 \end{align}
  \begin{proof}
     For $r=1$, we denote the elements of vector $A$ as $a_i$  and elements of vector $B$ as $b_j$, then the elements of matrix $E=A B^\top$  are: $E_{i,j}=a_i b_j$. We now derive the distribution of the unnormalised variables: $E_{i,j}$ using the formula for the distribution of the product of two independent random variables \citep{Rohatgi76,Grimmett93}:
     \begin{align}
         p(E_{i,j})=&\int_{-\infty}^\infty  p(a_i) p\left(b_j=\frac{E_{i,j}}{a_i}\right)\frac{1}{\lvert a_i \rvert} da_i,\\
         =&\left(\frac{p}{2s \Gamma\left(\frac{1}{p}\right)}\right)^2\int_{-\infty}^\infty  \exp\left(-\left\lvert\frac{a_i}{s}\right\rvert^p\right) \exp\left(-\left\lvert\frac{E_{i,j}}{a_is}\right\rvert^p\right)\frac{1}{\lvert a_i\rvert}da_i,\\
         =&2\left(\frac{p}{2s \Gamma\left(\frac{1}{p}\right)}\right)^2\int_0^\infty  \exp\left(-\left\lvert\frac{a_i}{s}\right\rvert^p\right) \exp\left(-\left\lvert\frac{E_{i,j}}{a_is}\right\rvert^p\right)\frac{1}{\lvert a_i\rvert}da_i,
     \end{align}
     where we have used symmetry of the integrand about $0$ to derive the final line. Now, making the substitution $x=\left( \frac{a_i}{s}\right)^p$, we have:
     \begin{align}
         \frac{da_i}{dx}=\frac{s{x}^{\frac{1}{p}-1}}{p},\quad a_i=sx^{\frac{1}{p}}
     \end{align}
     hence: 
     \begin{align}
         p(E_{i,j})=&\frac{p}{\left(s \Gamma\left(\frac{1}{p}\right)\right)^2}\frac{1}{2}\int_0^\infty  \exp\left(-x-\frac{1}{x}\frac{\left\lvert E_{i,j}\right\rvert^p}{s^{2p}}\right)\frac{1}{ x}dx.
     \end{align}
     Now, we use the identity \citep[Theorem 9.42]{Temme96}:
     \begin{align}
         K_0(z)=\frac{1}{2}\int_0^\infty \exp\left(-x-\frac{z^2}{4x}\right)\frac{1}{x} dx,
     \end{align}
     with $z=\frac{2\lvert E_{i,j}\rvert^\frac{p}{2}}{s^p}$ to yield:
     \begin{align}
p(E_{i,j})=&\frac{p}{\left(s \Gamma\left(\frac{1}{p}\right)\right)^2}K_0\left(\frac{2\lvert E_{i,j}\rvert^\frac{p}{2}}{s^p}\right),
     \end{align}
as required for \cref{eq:product_density}. Now we derive the marginal score function by applying the chain rule:

\begin{align}
    \partial_{E_{i,j}} \log p(E_{i,j})&=\partial_{E_{i,j}} \log K_0\left(\frac{2\lvert E_{i,j}\rvert^\frac{p}{2}}{s^p}\right),\\
    &=\partial_{z} \log K_0\left(z=\frac{2\lvert E_{i,j}\rvert^\frac{p}{2}}{s^p}\right) \partial_{E_{i,j}} \frac{2\lvert E_{i,j}\rvert^\frac{p}{2}}{s^p},\\
    &=\partial_{z} \log K_0\left(z=\frac{2\lvert E_{i,j}\rvert^\frac{p}{2}}{s^p}\right)  \frac{p\lvert E_{i,j}\rvert^{\frac{p}{2}-1}\textrm{sign}(E_{i,j})}{s^p},\\
    &=\frac{\partial_{z}  K_0\left(z=\frac{2\lvert E_{i,j}\rvert^\frac{p}{2}}{s^p}\right) }{K_0\left(z=\frac{2\lvert E_{i,j}\rvert^\frac{p}{2}}{s^p}\right)}\cdot \frac{p\lvert E_{i,j}\rvert^{\frac{p}{2}-1}\textrm{sign}(E_{i,j})}{s^p},\\
    &=-\frac{  K_1\left(\frac{2\lvert E_{i,j}\rvert^\frac{p}{2}}{s^p}\right) }{K_0\left(\frac{2\lvert E_{i,j}\rvert^\frac{p}{2}}{s^p}\right)}\cdot \frac{p\lvert E_{i,j}\rvert^{\frac{p}{2}-1}\textrm{sign}(E_{i,j})}{s^p},
\end{align}
where we have used the identity $\partial_z K_0(x)=-K_1(x)$ from \cref{eq:bessel_derivative}.
 \end{proof}
\end{theorem}
For $r>1$ we can derive $\hat{S}(\cdot)$ for the Gaussian sampling case:
\begin{theorem}\label{proof:marginal_score_gauss}
    Let Assumption~\ref{ass:ggd} hold and $p=2$. Then the distribution over marginals $p(E_{i,j})$ is:
    \begin{align}
        p(E_{i,j})=\frac{2\sqrt{r}\lvert \sqrt{r}E_{i,j}\rvert^\frac{r-1}{2}}{s^{r+1}\sqrt{\pi }\Gamma\left(\frac{r}{2}\right)}\cdot K_{\frac{r-1}{2}}\left(\frac{2\lvert \sqrt{r}E_{i,j} \rvert }{s^2}\right).
    \end{align}
    and the score function is (for $E_{i,j}\ne0$): 
    \begin{align}
        \hat{S}(E_{i,j})&=\frac{r-1}{ E_{i,j}}-\frac{2\sqrt{r}\textnormal{\textrm{sign}}(E_{i,j})}{s^2} \frac{ K_{\frac{r+1}{2}}\left(\frac{2\lvert \sqrt{r}E_{i,j} \rvert }{s^2}\right)}{K_{\frac{r-1}{2}}\left(\frac{2\lvert \sqrt{r}E_{i,j} \rvert }{s^2}\right)}.
    \end{align}
  \begin{proof}
 Each element $E_{i,j}$ is the sum of $r$ independent variables $u_{i,j,l}\coloneqq a_{i,l}b_{j,l}$ distributed according to \cref{eq:product_density} with $p=2$:
\begin{align}
    E_{i,j}=\frac{1}{\sqrt{r}}\sum_{l=1}^r a_{i,l}b_{j,l}=\frac{1}{\sqrt{r}}\sum_{l=1}^r u_{i,j,l}.
\end{align}
Let $Z_{i,j} =\sqrt{r}E_{i,j}$, hence:
\begin{align}
    Z_{i,j}=\sum_{l=1}^r u_{i,j,l}.
\end{align}
We first find the density $p(Z_{i,j})$. From \cref{eq:product_density}, the distribution of each $u_{i,j,l}$ is: 
\begin{align}
    p(u_{i,j,l})=\frac{2}{s^2\pi } K_0\left(\frac{2\lvert u_{i,j,l} \rvert }{s^2}\right)
\end{align}
 We use the fact that the PDF of a sum of $r$ independent random variables (i.e. $Z_{i,j}$) is given by the $r$-fold convolution of the individual PDFs. As convolution in the spatial domain is equal to multiplication in the frequency domain, the PDF $p(Z_{i,j})$ follows by taking Fourier transform of $p(u_{i,j,l})$, taking the power $r$ and then taking the inverse Fourier transform: 
\begin{align}
    p(Z_{i,j})=\left(\frac{2}{s^2\pi}\right)^r \mathcal{F}^{-1}\left[ \mathcal{F}\left[K_0\left(\frac{2\lvert \cdot\rvert }{s^2} \right)\right]^r\right](Z_{i,j}),
\end{align}
where recall from \cref{sec:notation} with $d=1$,  $\mathcal{F}[f](\omega)\coloneqq \int f(x) \exp(-i\omega x)dx$ denotes the Fourier transform and $\mathcal{F}^{-1}[\tilde{f}](x)\coloneqq \frac{1}{2\pi} \int \tilde{f}(\omega) \exp(i\omega x)d\omega,$ the inverse Fourier transform. Taking the Fourier transform of the Bessel function: 
        \begin{align}
            \mathcal{F}\left[K_0\left(\frac{2\lvert \cdot\rvert }{s^2} \right)\right](\omega)&=\int \exp(-i\omega x) K_0\left(\frac{2\lvert x\rvert }{s^2} \right) dx,\\
            &=\int \cos(\omega x) K_0\left(\frac{2 \lvert x\rvert }{s^2} \right) dx-i\int \sin(\omega x ) K_0\left(\frac{2 \lvert x\rvert }{s^2} \right) dx,\\
            &=\int \cos(\omega x) K_0\left(\frac{2 \lvert x\rvert }{s^2} \right) dx,\\
            &=2\int^\infty_0 \cos(\omega x) K_0\left(\frac{2 x }{s^2} \right) dx,\label{eq:cos_integral}
        \end{align}
        where we have used the fact that $K_0\left(\frac{2\lvert x\rvert }{s^2}\right)$ is an even function of $x$ and so its integral with $\sin(\omega x)$ in the second line is zero. Using a standard result, we can evaluate the integral in \cref{eq:cos_integral}  \citet[6.671 Integral 14]{Gradshtein15}:
        \begin{align}
           \mathcal{F}\left[K_0\left(\frac{2\lvert \cdot\rvert }{s^2} \right)\right](\omega)=\frac{\pi}{\sqrt{\omega^2+\left(\frac{2}{s^2}\right)^2}}, 
        \end{align}
        hence: 
        \begin{align}
            p(Z_{i,j})&=\left(\frac{2}{s^2\pi}\right)^r \mathcal{F}^{-1}\left[ \frac{\pi^r}{\left(\omega^2+\left(\frac{2}{s^2}\right)^2\right)^\frac{r}{2}}\right](Z_{i,j}),\\
            &=\left(\frac{2}{s^2}\right)^r \mathcal{F}^{-1}\left[ \left(\omega^2+\left(\frac{2}{s^2}\right)^2\right)^{-\frac{r}{2}}\right](Z_{i,j}),\label{eq:fourier-inverse_fourier}\\
            &=\left(\frac{2}{s^2}\right)^r \frac{1}{2\pi}\int \exp(i\omega  Z_{i,j}) \left(\omega^2+\left(\frac{2}{s^2}\right)^2\right)^{-\frac{r}{2}}d\omega,\\
             &=\left(\frac{2}{s^2}\right)^r \frac{1}{2\pi}\Bigg(\int \cos(\omega  Z_{i,j}) \left(\omega^2+\left(\frac{2}{s^2}\right)^2\right)^{-\frac{r}{2}}d\omega\\
             &\qquad+i\int \sin(\omega  Z_{i,j}) \left(\omega^2+\left(\frac{2}{s^2}\right)^2\right)^{-\frac{r}{2}}d\omega\Bigg),\\
             &=\left(\frac{2}{s^2}\right)^r \frac{1}{2\pi}\int \cos(\omega  Z_{i,j}) \left(\omega^2+\left(\frac{2}{s^2}\right)^2\right)^{-\frac{r}{2}}d\omega,\\
             &=\left(\frac{2}{s^2}\right)^r \frac{1}{\pi}\int^\infty_0 \cos(\omega  Z_{i,j}) \left(\omega^2+\left(\frac{2}{s^2}\right)^2\right)^{-\frac{r}{2}}d\omega,\label{eq:inverse_integral}
        \end{align}
        where we have used the fact that the integrand is an even function and so its integral with $\sin(\omega Z_{i,j})$ is zero to derive the  penultimate line. To evaluate the integral in \cref{eq:inverse_integral} we apply \citet[3.771 Integral 2]{Gradshtein15}:
        \begin{align}
             p(Z_{i,j})&=\left(\frac{2}{s^2}\right)^r \cdot\frac{1}{\sqrt{\pi}\Gamma\left(\frac{r}{2}\right)}\left(\frac{s^2\lvert Z_{i,j}\rvert  }{4}\right)^{\frac{r-1}{2}} \cdot K_{\frac{r-1}{2}}\left(\frac{2\lvert Z_{i,j} \rvert }{s^2}\right),\\
             &=\frac{2\lvert Z_{i,j}\rvert^\frac{r-1}{2}}{s^{r+1}\sqrt{\pi }\Gamma\left(\frac{r}{2}\right)}\cdot K_{\frac{r-1}{2}}\left(\frac{2\lvert Z_{i,j} \rvert }{s^2}\right).
        \end{align}
        Using the transformation of variables $E_{i,j}=\frac{1}{\sqrt{r}} Z_{i,j}$ yields our desired results:
        \begin{align}
            p(E_{i,j})&=\sqrt{r}p(Z_{i,j}=\sqrt{r}E_{i,j}),\\
            &=\frac{2\sqrt{r}\lvert \sqrt{r}E_{i,j}\rvert^\frac{r-1}{2}}{s^{r+1}\sqrt{\pi }\Gamma\left(\frac{r}{2}\right)}\cdot K_{\frac{r-1}{2}}\left(\frac{2\lvert \sqrt{r}E_{i,j} \rvert }{s^2}\right).
        \end{align}
        
        Now, we derive the score function:
        \begin{align}
            \partial_{E_{i,j}} \log p(E_{i,j})&= \frac{r-1}{2 }\cdot\partial_{E_{i,j}}\log \lvert \sqrt{r}E_{i,j}\rvert+\partial_{E_{i,j}}\log K_{\frac{r-1}{2}}\left(\frac{2\lvert \sqrt{r}E_{i,j} \rvert }{s^2}\right),\\
            &= \frac{r-1}{2 E_{i,j}}+\frac{2\partial_{E_{i,j}}\lvert \sqrt{r}E_{i,j} \rvert }{s^2} \frac{\partial_x K_{\frac{r-1}{2}}\left(x=\frac{2\lvert \sqrt{r}E_{i,j} \rvert }{s^2}\right)}{K_{\frac{r-1}{2}}\left(\frac{2\lvert \sqrt{r}E_{i,j} \rvert }{s^2}\right)},\\
            &= \frac{r-1}{2 E_{i,j}}+\frac{2\sqrt{r}\textrm{sign}(E_{i,j})}{s^2} \frac{\partial_x K_{\frac{r-1}{2}}\left(x=\frac{2\lvert \sqrt{r}E_{i,j} \rvert }{s^2}\right)}{K_{\frac{r-1}{2}}\left(\frac{2\lvert \sqrt{r}E_{i,j} \rvert }{s^2}\right)},
        \end{align}
        Now, from \cref{eq:derivative_general} for $E_{i,j}\ne0$:
        \begin{align}
            \frac{\partial_x K_{\frac{r-1}{2}}\left(x\right)}{K_{\frac{r-1}{2}}\left(x\right)}&=\frac{\frac{r-1}{2 x} K_{\frac{r-1}{2}}\left(x\right)-K_{\frac{r+1}{2}}\left(x\right)}{K_{\frac{r-1}{2}}\left(x\right)},\\
            &=\frac{r-1}{2 x}-\frac{K_{\frac{r+1}{2}}\left(x\right)}{K_{\frac{r-1}{2}}\left(x\right)},\\
            \implies   \partial_{E_{i,j}} \log p(E_{i,j})&= \frac{r-1}{2 E_{i,j}}+\frac{(r-1)\textrm{sign}(E_{i,j})}{2\lvert E_{i,j}\rvert}-\frac{2\sqrt{r}\textrm{sign}(E_{i,j})}{s^2} \frac{ K_{\frac{r+1}{2}}\left(\frac{2\lvert \sqrt{r}E_{i,j} \rvert }{s^2}\right)}{K_{\frac{r-1}{2}}\left(\frac{2\lvert \sqrt{r}E_{i,j} \rvert }{s^2}\right)}, \\
            &= \frac{r-1}{2 E_{i,j}}+\frac{(r-1)}{2E_{i,j}}-\frac{2\sqrt{r}\textrm{sign}(E_{i,j})}{s^2} \frac{ K_{\frac{r+1}{2}}\left(\frac{2\lvert \sqrt{r}E_{i,j} \rvert }{s^2}\right)}{K_{\frac{r-1}{2}}\left(\frac{2\lvert \sqrt{r}E_{i,j} \rvert }{s^2}\right)},\\
            &=\frac{r-1}{ E_{i,j}}-\frac{2\sqrt{r}\textrm{sign}(E_{i,j})}{s^2} \frac{ K_{\frac{r+1}{2}}\left(\frac{2\lvert \sqrt{r}E_{i,j} \rvert }{s^2}\right)}{K_{\frac{r-1}{2}}\left(\frac{2\lvert \sqrt{r}E_{i,j} \rvert }{s^2}\right)},
        \end{align}
         as required.
    \end{proof}
\end{theorem}

\newpage

\newpage

\section{EGGROLL Speed}
\label{sec:speed_appendix}


All timings were done on a single GPU on a GH200 (equivalent to a single H100) for a linear model with dimension 8192 in bfloat16, allowing a maximum batch size of 1024. For the graph in~\cref{fig:batch_timings}, we pre-generate the noises instead of integrating the noise generation into the forward pass.

\begin{figure}[htb!]
    \centering
    \includegraphics[width=0.4\linewidth]{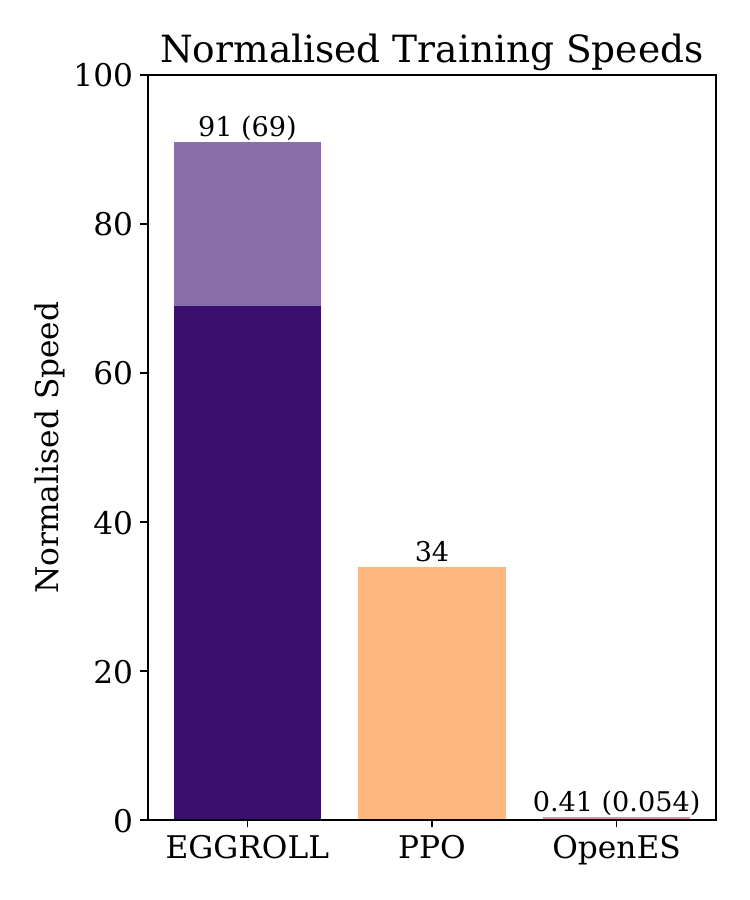}
    \caption{Relative speed of EGGROLL, when including jax noise regeneration.}
    \label{fig:eggroll_speed_full}
    
\end{figure}

In~\cref{fig:eggroll_speed_full}, we consider the impact of regenerating noises on-the-fly using jax PRNG. The darker area and value in parenthesis for EGGROLL and OpenES indicate the speed when regenerating noises on-the-fly, while the full bar indicates the speed when the noises are already generated. 

We regenerate noises on the fly in our primary jax codebase, but pre-generating the EGGROLL perturbations beforehand is also a practical possibility since low-rank perturbations only require a small amount of memory, proportional to the square root of the size of the original parameter matrices.

\newpage

\section{Arithmetic Intensity Analysis} \label{sec:arithmetic_intensity_appendix}

In this section, we derive the arithmetic intensity of standard batched inference, Gaussian matrix ES, and EGGROLL. We calculate arithmetic intensity as the number of operations divided by the total number of bytes read from or written to. For context, for the (b)float16 datatype on an H100 GPU, there are approximately 1000 teraFLOPS of compute (without sparsity) and 3.35 TB/s of GPU memory bandwidth, meaning that the roofline threshold is approximately 300 ops/byte, defined as the minimum for computation needed for it to be the bottleneck instead of memory movement.

In the following subsections, we are considering a single linear layer with mean parameter $M \in \mathbb{R}^{d_{out} \times d_{in}}$ and a batch of inputs $u \in \mathbb{R}^{B \times d_{in}}$. All operations occur with a precision of $s$ bytes per element.

\subsection{Arithmetic Intensity of Standard Batched Inference}

In standard batched inference, we wish to simply calculate $uM^T$. The total bytes read as input are $B \times d_{in} \times s$ (for $u$) and $d_{out} \times d_{in} \times s$ (for $M$), and the total bytes written as output are $B \times d_{out} \times s$. The total number of operations are $B \times d_{in} \times d_{out} \times 2$ since matrix multiplication requires both multiplications and additions for each element of $u$ across all of $d_{out}$. Therefore, the arithmetic intensity is: $$\frac{B \times d_{in} \times d_{out} \times 2}{B \times d_{in} \times s + B \times d_{out} \times s + d_{out} \times d_{in} \times s}.$$

When $s=2$ (for (b)float16) and $d_{out}=d_{in}=m$, the arithmetic intensity simplifies to $$\frac{Bm}{2B + m}.$$

The batch size needed to achieve a desired arithmetic intensity of $A$ is derived as follows:
\begin{align}
    Bm &= 2AB + Am\\
    Bm-2AB &= Am\\
    B &= \frac{Am}{m-2A}\\
\end{align}
Therefore, achieving an arithmetic intensity of 300 ops/byte with $m=8192$ requires a minimum batch size of 324.

\subsection{Arithmetic Intensity of Gaussian Matrix ES}

In Gaussian matrix ES, we assume access to pre-generated perturbations of shape $\mathbb{R}^{B \times d_{out} \times d_{in}}$. The total bytes read as input are $B \times d_{in} \times s$ (for $u$) and $B \times d_{out} \times d_{in} \times s$ (for $M$), and the total bytes written as output are $B \times d_{out} \times s$. Otherwise, the total number of operations is identical to standard batched inference, giving us an arithmetic intensity of $$\frac{B \times d_{in} \times d_{out} \times 2}{B \times d_{in} \times s + B \times d_{out} \times s + B\times d_{out} \times d_{in} \times s} = \frac{d_{in} \times d_{out} \times 2}{d_{in} \times s + d_{out} \times s + d_{out} \times d_{in} \times s}.$$

When $s=2$ (for (b)float16) and $d_{out}=d_{in}=m$, the arithmetic intensity simplifies to $$\frac{m}{2 + m}.$$

This means that arithmetic intensity is always strictly less than 1, regardless of batch size or dimensionality. The common way to increase arithmetic intensity is to bring it closer to standard batched inference, reusing the same perturbation across multiple inputs. For instance, when $m=8192$, achieving an arithmetic intensity of 300 ops/byte requires that each perturbation is reused at least 324 times, and smaller values of $m$ need to be reused even more often.

\subsection{Arithmetic Intensity of EGGROLL}

For EGGROLL, we assume access to the pre-generated decomposed perturbations $A \in \mathbb{R}^{B \times d_{out} \times r}$ and $B \in \mathbb{R}^{B \times d_{in} \times r}$. Therefore, the bytes read as pure input are $B\times d_{in} \times s + B \times (d_{in} + d_{out}) \times r \times s + d_{out} \times d_{in} \times s$ and the bytes written as pure output are $B \times d_{out} \times s$. However, the efficient low-rank perturbation calculation requires writing and reading an intermediate matrix of shape $B \times r$, so the total bytes read are $$(B\times d_{in} + B \times (d_{in} + d_{out}+2) \times r + d_{out} \times d_{in} + B \times d_{out}) \times s.$$

The total number of operations includes the amount for standard batch inference, $B \times d_{in} \times d_{out} \times 2$, along with the rank-$r$ perturbations, $B \times (d_{in} + d_{out}) \times r \times 2$, and the final sum between the main calculation and perturbation $B \times d_{out}$. Therefore, the arithmetic intensity is $$\frac{B \times d_{in} \times d_{out} \times 2 + B \times (d_{in} + d_{out}) \times r \times 2 + B \times d_{out}}{(B\times d_{in} + B \times (d_{in} + d_{out}+2) \times r + d_{out} \times d_{in} + B \times d_{out}) \times s}.$$

When $s=2$ (for (b)float16) and $d_{out}=d_{in}=m$, the arithmetic intensity simplifies to
\begin{align}
    &\frac{Bm + 2Br + \frac{B}{2}}{B + B r(2+\frac{2}{m}) + m + B}\\
    &= \frac{m + 2r + \frac{1}{2}}{2 + r(2+\frac{2}{m}) + \frac{m}{B}.}
\end{align}

The batch size needed to achieve a desired arithmetic intensity of $A$ is derived as follows:
\begin{align}
    2A + rA(2+\frac{2}{m}) + \frac{Am}{B} &= m + 2r + \frac{1}{2}\\
    \frac{Am}{B} &= m + 2r + \frac{1}{2} - 2A - rA(2+\frac{2}{m})\\
    B &= \frac{Am}{m - 2A + 2r + \frac{1}{2} - rA(2+\frac{2}{m})}
\end{align}

Note that the only difference with the critical batch size of standard batched inference is the additional $2r + \frac{1}{2} -rA(2+\frac{2}{m})$ in the denominator. Therefore, achieving an arithmetic intensity of 300 ops/byte with $m=8192$ and $r=1$ requires a minimum batch size of 352, compared to 324 for standard batched inference. This means that EGGROLL can saturate compute with unique perturbations per input, unlike Gaussian matrix ES.

Note that there is an overhead of $Bm(4r+1)$ flops relative to standard batched inference, resulting in an additional compute rate of $\frac{Bm(4r+1)}{2Bm^2} = \frac{4r+1}{2m}$, which is effectively negligible for large enough matrices.

\newpage

\section{EGG Architecture} \label{sec:egg_appendix}

In the following section, we detail the design of our EGG model, which follows the high-level structure of modern pre-layernorm decoder-only language models, but replaces self-attention with a modified minGRU and standard layernorms with a custom variant to enable pure integer training. See~\cref{alg:egg} for an overview of the forward pass of the EGG architecture.

\begin{algorithm}
\caption{EGG forward pass}  \label{alg:egg}
\begin{algorithmic}
\REQUIRE Input token $t \in \mathbb{U}_8$, input state $s \in \mathbb{I}^{l\times D}_8$, network parameters $\theta$
\ENSURE Output vector $y \in \mathbb{I}^{D}_8$ and output state $s' \in \mathbb{I}^{l\times D}_8$
\STATE $s' \gets \mathbb{I}^{l\times D}_{8}$ initialised to 0
\STATE $y \gets \text{EMBED}(\theta_\text{emb}, t)$
\FOR{$i \in \{0, \dots, l-1\}$}
    \STATE $y', s'_i \gets \text{GRU}(\theta_{\text{gru},i}, \text{LN}(\theta_{\text{ln1},i},y), s_i)$
    \STATE $y \gets \mathbb{I}_8(\mathbb{I}_{32}(y')+\mathbb{I}_{32}(y))$
    
    \STATE $y' \gets \text{MLP}(\theta_{\text{mlp},i}, \text{LN}(\theta_{\text{ln2},i},y))$
    \STATE $y \gets \mathbb{I}_8(\mathbb{I}_{32}(y')+\mathbb{I}_{32}(y))$
\ENDFOR
\STATE $y \gets \text{LN}(\theta_{\text{lnout},i},y) @ \theta_\text{head}^T$
\end{algorithmic}
\end{algorithm}

\subsection{Motivation}

Since EGGROLL does not rely on gradients, we can explicitly design a language model architecture to be efficient and hardware-friendly at inference time. In particular, we design EGG under the following constraints to emphasise the flexibility of EGGROLL:

\paragraph{Pure Integer Training:} On H100 systems, int8 is the fastest datatype and int8 matrix multiplication with int32 accumulation is the fastest tensor core operation. Furthermore, integer datatypes are much simpler to implement in hardware, providing massive energy savings for high-throughput systems~\citep{6757323}. Therefore, we keep all weights in int8 and all activations in integer formats, \textit{never} casting to floating point at any point during training. This stands in contrast to the standard approach for language model quantisation through "quantisation aware training" with backpropagation, where floating point activations are still necessary~\citep{wang2023bitnetscaling1bittransformers}.

\paragraph{Nonlinear RNN:} Modern language models use sequence-parallel architectures like Transformers and SSMs, since they enable stable gradients without backpropagation through time. However, most of these architectures cannot handle simple state tracking~\citep{pmlr-v235-merrill24a}, whereas classic recurrent networks like LSTMs and GRUs can do so with a single layer. Since EGGROLL does not require backpropagation through time, we can train on unbounded sequence lengths~\citep{li2023variance} with nonlinear RNNs of broader complexity classes. Specifically, we develop a variant of the minGRU model~\citep{heck2017simplifiedminimalgatedunit} that performs all operations in integer formats.
    
\paragraph{Removal of all Activation Functions:} Inspired by~\citet{foerster2017nonlinear}, we remove all activation functions, like the rectified linear unit and hyperbolic tangent, due to the nonlinearity present in the int8 datatype. Specifically, the saturated addition of int8 values provides sufficient nonlinearity due to the implicit clipping of values to the int8 dynamic range, which evolution strategies can exploit.

\subsection{Notation and Operations}

We use the constant $l \in \mathbb{Z}^+$ to denote the number of layers of the model and $D=4^d$ as the hidden dimension of the model, where $d \in \mathbb{Z}^+$.

We use $\mathbb{I}_n$ to denote an $n$-bit signed integer and $\mathbb{U}_n$ to denote an $n$-bit unsigned integer. We denote casting vector $\vec{u}$ to format $\mathbb{I}_n$ as $\mathbb{I}_n(\vec{u})$, which implicitly includes clipping to the bounds of the datatype. To ensure symmetry between positive and negative values of each datatype, we consider the value $-2^{n-1}$ to be invalid for datatype $\mathbb{I}_n$; for instance, for 8-bit signed integers we only allows value from -127 to 127.

We use the following operations:
\begin{itemize}
    \item $\vec{u} @ M$ indicating scaled vector-matrix multiplication of $\mathbb{I}^n_8\times \mathbb{I}^{n,m}_8 \rightarrow \mathbb{I}^{m}_{8}$, corresponding to int8 tensor core multiplication with int32 accumulation and scaling. The details of this operation are described in~\cref{app:egg:matmul}.
    \item $a \cdot b$ indicates dot product with int32 accumulation, $\mathbb{I}_8^n \times \mathbb{I}_8^n \rightarrow \mathbb{I}_{32}$, and $a \odot b$ indicates the Hadamard (elementwise) product.
    \item Standard integer operations: $+$ for addition, $-$ for subtraction, and $\odot$ for element-wise multiplication.
    \item $|u|$ indicates taking the element-wise absolute value of $u$, $\mathbb{I}^n \rightarrow \mathbb{I}^n$.
    \item $\text{sign}(u)$ indicates taking the element-wise sign of $u$, giving 1 for positive values, -1 for negative values, and 0 for zero.
    \item $\text{sum}(u)$ indicates taking the sum of all elements in $u$ (casting to $\mathbb{I}_{32}$ to prevent overflow): $\mathbb{I}^n \rightarrow \mathbb{I}_{32}$.
    \item $u \gg n$ indicates an elementwise bitwise right shift by $n$, which is typically equivalent to $2^{-n}u$. Similarly, $u \ll n$ indicates a bitwise left shift by $n$, which is typically equivalent to $2^nu$.
    \item Square-bracket indexing. For instance $M[i,j]$ extracts the element at index $i$ in axis 0 and index $j$ in axis 1, following the zero-based indexing convention.
\end{itemize}

\subsection{Parameter Initialisation}

The standard initialisation for matrix parameters in our model is rounding 16 times a sample from the standard normal, and casting to $\mathbb{I}_8$. This can be precomputed on a CPU since this is only done once at the start of training.

The egg model has the following parameters (where an additional subscript of $i$ indicates that there is a version of this parameter for each layer of the model):
\begin{itemize}
    \item $\theta_\text{emb} \in \mathbb{I}^{256\times D}_8$, following standard initialisation.
    \item $\theta_\text{head} \in \mathbb{I}^{256\times D}_8$, following standard initialisation.
    \item $\theta_\text{lnout} \in \mathbb{I}_8^D$, initialised to 16 for each element.
    \item $\theta_{\text{ln1},i}, \theta_{\text{ln2},i} \in \mathbb{I}_8^D$, initialised to 16 for each element
    \item $\theta_{\text{mlp},i,1} \in \mathbb{I}_8^{4D \times D}$ and $\theta_{\text{mlp},i,2} \in \mathbb{I}_8^{D \times 4D}$, following standard initialisation.
    \item $\theta_{\text{GRU},i,[\text{Wf,Uf,Wh,Uh}]} \in \mathbb{I}_8^{D \times D}$, following standard initialisation.
    \item $\theta_{\text{GRU},i,[\text{bfm bh}]} \in \mathbb{I}_8^D$, initialised to 0 for each element.
\end{itemize}

In total there are $513D+l(4D+12D^2)$ parameters in the model.

\subsection{Matrix Multiplication}
\label{app:egg:matmul}

Tensor cores in GPUs are able to calculate fast vector-matrix multiplications with int32 accumulation as $uM \in \mathbb{I}_{32}^m$ where $u \in \mathbb{I}_8^n$ and $M \in \mathbb{I}_8^{n \times m}$. For our purposes, we define $u@M$ as a scaled multiplication:

$$ u @ M := \mathbb{I}_8\left(\frac{uM}{16\sqrt{n}}\right).  $$

Note that when $n=4^d$, the division operation just becomes a right-shift by $4+d$, which is fast to calculate.

We choose this scaled matrix multiplication because we initialise $M$ to 16 times standard normal samples for each element, so dividing by $16\sqrt{n}$ preserves the magnitude of $u$ for the output. In particular, if all elements of $u$ and $M$ are drawn from independently from the standard normal distribution multiplied by 16, the central limit theorem tells us that the expected value per element of the output will be $256\sqrt{n}$, so dividing by $16\sqrt{n}$ preserves the standard deviation of 16.

\subsection{Embedding}

Our embedding function takes as input an embedding matrix $\theta_\text{emb} \in \mathbb{I}^{256\times D}_8$ and an input token $t \in \mathbb{U}_8$, and simply outputs the vector corresponding to that token: $\theta_\text{emb}[t] \in \mathbb{I}^D_8$.

\subsection{Layer Normalisation (LN)}

Our layer normalisation operation involves multiplying our input $u \in \mathbb{I}^D_8$ with a weight $\theta_\text{ln} \in \mathbb{I}^D_8$ before dividing by the mean absolute value of $u$.

We decide to divide by the mean absolute value of the input instead of the more common root-mean-squared since square roots are expensive on integers. Note that the $L1$ norm after dividing the input by the mean absolute value (when using real numbers) is $D$ instead of 1, which we intentionally choose to preserve more bits of information given the limited range of $\mathbb{I}_8$.

We calculate the mean absolute value of input $u$ as:
$$ u_\text{mav} = \mathbb{I}_8(\text{sum}(|u|) \gg (2d)),$$

Note that we can safely cast the mean absolute value to an $\mathbb{I}_8$ without overflow given the properties of the mean of a set, though we lose precision due to truncating the fractional component.

The output of layernorm is calculated as:
$$ \text{DIVIDE}(\mathbb{I}_{16}(u) \odot \mathbb{I}_{16}(\theta_\text{ln}), u_\text{mav}). $$

Since division is an expensive operation, we precompute it using a lookup table. Note that the product of two $\mathbb{I}_8$ values will always remain in the dynamic range of $\mathbb{I}_{16}$, so our lookup table will be of shape $2^{16}\times 2^8$.

\subsection{MLP}

Each MLP block consists of two weight parameters: $\theta_1 \in \mathbb{I}_8^{4D \times D}$ and $\theta_2 \in \mathbb{I}_8^{D \times 4D}$. Given an input $u \in \mathbb{I}_8^D$, we calculate the output as:
$$ (u @ \theta_1^T) @ \theta_2^T.$$

Note that we do not use an activation function, because the $@$ operation is already nonlinear due to the saturated conversion from $\mathbb{I}_{32}$ to $\mathbb{I}_8$

\subsection{GRU}

Each GRU block accepts an input vector and state $u, s \in \mathbb{I}^D_8$ consists of 6 weight parameters: $\theta_\text{Wf}, \theta_\text{Uf}, \theta_\text{Wh}, \theta_\text{Uh} \in \mathbb{I}_8^{D \times D}$ and $\theta_\text{bf},\theta_\text{bh} \in \mathbb{I}_8^D$.

Using these weight matrices, we calculate the following vectors:
\begin{align}
     f &= \sigma(\mathbb{I}_8(\mathbb{I}_{32}(u@ \theta_\text{Wf}^T) + \mathbb{I}_{32}(s @ \theta_\text{Uf}^T) + \mathbb{I}_{32}(\theta_\text{bf}))),\\
     \hat{f} &= \mathbb{I}_8(((\mathbb{I}_{32}(f) + 127)\odot\mathbb{I}_{32}(s)) \gg 8),\\
     \hat{h} &= \phi(\mathbb{I}_8(\mathbb{I}_{32}(u@ \theta_\text{Wh}^T) + \mathbb{I}_{32}(\hat{f} @ \theta_\text{Uh}^T) + \mathbb{I}_{32}(\theta_\text{bh}))),\\
     h &= s + \mathbb{I}_8(((\mathbb{I}_{32}(f) + 127) \odot (\mathbb{I}_{32}(\hat{h}) - \mathbb{I}_{32}(s))) \gg 8),
\end{align}



where $h$ is the output and the new hidden state. In the typical GRU, $\sigma$ stands for the sigmoid function while $\phi$ stands for the hyperbolic tangent, but we find that setting these as identity operations is sufficient due to the nonlinearity already present in the clipped addition. One can view this clipped addition operation as scaled and shifted version of the ``hard" tanh and sigmoid operators.

To explain why we perform these operations, we can analyse this relative to the original GRU. The $f$ vector for the standard GRU has all elements between 0 and 1 due to the sigmoid, but our elements are between -127 and 127. Therefore, to calculate $\hat{f}$ (which is typically just $f \odot s$), we first add 127 to $f$, getting the range between 0 and 254 before multiplying by $s$ before bit-shifting right by 8 again to bring our values back to the $\mathbb{I}_8$ dynamic range. We apply similar logic to calculate the final $h$, which is typically just $h=s+f\odot(\hat{h}-s)$ but needs to be rescaled to keep the int8 dynamic range.

\subsection{Fitness Calculation in Integer Types}

The ``fitness'' used in language model pretraining is the log-likelihood of correctly generating the next token, treating the outputs of the language model as logits (unnormalised log probabilities). If $t' \in \mathbb{U}_8$ is the next token to predict and $y \in \mathbb{I}^{256}_8$ are the logits, we can calculate the log likelihood as follows:
\begin{align}
    y' &= \mathbb{I}_{32}(y) + 128, \\ 
    o &= y'[t'] - \text{LOG2}[\text{sum}(\text{EXP2}[y'])],
\end{align}
where $o$ is the loss for one token. We implement $\text{EXP2}$ and $\text{LOG2}$ as lookup tables, where
\begin{align}
 \text{EXP2}[i] &= 16\times2^{i/16}, \\
 \text{LOG2}[i] &= 16\times\log_2(i / 16).
\end{align}
Note that each element in EXP2 for any $\mathbb{U}_{8}$ input requires at most 20 bits, so the sum of exponents across all possible choices is at most 28 bits, meaning we have to precompute LOG2 for $2^{28}$ values.

\section{EGG Pretraining with Integer EGGROLL}

The core ideas of EGGROLL still apply in this integer-based training setting, but we have to make some modifications to ensure it only uses integer operations.

\subsection{Adding EGGROLL Perturbations}

For parameter $\theta \in \mathbb{I}_8^{m \times n}$ that represents a matrix multiplication, we first sample rank-1 perturbation vectors for each index in the batch: $A \in \mathbb{I}_8^{m}$ and $B \in \mathbb{I}_8^{n}$. We sample these vectors from the standard random normal multiplied by 16 and rounded to the nearest $\mathbb{I}_8$ (clipping if necessary). To prevent the use of floating-point arithmetic on the accelerator, we pre-generate a large matrix of these random values, randomly indexing into it to get the perturbation vectors.

Given an input $u \in \mathbb{I}_8^{n}$, instead of calculating $u@\theta^T$, we calculate

$$ \mathbb{I}_8\left(\frac{u\theta^T + ((u \cdot B) \mathbb{I}_{32}(A) \gg (4+\hat{\sigma}))}{16\sqrt{n}}\right). $$

The value of $\hat{\sigma}$ is a hyperparameter, related to the $\sigma$ in the main paper as $\sigma=2^{-\hat{\sigma}}$. Note that the batched forward pass remains efficient since it still simply performs a batched vector-vector dot product in int8 (with int32 accumulate) and a batched vector-scalar product in int32.

We apply this same logic to the embedding matrix, since we can interpret $\theta[t]$ as $\text{one\_hot}(t)\theta$ and still apply our rank-1 updates in that context. In practice, this means replacing $u\cdot B$ with $B[t]$.


\subsection{Fitness Shaping}

We employ a simple fitness shaping scheme based on antithetical pairs. Specifically, given raw fitnesses $s^+, s^-$, for the positive and negative sample of the antithetical pair respectively, the transformed fitness for the noise is:
$$ \text{sign}(s^+ - s^-), $$

Note that the only possible values for the fitness after shaping are $\{-1, 0, 1\}$.

\subsection{Parameter Update}

For parameter $\theta \in \mathbb{I}_8^{m \times n}$ that represents a matrix multiplication (or embedding vector), suppose the sampled batch of rank-1 perturbation vectors are $A \in \mathbb{I}_8^{N\times m}$ and $B \in \mathbb{I}_8^{N \times n}$, and let the fitnesses after shaping be $F \in \mathbb{I}_8^{N}$. Then we calculate an intermediate value $E \in \mathbb{I}_{32}^{m \times n}$ as:

$$ E =  (\text{diag}(F)A)^TB .$$

We use $E$ to determine if each element of $\theta$ should be increased or decreased. In particular, when the absolute value of $E$ is above a pre-specified threshold we move $\theta$ by one discrete bin in the direction of the sign of $E$. Since there are only 255 unique values for each element in $\mathbb{I}_8$, restricting updates to single bins improves stability without compromising the ability for a parameter to get to any other value with relatively few updates. In particular, we have a real-valued hyperparameter, $\alpha \in (0, 1)$ such that the threshold equals

$$\mathbb{I}_{32}\left(16\times\Phi^{-1}\left(\frac{1-\alpha}{2}\right)\right)\times 16\sqrt{N},$$

where $\Phi$ is the normal cumulative distribution function. Note that this threshold can be precalculated on a CPU. We observe that $\alpha$ approximately equals the fraction of parameters that are updated at each step.

We currently do not incorporate any momentum or other optimiser states, but this remains critical future work to improve the speed of convergence for pure integer training.

Across model sizes and population size, we find that setting $\hat{\sigma}$ to 4 and letting $\alpha$ decay over training steps as $\frac{1}{.015t + 1}$ gives consistently strong results.

\section{EGG Ablations}
\label{sec:egg_ablation}

In our main experiments, we use a fixed data batch size of 16 sequences for population sizes 2 and powers of 4 ranging from 4 to $4^{10}=1048576$. In this section, we vary the batch size by powers of 4, ranging from 4 to $4^5=1024$, while varying population size by powers of 4 from 16 to 1048576. When the batch size, $b$ is greater than half of the population size, $N$, we give each antithetical pair $\frac{2b}{N}$ sequences, functionally giving a cleaner fitness signal to each member of the population. This also means that the number of parallel "inferences" required is $\max(2b, N)$.

\begin{figure}[htb!]
    \centering
    \includegraphics[width=0.75\linewidth]{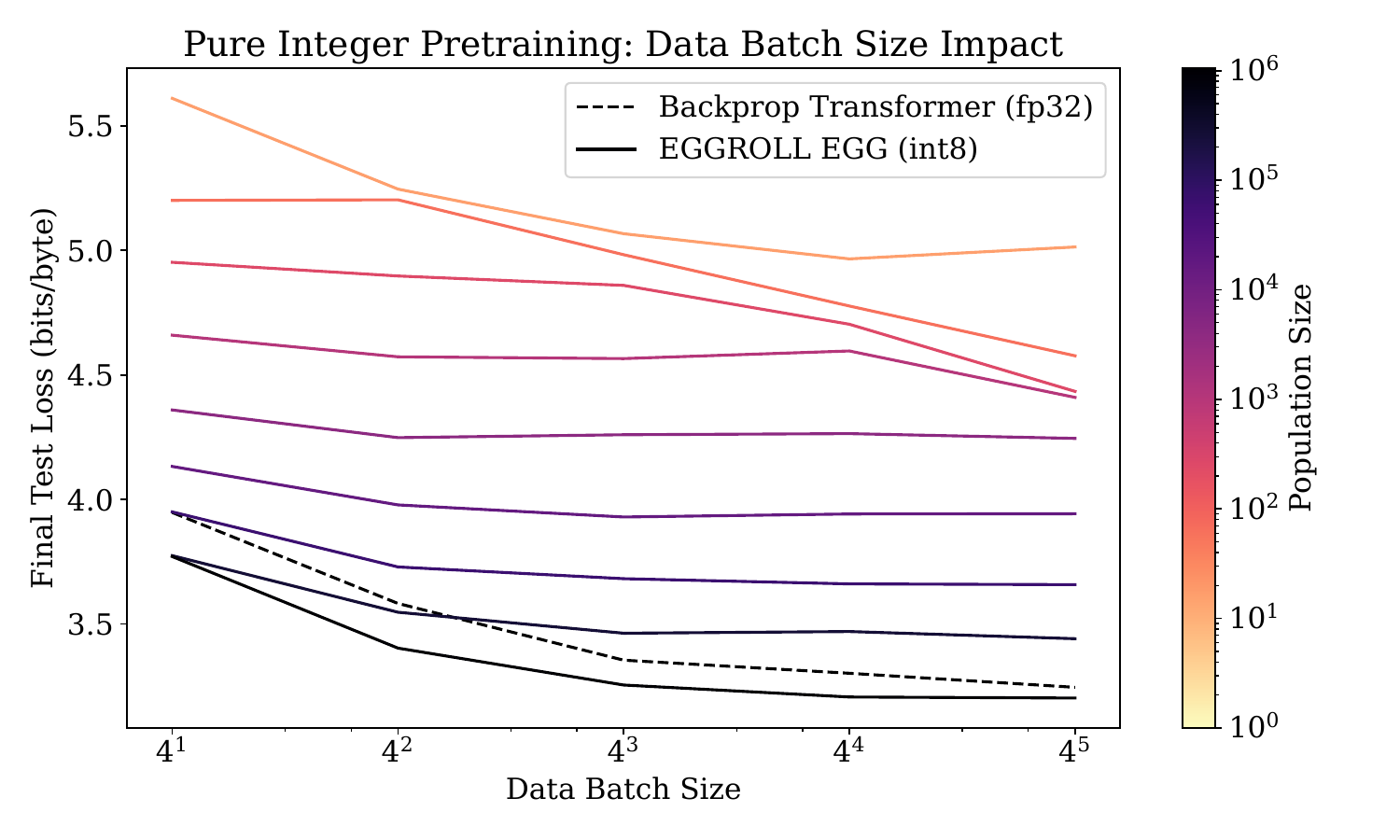}
    \caption{Test loss curves when varying data batch size and population size.}
    \label{fig:data_efficient_pretraining}
\end{figure}

In \cref{fig:data_efficient_pretraining}, we observe that the final test loss for each population size is relatively constant beyond a specific data batch size threshold. At the top right of the figure, we observe a decrease in loss for small population sizes after $b > \frac{N}{2}$, which is an artifact of the increased compute usage necessary to use the full data batch. Ignoring this artifact, the minimum batch size for near-optimal performance at a given population size $N$ appears to be $\frac{N}{4^6}$. We see that large population sizes need larger data batches for improved performance, since a batch size of 4 results in nearly identical performance for population sizes $4^9=262144$ and $4^{10}=1048576$, but this diverges as data batch size increases.

\section{Distributed EGGROLL Framework}
\label{sec:distributed_egg}

To facilitate the large-scale experiments, where we scale population sizes beyond 1M, we develop a lightweight distributed training framework designed to minimise network overhead.

\subsection{Base-3 Fitness Packing and Bandwidth Efficiency} \label{app:network_bandwidth}
A key bottleneck in distributed training is the communication of gradients or results. We address this via a custom base-3 packing scheme for fitness vectors. Since workers evaluate perturbations in antithetic pairs, the raw signal is discretised into ternary values $\{+1, 0, -1\}$. These are mapped to $\{0, 1, 2\}$ and packed five at a time into a single byte:

$$byte = \sum_{i=0}^{4} v_i \cdot 3^i$$

This yields an effective bitrate of $1.6$ bits per value (near the $\log_2 3 \approx 1.585$ theoretical limit). Consequently, the network payload per chunk is approximately $52 + \text{chunk\_size}/10$ bytes, rendering bandwidth usage independent of model size.

\subsection{System Architecture}
The system employs a Coordinator-Worker topology. The Coordinator maintains the global state and assigns population chunks to Workers. Workers calculate fitness on GPU, apply signal shaping (chunk mean filtering, adaptive thresholding), and return only the packed ternary fitness, minimising traffic significantly compared to standard gradient transmission.

\section{Fine-tuning of Integer Quantised Models} \label{sec:finetuning_appendix}
\subsection{Quantisation Procedure}
To maximise population throughput and reduce device memory during EGGROLL fine-tuning, we represent the large matrix-multiplication parameters of RWKV in an int8 weight format while keeping non-matmul parameters (e.g., small biases / bookkeeping tensors) in floating point, bf16. Following \citet{jacob2017quantizationtrainingneuralnetworks}, for each weight matrix $W \in \mathbb{R}^{d_{\text{in}}\times d_{\text{out}}}$, we use symmetric per-channel int8 quantisation with an absmax scale. For each output channel we first compute:
\begin{equation}
    s_i = \max\left(\frac{\max_{j} |W_{i,j}|}{127}, \epsilon\right),
\end{equation}
where $\epsilon$ is some small scalar. Then, we store each $s_i$ in bf16, and quantise weights as
\begin{equation}
    Q_{i,j} = \text{clip}\left(\text{round}\left(\frac{W_{i, j}}{s_i}\right), -127, 127\right) \in \mathbb{I}_8 .
\end{equation}
Every matrix parameter is stored as a dictionary containing the quantised weight matrix $Q$, the scale parameters per channel $\{s_i\} \forall i \in 1, \hdots,  d_{\text{out}}$ and an input scale factor $s_x$ in bf16 precision. At runtime, the forward pass is computed by scaling the input vector by $s_x$ and the quantised matrix $Q$ with the scales per channel, $[s_1, \hdots, s_{d_{\text{out}}}]$,
\begin{equation}
    x_{n+1} = (x_n \odot s_x)^T (W \odot [s_1, \hdots, s_{d_{\text{out}}}]).
\end{equation}

\subsection{Integrating integer-quantised EGGROLL with Adam}
\label{sec:quant_eggroll_adam}

EGGROLL performs black-box (ES) optimisation directly over the parameter representation used in the forward pass, including integer quantised weights.
We integrate this with the Adam optimiser \citep{kingma2014adam} by maintaining Adam's moment estimates in bf16, while enforcing that all quantised tensors remain on the int8 lattice.

\paragraph{ES gradients.}
EGGROLL estimates gradients via antithetic ES perturbations and score-weighted averaging.
This yields a bf16 gradient estimate for:
(i) floating-point parameters (when present),
(ii) quantised matrix parameters via a low-rank perturbation pathway,
and (iii) scale parameters $\{s_i\} \forall i \in 1, \hdots,  d_{\text{out}}$ and $s_x$ via explicit scale perturbations.
We then pass these gradients to Adam (Optax), which produces an update tensor \(u\) for each parameter leaf.


\paragraph{Adam updates for int8 tensors (discretised).}
For integer parameters (notably int8), Adam produces a real-valued proposal \(u\) (stored in bf16).
Since the parameter itself must remain int8, we convert this proposal into a sparse unit-step update using a normalised thresholding rule. Let \(Q\in\mathbb{Z}_8^{m\times n}\) be an int8 tensor and \(u\in\mathbb{R}^{m\times n}\) be Adam's proposed update.
We compute a per-tensor z-score normalisation
\[
z = \frac{u-\mu(u)}{\sigma(u)+10^{-8}},
\]
then apply a threshold \(\tau\) to form the integer step
\[
\Delta = \mathrm{sign}(z)\cdot \mathbbm{1}\{|z|\ge \tau\}\in\{-1,0,+1\}^{m\times n}.
\]
Finally we update by unit increments and clip to the valid int8 range:
\[
Q \leftarrow \mathrm{clip}(Q+\Delta,\,-127,127).
\]
Intuitively, Adam supplies a magnitude- and history-aware proposal, while the discretisation enforces the integer constraint and yields a stable, sparse update pattern (only entries with sufficiently large normalised updates are modified).

\paragraph{Memory considerations.}
We store Adam's optimiser state (moments) in bf16 for all array-valued leaves to reduce memory footprint, while keeping scalar bookkeeping in full precision.
This keeps the dominant memory cost of optimisation close to that of the parameters themselves, which is particularly important when fine-tuning large models with large ES populations.

\paragraph{Model distillation.} We distil a non-quantised model into the quantised RWKV-7 model by matching the two distributions in  teacher forced examples from GSM8k. More specifically, the fitness for a given set of parameters, $\mu_i$, is computed as follows:
\begin{equation}
    f_{\mu_i}(x_{1:T}) = \sum_{t=1}^T \text{KL}\left(p_t||q_t(\cdot; \mu_i)\right),
\end{equation}
where $x_{1:T}$ is a subsequence of tokens taken from the solutions of GSM8K and $\text{KL}\left(p_t||q_t(\cdot; \mu_i)\right)$ is the Kullback-Leibler divergence between the distribution of the non-quantised model, $p_t$, and the distribution of the quantised model $q_t$ over the vocabulary at token $t$. 

\section{Fine-tuning Pretrained Transformer LLMs with Verifiable Rewards}\label{sec:vllm-exps}

This section describes compares EGGROLL to standard RL from Verifiable Rewards (RLVR). We first describe our experimental results, before  including details of the infrastructure used to run these experiments.

\subsection{Results}\label{subsec:vllm-exps-results}

Here we demonstrate that EGGROLL can be used to fine-tune pre-trained LLMs on verifiable rewards. We use the vLLM library \cite{kwon2023efficient} for efficient inference. More infrastructure detail is given in \Cref{subsec:vllm-infrastructure}.

\begin{figure}[h!]
    \centering
    \includegraphics[width=0.9\linewidth]{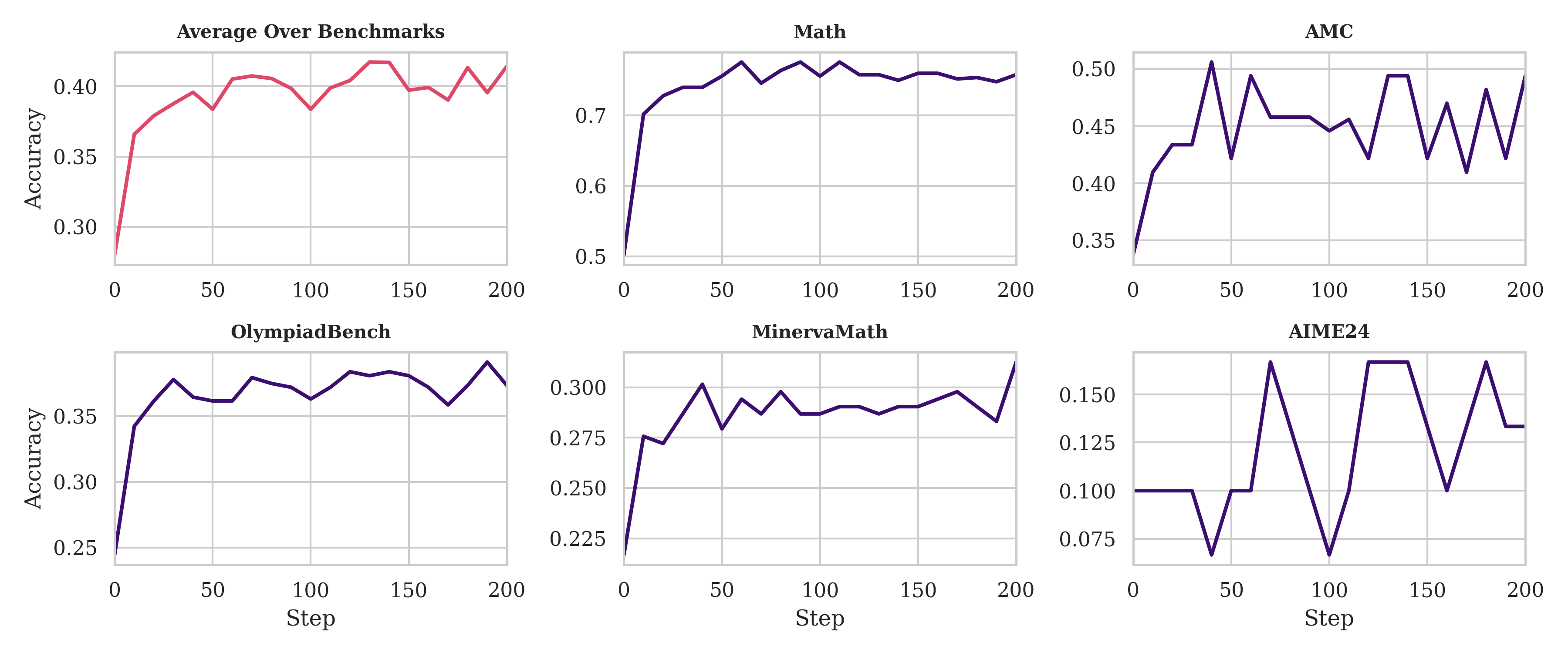}
    \caption{Training curves for fine-tuning Qwen3-4B-Base on the DeepScaleR math dataset. Similar to RL from Verifiable Rewards (RLVR), we see that optimising with EGGROLL is able to improve chain-of-thought reasoning performance on a range of math benchmarks.}
    \label{fig:vllm-4b-base}
\end{figure}

We first fine-tune the Qwen3-4B-Base model \cite{yang2025qwen3technicalreport} on the DeepScaleR \cite{deepscaler_dataset_2025}, a dataset of 40k maths questions. As in standard RLVR, the model generates a chain-of-thought (CoT) followed by a final answer. Fitness is then simply calculated by extracting the final answer and comparing it to the ground truth answer \cite{shao2025deepseekmathv2selfverifiablemathematicalreasoning}. We evaluate performance on MATH500 \cite{mathdataset}, OlympiadBench \cite{olympiadbench}, AIME24 \cite{aime}, AMC, and MinervaMath \cite{minerva}. Training curves are shown in \Cref{fig:vllm-4b-base}. Here we see that fine-tuning with EGGROLL significantly improves performance over the base model. In \Cref{tab:vllm-4b-base} we show final accuracies with EGGROLL and with the equivalent RL experiment. The RL values are taken from \cite{liu2025gemgymagenticllms}, and we match all the relevant shared hyperparameters and setup, such as maximum response length and prompt phrasing. We see that EGGROLL is able to match the RL optimisation with very minimal hyperparameter tuning, a LoRA rank of 1 and a moderately small population size of 2048. Full hyperparameter details are given in \Cref{tab:vllm_exp_params}.

\begin{table}[h!]
\centering
\begin{tabular}{lllllll }
\hline
 & MATH500 & OlympiadBench & AIME24 & AMC  & MinervaMath & \textbf{Average} \\ \hline
\textit{Qwen3-4B-Base}            & 50.2    & 24.4          & 10.0   & 33.7 & 21.7        & 28.0    \\
\:\:\:\:+EGGROLL                 & \textbf{75.8}    & \textbf{37.3}          & 13.3   & \textbf{49.4} & 31.3        & \textbf{41.4}    \\
\:\:\:\:+RL                      & 67.4    & 33.5          & \textbf{16.7}   & \textbf{49.4} & \textbf{40.1}        & \textbf{41.4}    \\ \bottomrule
\end{tabular}\label{tab:vllm-4b-base}
\caption{Final test accuracies when training on the DeepScaleR dataset to optimise verifiable rewards with EGGROLL and RL. We see that EGGROLL significantly boosts performance from the base model and is able to match the equivalent RL experiment.}
\end{table}

\begin{figure}[h!]
    \centering
    \includegraphics[width=0.6\linewidth]{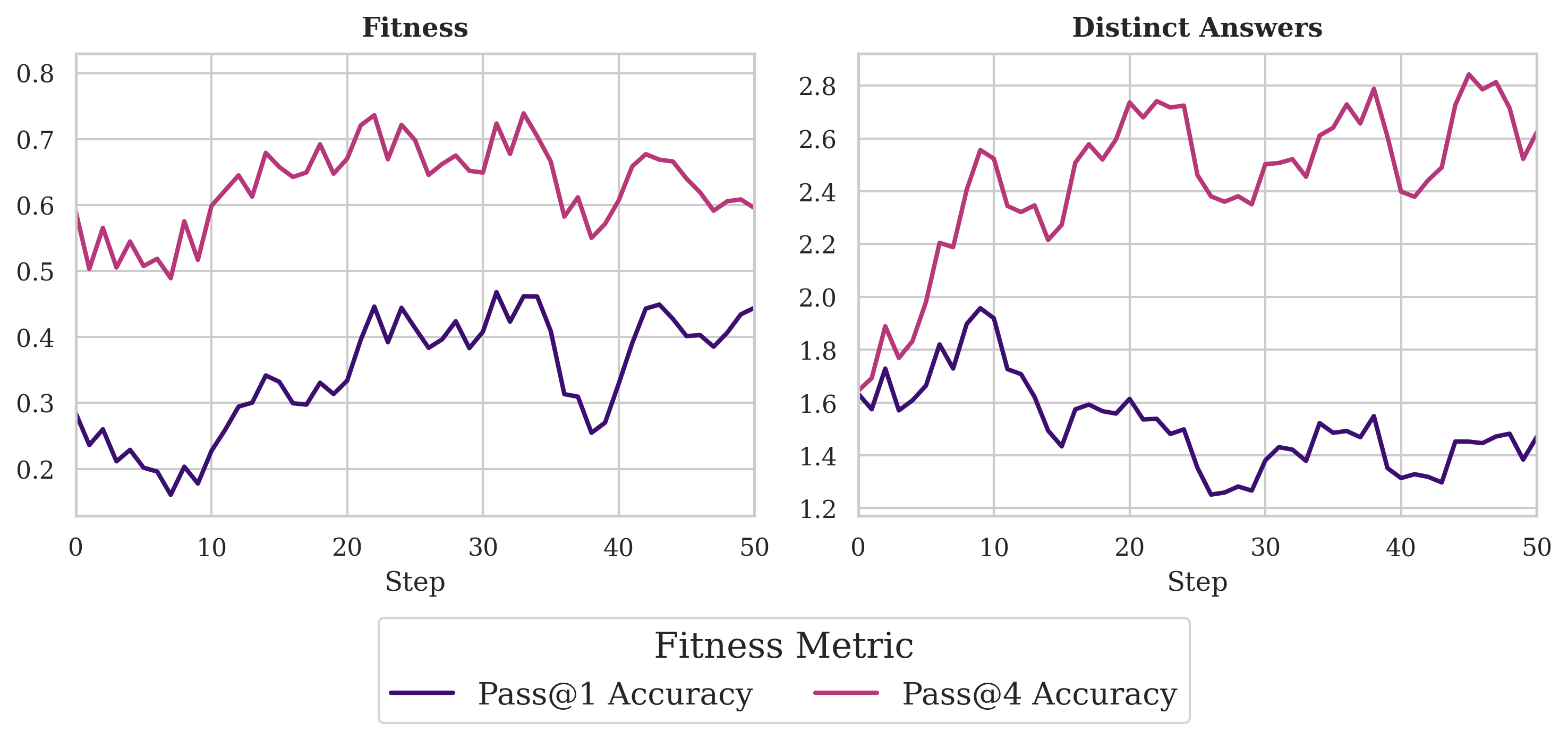}
    \caption{Using EGGROLL to optimise non-differentiable objectives. \textit{Left}: Fitness curves comparing training with pass@1 (differentiable) versus pass@k (non-differentiable), where $K=4$. \textit{Right}: The mean number of unique final answers generated per 4-sample set. We observe that when optimizing for pass@k increases answer diversity, whereas optimizing for zero-shot accuracy (pass@1) reduces it.}
    \label{fig:vllm-pass_at_k}
\end{figure}

Since EGGROLL can be used to optimise non-differentiable objectives we next try optimising for pass@k. While zero-shot (pass@1) is differentiable, the pass@k objective is not as it depends on multiple samples from the model. This means it cannot be optimised easily with RL. In \Cref{fig:vllm-pass_at_k} we fine-tune the Qwen3-1.7B model on the DeepScaleR dataset with a population size of 256, LoRA rank 1, and $K=4$. We see that EGGROLL successfully optimises both the pass@1 (differentiable) and pass@k (non-differentiable) objectives. In \Cref{fig:vllm-pass_at_k} \textit{(right)} we plot the number of distinct answers in 4 samples from the model. We see then when optimising for pass@k the answer diversity sampled by the model increases over training, whereas when optimising for zero-shot (pass@1) the model collapses towards a single final answer.

\subsection{Training Infrastructure for Large-Scale Transformer LLMs}
\label{subsec:vllm-infrastructure}

EGGROLL facilitates the fine-tuning of transformer-based LLMs at scale. We achieve this by repurposing the \texttt{vLLM} inference engine, leveraging its high-throughput kernel implementations and native support for multi-LoRA serving. The system utilises vLLM’s native Tensor Parallelism (TP) to shard the model weights across the GPUs within a node, while cross-node parallelisation is employed for the concurrent evaluation of the LoRA population.

To render ES-based optimisation feasible and efficient across a wide range of model sizes, we implement several critical systems-level optimisations:

\paragraph{Custom \texttt{WorkerExtension} and Sharding-Aware Updates}
By implementing a custom \texttt{WorkerExtension}, we effectively convert the vLLM inference engine into a training-capable runtime. This extension allows the optimisation logic to reside within the GPU process space, enabling direct, in-place manipulation of the model’s weights. A significant complexity of this integration is vLLM’s internal tensor parallelism, which frequently fuses weights (e.g. combining \texttt{q\_proj}, \texttt{k\_proj}, and \texttt{v\_proj} into a single \texttt{qkv\_proj} tensor). Our update mechanism is explicitly ``sharding-aware''; it constructs a dictionary which maps individual LoRA updates to the specific fused slices held by each local GPU rank. This ensures that the global ES update is mathematically consistent across all distributed shards.

\paragraph{Layer-wise Memory Management}
To prevent out-of-memory (OOM) errors during the update phase, the \texttt{WorkerExtension} performs the ES weight application in a streaming, layer-wise fashion. By processing one layer at a time and clearing temporary buffers, the memory overhead of the update remains independent of the total model depth. This allows for the fine-tuning of models of very different sizes with a VRAM footprint barely exceeding that of standard inference.

\paragraph{Direct GPU-to-GPU Weight Synchronization}
After computing the ES update on the primary rank, we broadcast the updated parameters to all model instances using NCCL via \texttt{PyNcclCommunicator}. This approach bypasses CPU-based communication and instead uses hardware interconnects to transfer weights directly between GPUs, preventing synchronization from becoming a bottleneck when scaling to more nodes.

\paragraph{Meta-Device Blueprinting}
To initialise models that exceed the physical RAM of the control node, we employ Meta-Device Initialisation. Using \texttt{accelerate}’s \texttt{init\_empty\_weights}, we instantiate a ``meta'' version of the model to derive the weight shapes and sharding requirements for the LoRA adapters. This allows the system to generate a complete parameter blueprint for models of arbitrary size without ever allocating the full weight tensors in system memory.

\paragraph{\texttt{vLLM} Engine Settings} Throughout the different experiments with \texttt{vLLM}, we use the following engine settings. These generally allow for high throughput across model sizes (e.g. at least 800 tokens/second), but we haven't performed hyperparameter sweeps, so potentially faster, more memory-efficient settings may be used for improved results.

\begin{table}[ht]
  \centering
  \begin{tabular}{ll}
    \toprule
    Parameter & Value \\
    \midrule
    Tensor parallel size & 2,4 \\
    Data type & auto \\
    Enable prefix caching & True \\
    Enforce eager execution & True \\
    Enable LoRA & True \\
    Max LoRAs & $\lceil \text{population\_size} / \text{num\_engines} \rceil$ \\
    GPU memory utilisation & 0.90 \\
    Max number of sequences & 384 \\
    Max model length & $\max(1024, 512 + \text{max\_tokens})$ \\
    Max batched tokens & $\text{prompt\_batch\_size} \times 1024$ \\
    Load format & auto \\
    \bottomrule
  \end{tabular}
  \caption{\texttt{vLLM} engine configuration parameters to allow for high throughput EGGROLL training on large-scale transformer LLMs.}
  \label{tab:vllm_params}
\end{table}

\begin{table}[ht]
  \centering
  \begin{tabular}{ll}
    \toprule
    Parameter & Value \\
    \midrule
    Population size & 256, 2048 \\
    Sigma & 0.001 \\
    Learning Rate & 0.001 \\
    Max Response Length & 4096 \\
    Temperature & 0.0, 0.7 \\
    Samples Per Prompt & 1, 4 \\
    Pass at K & True, False \\
    LoRA Rank & 1 \\
    LoRA Reuse Steps & 4 \\
    \bottomrule
  \end{tabular}
  \caption{Hyperparameters for the verifiable reward transformer fine-tuning experiments in \Cref{subsec:vllm-exps-results}.}
  \label{tab:vllm_exp_params}
\end{table}

\newpage

\section{Fine-tuning Time Series Foundation Model: High-Frequency Trading}
\label{sec:eggroll_trading}

The preceding experiments demonstrate the effectiveness of EGGROLL on natural language reasoning tasks. We now investigate whether EGGROLL can effectively fine-tune pretrained foundation models on a fundamentally different data modality: structured time series. We focus on high-frequency trading (HFT) for two reasons. First, HFT generates data at an unprecedented scale. The S\&P 500 constituents alone produced approximately 3.8 trillion tokens of order flow data between 2016 and 2021, comparable to the largest natural language corpora. Second, the domain presents a well-defined downstream task (order execution) with a natural reward signal: the realised profit and loss, also known as PnL, making it amenable to fine-tuning via evolution strategies.


Order execution takes place in limit order books (LOBs), which are the mechanism upon which modern financial exchanges operate \citep{gould2013limit,bouchaud2018trades}. They allow market participants to submit limit orders that specify the details of intended transactions. Specifically, each limit order contains the order type, direction, price, and quantity. The continuous stream of these orders is known as the order flow. LOBs aggregate the limit orders that have not been matched yet. Unlike natural language, where tokens are purely symbolic, order flow messages comprise both categorical values (e.g., order type, direction) and numerical values (e.g., price, quantity) in which magnitude carries semantic meaning. This structure provides a distinct test of EGGROLL's ability to fine-tune foundation models on time series sequential data.

A central objective in this context is order execution, which consists of buying or selling a specified quantity of an asset within a given time window. The goal is to maximise profit by transacting at favourable prices. In prior reinforcement learning approaches to this problem, the action space is usually simplified \citep{frey2023jax, mohl2025jaxmarlhft, ning2021double}. In contrast, we aim to give the model full flexibility in choosing limit orders, i.e., to freely choose the order type, direction, price, and quantity. We achieve this by tokenising the limit order book messages and providing the model with a token-level action space.

Foundation models have recently been used to generate synthetic order flow \citep{NagyGenerativeAI2023,li2025mars} and have been shown to replicate realistic market behaviour \citep{nagy2025lobbench} through next-token prediction. We therefore first pretrain a foundation model on tokenised limit order book messages, and then fine-tune it using EGGROLL for the order execution task. The pretraining follows the approach of \citet{NagyGenerativeAI2023}: we employ an S5 model architecture \citep{smith2023s5} that generates next-token probabilities, with cross-entropy as the training loss. The pretraining is conducted on the LOBSTER data set \citep{huang2011lobster} for the Google stock (GOOG) in 2022, which contains around 25B tokens.

Subsequently, we fine-tune the model using EGGROLL. The training parameters are summarised in Table~\ref{tab:eggroll_training}. The task is to execute a sell order of $Q=30$ shares within a horizon of $T=10$ steps. In each episode, the LOB is initialised based on a LOB snapshot followed by 10 warm-up background messages. In each step, the population members generate their messages, which are then followed by 50 real background data messages. The orders are executed using the Jax-LOB \citep{frey2023jax} simulator. We perform the fine-tuning on a fixed time window for GOOG in January 2023. Following \citep{lora2025ssm}, we apply LoRA with rank 4 on all projection matrices while freezing SSM parameters and layer norms. Performance is evaluated using PnL based on the executed prices and the initial mid price. Specifically, for a sell task of total quantity $Q$, the PnL is computed as
\[
\sum_{i=1}^{N} q_i p_i - Q\, P_{\text{mid}}^{\text{init}},
\]
where $q_i$ and $p_i$ denote the quantity and price of the $i$-th executed trade and $P_{\text{mid}}^{\text{init}}$ is the mid-price at the beginning of the execution window. If the agent does not execute the entire quantity by the end of the episode, an automatic market order is submitted selling the remaining quantity. To improve robustness to outliers, fitness is defined as a rank-based transformation of the PnL. Specifically, for a population of size $M$, the PnL values
\[
\mathcal{P} = \{ \mathrm{PnL}_1, \dots, \mathrm{PnL}_M \},
\]
are mapped to the interval $[-0.5, 0.5]$, where $\mathrm{rank}(\mathrm{PnL}_i) \in \{0, \dots, M - 1\}$ denotes the rank of the $i$-th individual's PnL:
\[
F_i =  \frac{1}{2} - \frac{\mathrm{rank}(\mathrm{PnL}_i)}{M - 1} ,
\]

\begin{figure}[t]
    \centering
    \includegraphics[width=0.65\linewidth]{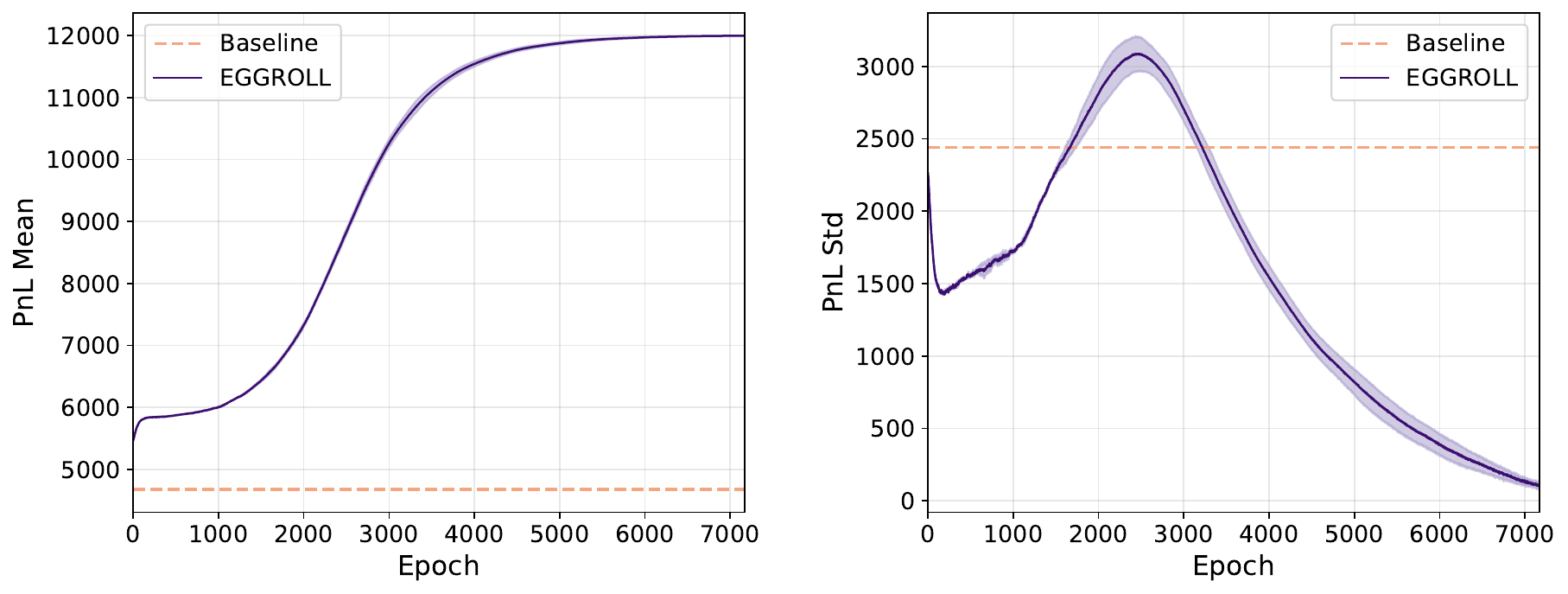}
    \caption{Training curves for order execution with EGGROLL. \textbf{Left}: Mean PnL over training epochs for the baseline ($\sigma=0$, orange dashed) and EGGROLL ($\sigma=0.01$, blue solid). \textbf{Right}: PnL standard deviation over training epochs. Shaded regions indicate the interquartile range across runs.}
    \label{fig:lob_extended_training}
\end{figure}

\begin{table}[ht]
  \centering
  \begin{tabular}{ll}
    \hline
    Hyperparameter & Value \\
    \hline
    Model & LOBS5-360M \\
    Parallel generations per GPU & 2{,}048 \\
    Total parallel generations & 65{,}536 \\
    LoRA rank & 4 \\
    Sigma & 0.01 \\
    Learning rate $\eta$  & 0.001 \\
    Epochs & 6{,}500 \\
    \hline
  \end{tabular}
  \caption{Model and EGGROLL fine-tuning settings for high-frequency trading.}
  \label{tab:eggroll_training}
\end{table}

Training curves over 6,500 epochs are shown in Figure~\ref{fig:lob_extended_training}. The baseline policy ($\sigma=0$), corresponding to the pretrained model, achieves a mean PnL of approximately 4,700. In contrast, EGGROLL fine-tuning ($\sigma=0.01$) improves the mean PnL to around 12,000, corresponding to a roughly 155\% improvement over the baseline. The right panel of Figure~\ref{fig:lob_extended_training} depicts the PnL standard deviation during fine-tuning: it initially increases to around 3,100 during the first 2,500 epochs, which corresponds to an exploration phase where the population tries out diverse strategies, before decreasing to approximately 400 by the end of training, indicating that the population concentrates around a high-performing policy.

\section{Experimental Details} 
\label{sec:experiment_appendix}
\subsection{Multi Agent Reinforcement Learning Experiments} \label{app:marl}

\begin{figure}[t]
    \centering
    \includegraphics[width=0.85\linewidth]{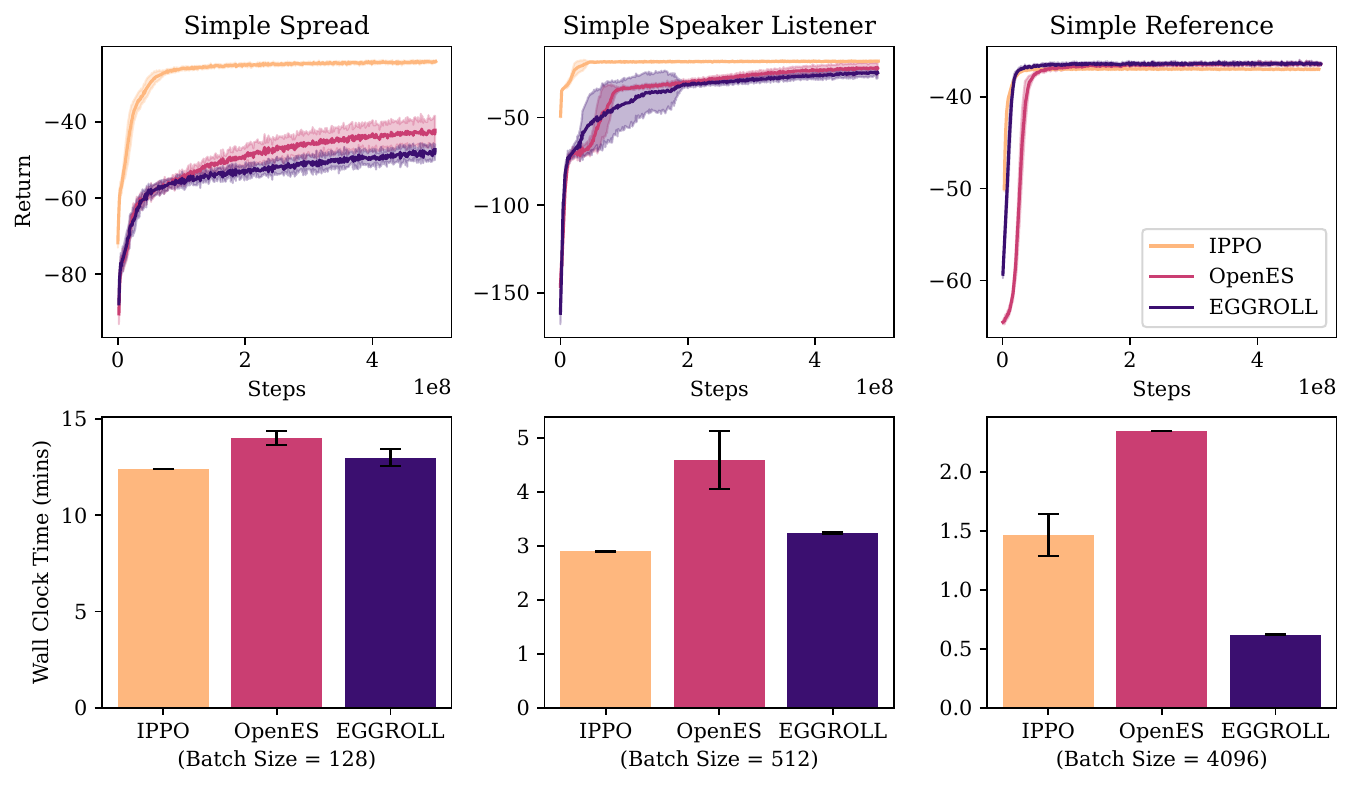}
    \caption{Training curves and wall clock times for cooperative Multi Particle Environments. Hyperparameter optimisation yielded equal batch sizes for all algorithms on the same environment. All EGGROLL runs used rank 1 perturbations. Shaded regions are standard errors of mean values.}
    \label{fig:marl_mpe}
\end{figure}

\begin{table}[htb!]
\centering
\begin{minipage}{0.48\textwidth}
\centering
\caption{Hyperparameter Ranges Used in MPE Sweeps for EGGROLL and OpenES}
\begin{tabular}{lcc}
\hline
Hyperparameter & Values \\
\hline
activation & pqn,\ tanh \\
pop\_size & 128, 512,\ 1024,\ 2048,\ 4096 \\
learning\_rate & 0.01,\ 0.05,\ 0.1,\ 0.5 \\
lr\_decay & 0.3,\ 0.7,\ 1.0 \\
sigma & 0.1,\ 0.2,\ 0.3,\ 0.4,\ 0.5 \\
rank\_transform & true,\ false \\
\hline
\end{tabular}
\end{minipage}
\hfill
\begin{minipage}{0.48\textwidth}
\centering
\caption{Hyperparameter Ranges Used in MPE Sweeps for IPPO}
\begin{tabular}{lccc}
\hline
Hyperparameter & Values \\
\hline
activation & relu,\ tanh \\
pop\_size & 128, 512,\ 1024,\ 2048,\ 4096 \\
learning\_rate & 5e-5,\ 1e-4,\ 2.5e-4,\ 1e-3 \\
entropy\_coef & 0.001,\ 0.005,\ 0.01 \\
\hline
\end{tabular}
\end{minipage}
\end{table}

\begin{table}[htb!]
\centering
\begin{minipage}{0.48\textwidth}
\centering
\caption{MPE Simple Spread v3}
\begin{tabular}{lccc}
\hline
Hyperparameter & eggroll & open\_es & ippo \\ \hline
activation & tanh & tanh & tanh \\
deterministic\_policy & true & true & false \\
learning\_rate & 0.01 & 0.01 & 0.001 \\
lr\_decay & 0.7 & 0.7 & linear \\
layer\_size & 64 & 64 & 64 \\
n\_layers & 3 & 3 & 3\\
pop\_size & 128 & 128 & 128\\
optimizer & adamw & adamw & adam\\
rank & 1 & 1 & - \\
rank\_transform & false & false & - \\
sigma & 0.5 & 0.5 & - \\
n\_minibatches & - & - & 4 \\
update\_epochs & - & - & 4 \\
gamma & - & - & 0.99 \\
gae\_lambda & - & - & 0.95 \\
epsilon\_clip & - & - & 0.2 \\
entropy\_coef & - & - & 0.01 \\
value\_coef & - & - & 0.5 \\
max\_grad\_norm & - & - & 0.5 \\
\hline
\end{tabular}
\end{minipage}
\hfill
\begin{minipage}{0.48\textwidth}
\centering
\caption{MPE Simple Speaker Listener v4}
\begin{tabular}{lccc}
\hline
Hyperparameter & eggroll & open\_es & ippo \\ \hline
activation & tanh & tanh & relu \\
deterministic\_policy & true & true & false \\
learning\_rate & 0.01 & 0.01 & 0.001 \\
lr\_decay & 0.7 & 0.3 & linear \\
layer\_size & 64 & 64 & 64 \\
n\_layers & 3 & 3 & 64 \\
pop\_size & 512 & 512 & 512 \\
optimizer & adamw & adamw & adam \\
rank & 1 & 1 &  - \\
rank\_transform & true & true & - \\
sigma & 0.5 & 0.5 & - \\
n\_minibatches & - & - & 4 \\
update\_epochs & - & - & 4 \\
gamma & - & - & 0.99 \\
gae\_lambda & - & - & 0.95 \\
epsilon\_clip & - & - & 0.2 \\
entropy\_coef & - & - & 0.005 \\
value\_coef & - & - & 0.5 \\
max\_grad\_norm & - & - & 0.5 \\
\hline
\end{tabular}
\end{minipage}
\end{table}

\begin{table}[htb!]
\centering
\centering
\caption{MPE Simple Reference v3}
\begin{tabular}{lccc}
\hline
Hyperparameter & eggroll & open\_es & ippo \\ \hline
activation & pqn & tanh & relu \\
deterministic\_policy & true & true & false \\
learning\_rate & 0.01 & 0.01 & 0.001 \\
lr\_decay & 0.3 & 0.3 & linear\\
layer\_size & 64 & 64 & 64 \\
n\_layers & 3 & 3 & 3\\
pop\_size & 4096 & 4096 & 4096 \\
optimizer & adamw & adamw & adam \\
rank & 1 & 1 & - \\
rank\_transform & false & true & - \\
sigma & 0.1 & 0.3 & - \\
n\_minibatches & - & - & 4 \\
update\_epochs & - & - & 4 \\
gamma & - & - & 0.99 \\
gae\_lambda & - & - & 0.95 \\
epsilon\_clip & - & - & 0.2 \\
entropy\_coef & - & - & 0.01 \\
value\_coef & - & - & 0.5 \\
max\_grad\_norm & - & - & 0.5 \\
\hline
\end{tabular}
\end{table}
\newpage
We train on three cooperative Multi Particle Environments (MPEs) \citep{lowe2017multi} implemented in JaxMARL \citep{jaxmarl} with feed-forward networks of width 64 and depth 3, performing Bayesian hyperparameter optimisation for each environment and algorithm. All runs were executed on NVIDIA A100-SXM4-40GB GPUs. We find that the optimal batch size is consistent across algorithms on the same environment. \Cref{fig:marl_mpe} shows that EGGROLL with rank 1 trains up to 2.4 times faster than OpenES for large batch sizes while staying competitive in performance.

\newpage

\vspace*{32pt}

\subsection{Reasoning Fine-tuning Experiments: Countdown}

We ran a Bayesian hyper-parameter sweep \citep{snoek2012practicalbayesianoptimizationmachine} for both GRPO and EGGROLL and used the best set found to run the experiments in figure~\ref{fig:eggroll_countdown_1}. For GRPO we swept over sampling temperature and learning rate, whereas for EGGROLL we swept over the standard deviation of the ES sampling ($\sigma$) and the learning rate scale. The best hyper-parameters found are detailed on tables~\ref{tab:eggroll_hparams_countdown} (EGGROLL) and~\ref{tab:grpo_hparams_coundown} (GRPO). All of the experiments run in 8 hours on a NVIDIA H200 GPU.

\begin{figure*}[t]
  \centering
  \begin{subfigure}[t]{0.493\textwidth}
    \centering
    \includegraphics[width=\linewidth]{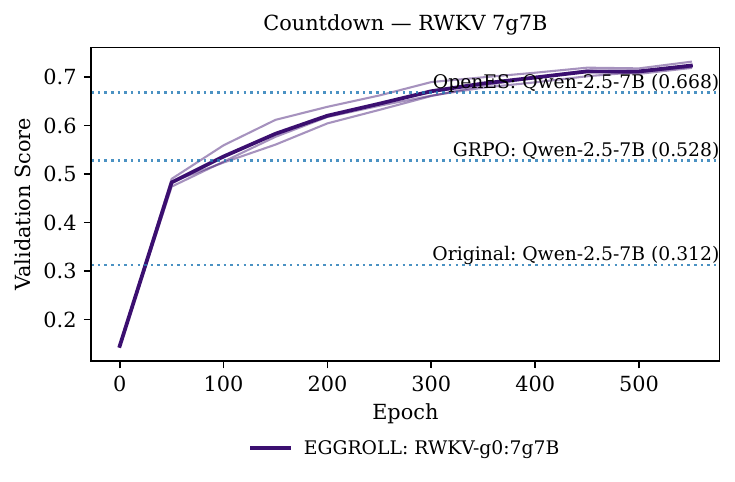}
    \caption{}
    \label{fig:eggroll_countdown_7B}
  \end{subfigure}
  \hfill
  \begin{subfigure}[t]{0.5\textwidth}
    \centering
    \includegraphics[width=\linewidth]{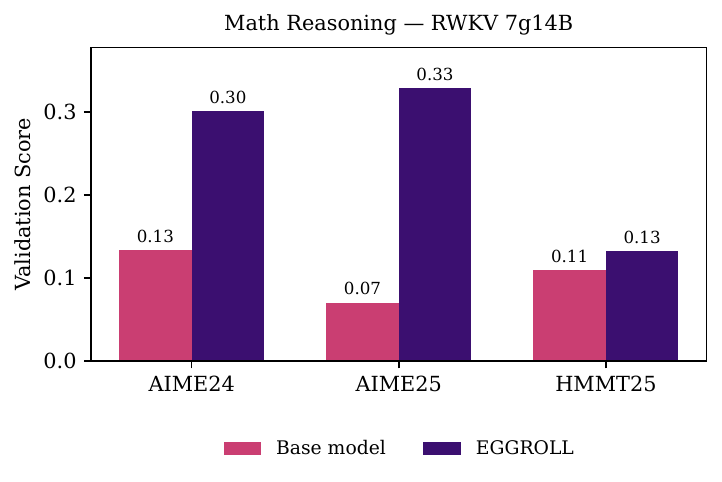}
    \caption{}
    \label{fig:eggroll_hard_reasoning_14B}
  \end{subfigure}
  \caption{(a) Comparison of our finetuned RWKV 7G 7 billion parameter model using 8 GPUS with the results reported by \cite{qiu2025evolutionstrategiesscalellm} on similarly sized Qwen models. (b) Performance of our finetuned RWKV 7G 14 billion parameter model on hard reasoning tasks using 32 GPUs for 12 hours. The model was trained using the DeepScaleR dataset and the best checkpoint was chosen by evaluating on AIME24. Due to the size of the model we were not able to run similar baseline experiments using GRPO.}
  \label{fig:eggroll_hard_reasoning_v0}
\end{figure*}

\begin{table}[h]
  \centering
  \begin{tabular}{ll}
    \toprule
    Hyperparameter & Value \\
    \midrule
    Model & RWKV 7g1.5B \\
    Optimiser & Gradient descent \\
    ES standard deviation $\sigma$ & $7 \times 10^{-4}$ \\
    Rank $r$ & 1 \\
    Learning-rate scale $\eta_{\text{scale}}$ & 0.125 \\
    Population size & 256 \\
    Parallel generations per GPU & 1536 \\
    Prompts per epoch & 6 \\
    Generation / thinking length & 1000 tokens \\
    Train / val temperature & 0 / 0 \\
    Parallel validations & 128 \\
    \bottomrule
  \end{tabular}
  \caption{Key hyperparameters for EGGROLL training on Countdown with FastRWKV-7g1.5B.}
  \label{tab:eggroll_hparams_countdown}
\end{table}

\begin{table}[h]
  \centering
  \begin{tabular}{ll}
    \toprule
    Hyperparameter & Value \\
    \midrule
    Model & RWKV 7g1.5B \\
    Optimiser & Radam \\
    Learning rate $\eta$ & $3 \times 10^{-6}$ \\
    Generations per prompt $G$ & 8 \\
    Parallel generations per GPU & 64 \\
    Prompts per epoch & 8 \\
    Generation length & 1000 tokens \\
    Number of minibatches & 4 \\
    PPO clip parameter $\epsilon_{\text{clip}}$ & 0.2 \\
    Train / val temperature & 1 / 0 \\
    Parallel validations & 128 \\
    \bottomrule
  \end{tabular}
  \caption{Key hyperparameters for GRPO training on Countdown with AssociativeScanRWKV-7g1.5B.}
  \label{tab:grpo_hparams_coundown}
\end{table}

We also run an experiment where we increase the number of GPUs to 8 and use a bigger model, RWKV 7g7B, on the Countdown task, allowing for stronger final performance. Notably, we compare to the results reported by \cite{qiu2025evolutionstrategiesscalellm} on Countdown. Figure~\ref{fig:eggroll_countdown_7B} shows that starting from our significantly weaker model (RWKV 7g7B v.s. Qwen 2.5-7B), we are able to train to a higher validation accuracy (72.9\%), v.s. the ones reported for training with GRPO (52.8\%) and Open ES (66.8\%). \cite{qiu2025evolutionstrategiesscalellm} do not report the wall clock time or the hardware used for their experiments which makes it difficult to establish a fair comparison.

\subsection{Reasoning Fine-tuning Experiments: GSM8K}

We used the hyper-parameters found for Countdown as a starting point and reduced the learning rates for both GRPO and EGGROLL using linear search until we found the best performing one on the validation set. Our experiments for GSM8K run on 8 NVIDIA H200 GPUS for 8 hours each. We also increase the standard deviation, $\sigma$, parameter for ES (from $7 \times 10^{-4}$ to $2 \times 10^{-3}$) as the significantly bigger population sizes (8096 v.s. 512) allow for much more stable training and aggressive exploration.


\begin{table}[h]
  \centering
  \begin{tabular}{ll}
    \toprule
    Hyperparameter & Value \\
    \midrule
    Model & RWKV 7g7B \\
    ES standard deviation $\sigma$ & $2 \times 10^{-3}$ \\
    Rank $r$ & 1 \\
    Learning-rate scale $\eta_{\text{scale}}$ & 0.06 \\
    Generations per prompt $G$ & 512 \\
    Parallel generations per GPU & 1024 \\
    Total parallel generations & 8192 \\
    Prompts per epoch & 16 \\
    Generation length & 1000 tokens \\
    Noise reuse factor & 1 \\
    Freeze non-LoRA params & True \\
    Train / val temperature & 0 / 0 \\
    Parallel validations & 128 \\
    \bottomrule
  \end{tabular}
  \caption{Key hyperparameters for multi-GPU EGGROLL training on GSM8K with FastRWKV-7g7B.}
  \label{tab:eggroll_gsm8k_hparams}
\end{table}

\begin{table}[ht]
  \centering
  \begin{tabular}{ll}
    \toprule
    Hyperparameter & Value \\
    \midrule
    Model & RWKV 7g7B \\
    Learning rate $\eta$ & $1 \times 10^{-6}$ \\
    Generations per prompt $G$ & 8 \\
    Parallel generations per GPU & 32 \\
    Total parallel generations & 256 \\
    Prompts per epoch & 32 \\
    Generation length & 1000 tokens \\
    Number of minibatches & 16 \\
    Number of workers (processes) & 8 \\
    PPO clip parameter $\epsilon_{\text{clip}}$ & 0.2 \\
    Train / val temperature & 1 / 0 \\
    Parallel validations & 128 \\
    \bottomrule
  \end{tabular}
  \caption{Key hyperparameters for multi-GPU GRPO training on GSM8K with AssociativeScanRWKV-7g7B.}
  \label{tab:grpo_gsm8k_hparams}
\end{table}

\begin{table}[h]
  \centering
  \begin{tabular}{ll}
    \toprule
    Hyperparameter & Value \\
    \midrule
    Model & RWKV 7g7B \\
    Optimiser & EGGROLL (Quantised)) \\
    ES standard deviation $\sigma$ & 0.4 \\
    Rank $r$ & 1 \\
    Learning-rate scale $\eta_{\text{scale}}$ & $3\times 10^{-7}$ \\
    Population size & 8192 \\
    Parallel generations per GPU & 256 \\
    Prompts per epoch & 1 \\
    Generation / thinking length & 256 tokens \\
    Train / val temperature & 0 / 0 \\
    Parallel validations & 128 \\
    \bottomrule
  \end{tabular}
  \caption{Key hyperparameters for quantised EGGROLL training on GSM8K (teacher-forced) with RWKV-\texttt{7g7B}.}
  \label{tab:eggroll_hparams_gsm8k_distill}
\end{table}

\subsection{Reinforcement Learning Experiments} \label{app:rl}
\begin{figure}[!htp]
    \centering
    \includegraphics[width=1.0\linewidth]{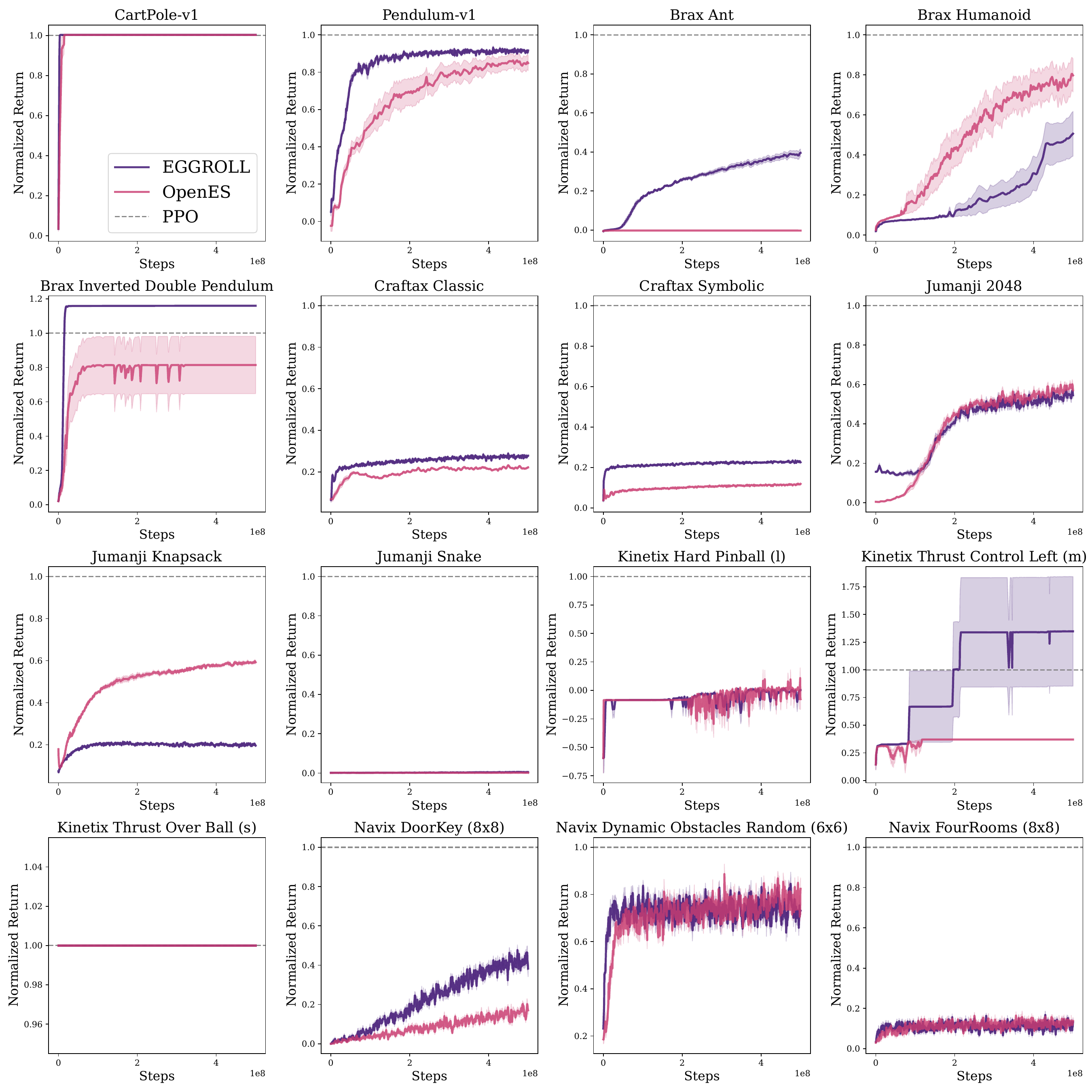}
    \caption{Comparison of reinforcement learning results: Mean returns for each environment and algorithm across 10 random seeds. The returns are evaluated using the mean of the parameters. HPO was conducted for each algorithm/environment pair. The shaded region is the standard error of the mean.}
    \label{fig:rl_returns_full}
\end{figure}
Next, we compare the performance of EGGROLL against standard OpenES as implemented in \cite{salimans2017} on reinforcement learning tasks. 
Given the small network sizes, we can use OpenES at this scale, but as network sizes increase, the use of vanilla OpenES becomes computationally infeasible.
We use the standard formulation of simply optimising for the final return in the environment. 
For both EGGROLL and OpenES, we perform hyperparameter optimisation (HPO) separately for each environment.
For each algorithm–environment pair, we define plausible ranges for all key hyperparameters based on prior work and preliminary experiments. 
We then perform 20 random search trials, where each trial corresponds to a single training run with a randomly sampled hyperparameter configuration. 
Each configuration is evaluated based on the final return achieved by the mean policy parameters at the end of training.
After all trials, we select the configuration that yields the highest final return. Using this best configuration, we then run 10 independent seeds to evaluate performance and report the mean and standard error of the mean across these seeds.

\begin{figure}[htb!]
    \centering
    \includegraphics[width=1.0\linewidth]{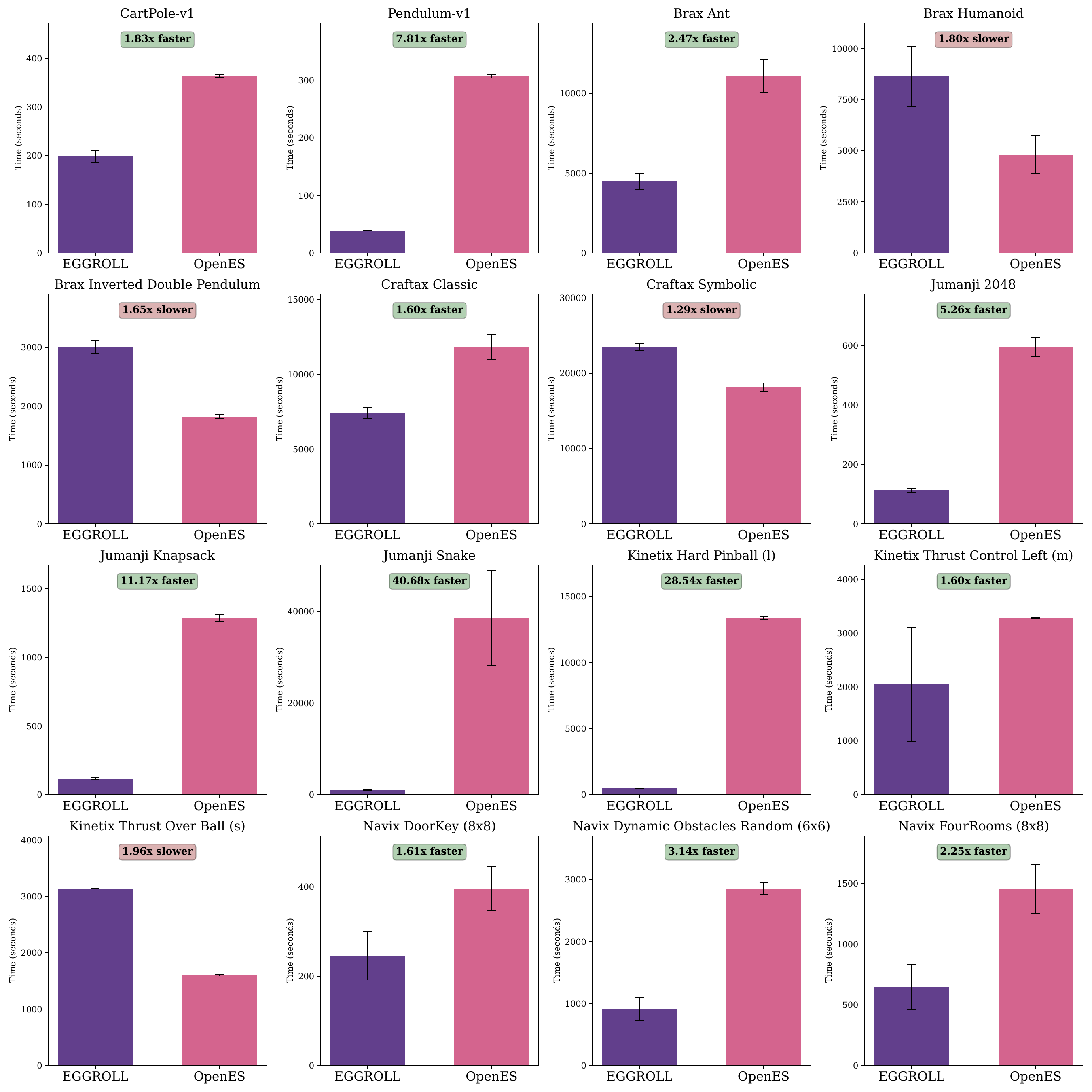}
    \caption{Comparison of reinforcement learning results: Mean and standard deviation of training time. Note that some of the timing difference is due to the differences in episode lengths, which is why the total time for EGGROLL sometimes appears longer than OpenES despite EGGROLL being faster on a per-timestep basis.}
    \label{fig:rl_timings_full}
\end{figure}

We use policy networks with 3 layers of 256 neurons and a range of environments that demonstrate different capabilities. 
We evaluate across the Navix \citep{pignatelli2024navixscalingminigridenvironments}, Craftax \citep{craftax}, Brax \citep{freeman2021braxdifferentiablephysics}, Kinetix \citep{Matthews25}, and Jumanji \citep{jumanji} suites of environments. 
We evaluate 16 environments in total. 
We choose environments that are not trivial or impossible for PPO to solve, according to the original papers. 
We also choose environments that belong to different categories (e.g., environment size in Kinetix or categories in Jumanji). 

We show a subsample of the evaluated environments in~\cref{fig:rl_combined} with the remaining results and hyperparameter details in Appendix~\ref{app:rl}. Our findings show that EGGROLL is competitive with Open ES on 7/16 environments, underperforms on 2/16, and outperforms on 7/16. This does not take into account the speed-ups when compared to using OpenES with full-rank updates (see Figure~\ref{fig:rl_timings_full}). We postulate that the reason for this performance increase is that the large networks are difficult to optimise for OpenES and lend themselves well to low-rank updates.

We present here the hyperparameter ranges we used for hyperparameter optimisation, as well as all hyperparameter settings for all the experiments. All RL experiments were run on an NVIDIA L40S GPU. For PPO, we use the same methodology to tune the hyperparameters as we did for OpenES and EGGROLL as described in~\cref{ssec:rl_exp}. We report the ranges and the final hyperparameters here. We train PPO agents using Rejax \citep{rejax}. We use the activation function from~\citet{gallici2025simplifying} in our experiments, which we refer to as the ``pqn'' activation function in our hyperparameter tables.

\begin{table}[htb!]
\centering
\caption{Hyperparameter Ranges for EGGROLL and OpenES}
\begin{tabular}{l l}
\hline
\textbf{Hyperparameter} & \textbf{Values} \\
\hline
pop\_size & 512,\ 1024,\ 2048,\ 4096 \\
n\_parallel\_evaluations & 1,\ 4,\ 8 \\
rank & 1,\ 2,\ 4 \\
optimizer & adamw,\ sgd,\ adam \\
learning\_rate & 1e-3,\ 1e-2,\ 1e-1 \\
lr\_decay & 0.995,\ 0.999,\ 0.9995,\ 1.0 \\
sigma & 0.05,\ 0.2,\ 0.5 \\
sigma\_decay & 0.995,\ 0.999,\ 0.9995,\ 1.0 \\
rank\_transform & true,\ false \\
deterministic\_policy & true,\ false \\
\hline
\end{tabular}
\end{table}

\begin{table}[h]
\centering
\caption{Hyperparameter Ranges for PPO}
\begin{tabular}{ll}
\hline
\textbf{Hyperparameter} & \textbf{Values} \\
\hline

clip\_eps & 0.1,\ 0.2,\ 0.3 \\ 
ent\_coef & 0,\ 0.0001,\ 0.001 \\
gae\_lambda & 0.9,\ 0.95,\ 0.98 \\
gamma & 0.95,\ 0.99,\ 0.995,\ 0.999 \\

learning\_rate & 0.0001,\ 0.0003,\ 0.001 \\
max\_grad\_norm & 0.5,\ 1,\ 2 \\

layer\_size & 256 \\  
n\_layers & 3 \\

normalize\_observations & true \\ 
normalize\_rewards & false \\

num\_envs & 64,\ 128,\ 256 \\ 
num\_epochs & 4,\ 8,\ 16 \\
num\_minibatches & 16,\ 32,\ 64 \\
num\_steps & 64,\ 128,\ 256 \\

reward\_normalization\_discount & 0.99 \\  
skip\_initial\_evaluation & false \\

vf\_coef & 0.5,\ 0.75,\ 1 \\

\hline
\end{tabular}
\end{table}

\begin{table}[htb!]
\centering
\begin{minipage}{0.48\textwidth}
\centering
\caption{CartPole-v1}
\begin{tabular}{lcc}
\hline
Hyperparameter & eggroll & open\_es \\ \hline
activation & pqn & pqn \\
deterministic\_policy & false & true \\
learning\_rate & 0.1 & 0.1 \\
lr\_decay & 0.9995 & 0.9995 \\
layer\_size & 256 & 256 \\
n\_layers & 3 & 3 \\
n\_parallel\_evaluations & 1 & 4 \\
pop\_size & 2048 & 512 \\
optimizer & sgd & adamw \\
rank & 4 & / \\
rank\_transform & false & true \\
sigma & 0.2 & 0.5 \\
sigma\_decay & 0.999 & 0.9995 \\
\hline
\end{tabular}
\end{minipage}
\hfill
\begin{minipage}{0.48\textwidth}
\centering
\caption{Pendulum-v1}
\begin{tabular}{lcc}
\hline
Hyperparameter & eggroll & open\_es \\ \hline
activation & pqn & pqn \\
deterministic\_policy & false & true \\
learning\_rate & 0.01 & 0.01 \\
lr\_decay & 0.995 & 0.995 \\
layer\_size & 256 & 256 \\
n\_layers & 3 & 3 \\
n\_parallel\_evaluations & 1 & 4 \\
pop\_size & 4096 & 4096 \\
optimizer & adam & adamw \\
rank & 4 & / \\
rank\_transform & false & false \\
sigma & 0.05 & 0.05 \\
sigma\_decay & 0.995 & 1 \\
\hline
\end{tabular}
\end{minipage}
\end{table}

\begin{table}[htb!]
\centering
\begin{minipage}{0.48\textwidth}
\centering
\caption{brax/ant}
\begin{tabular}{lcc}
\hline
Hyperparameter & eggroll & open\_es \\ \hline
activation & pqn & pqn \\
deterministic\_policy & false & false \\
learning\_rate & 0.01 & 0.1 \\
lr\_decay & 0.9995 & 0.995 \\
layer\_size & 256 & 256 \\
n\_layers & 3 & 3 \\
n\_parallel\_evaluations & 1 & 8 \\
pop\_size & 2048 & 512 \\
optimizer & adam & adam \\
rank & 1 & / \\
rank\_transform & false & false \\
sigma & 0.05 & 0.05 \\
sigma\_decay & 0.9995 & 0.9995 \\
\hline
\end{tabular}
\end{minipage}
\hfill
\begin{minipage}{0.48\textwidth}
\centering
\caption{brax/humanoid}
\begin{tabular}{lcc}
\hline
Hyperparameter & eggroll & open\_es \\ \hline
activation & pqn & pqn \\
deterministic\_policy & true & false \\
learning\_rate & 0.1 & 0.1 \\
lr\_decay & 1 & 0.995 \\
layer\_size & 256 & 256 \\
n\_layers & 3 & 3 \\
n\_parallel\_evaluations & 8 & 8 \\
pop\_size & 4096 & 1024 \\
optimizer & adam & sgd \\
rank & 1 & / \\
rank\_transform & true & true \\
sigma & 0.2 & 0.2 \\
sigma\_decay & 0.9995 & 0.995 \\
\hline
\end{tabular}
\end{minipage}
\end{table}

\begin{table}[htb!]
\centering
\begin{minipage}{0.48\textwidth}
\centering
\caption{brax/inverted\_double\_pendulum}
\begin{tabular}{lcc}
\hline
Hyperparameter & eggroll & open\_es \\ \hline
activation & pqn & pqn \\
deterministic\_policy & true & true \\
learning\_rate & 0.1 & 0.1 \\
lr\_decay & 1 & 0.995 \\
layer\_size & 256 & 256 \\
n\_layers & 3 & 3 \\
n\_parallel\_evaluations & 1 & 1 \\
pop\_size & 2048 & 4096 \\
optimizer & adam & adam \\
rank & 2 & / \\
rank\_transform & true & true \\
sigma & 0.5 & 0.05 \\
sigma\_decay & 0.995 & 1 \\
\hline
\end{tabular}
\end{minipage}
\hfill
\begin{minipage}{0.48\textwidth}
\centering
\caption{craftax/Craftax-Classic-Symbolic-AutoReset-v1}
\begin{tabular}{lcc}
\hline
Hyperparameter & eggroll & open\_es \\ \hline
activation & pqn & pqn \\
deterministic\_policy & false & false \\
learning\_rate & 0.01 & 0.001 \\
lr\_decay & 0.995 & 0.995 \\
layer\_size & 256 & 256 \\
n\_layers & 3 & 3 \\
n\_parallel\_evaluations & 4 & 8 \\
pop\_size & 2048 & 4096 \\
optimizer & sgd & adamw \\
rank & 1 & / \\
rank\_transform & false & false \\
sigma & 0.05 & 0.05 \\
sigma\_decay & 1 & 0.995 \\
\hline
\end{tabular}
\end{minipage}
\end{table}

\begin{table}[htb!]
\centering
\begin{minipage}{0.48\textwidth}
\centering
\caption{craftax/Craftax-Symbolic-AutoReset-v1}
\begin{tabular}{lcc}
\hline
Hyperparameter & eggroll & open\_es \\ \hline
activation & pqn & pqn \\
deterministic\_policy & false & false \\
learning\_rate & 0.01 & 0.1 \\
lr\_decay & 0.999 & 0.995 \\
layer\_size & 256 & 256 \\
n\_layers & 3 & 3 \\
n\_parallel\_evaluations & 1 & 4 \\
pop\_size & 512 & 1024 \\
optimizer & sgd & adam \\
rank & 4 & / \\
rank\_transform & true & false \\
sigma & 0.05 & 0.5 \\
sigma\_decay & 0.999 & 1 \\
\hline
\end{tabular}
\end{minipage}
\hfill
\begin{minipage}{0.48\textwidth}
\centering
\caption{jumanji/Game2048-v1}
\begin{tabular}{lcc}
\hline
Hyperparameter & eggroll & open\_es \\ \hline
activation & pqn & pqn \\
deterministic\_policy & false & true \\
learning\_rate & 0.1 & 0.01 \\
lr\_decay & 1 & 0.999 \\
layer\_size & 256 & 256 \\
n\_layers & 3 & 3 \\
n\_parallel\_evaluations & 4 & 4 \\
pop\_size & 1024 & 1024 \\
optimizer & adamw & adamw \\
rank & 1 & / \\
rank\_transform & false & true \\
sigma & 0.5 & 0.05 \\
sigma\_decay & 0.9995 & 0.9995 \\
\hline
\end{tabular}
\end{minipage}
\end{table}

\begin{table}[htb!]
\centering
\begin{minipage}{0.48\textwidth}
\centering
\caption{jumanji/Knapsack-v1}
\begin{tabular}{lcc}
\hline
Hyperparameter & eggroll & open\_es \\ \hline
activation & pqn & pqn \\
deterministic\_policy & false & false \\
learning\_rate & 0.1 & 0.01 \\
lr\_decay & 0.999 & 1 \\
layer\_size & 256 & 256 \\
n\_layers & 3 & 3 \\
n\_parallel\_evaluations & 4 & 1 \\
pop\_size & 1024 & 2048 \\
optimizer & sgd & adamw \\
rank & 4 & / \\
rank\_transform & true & true \\
sigma & 0.05 & 0.5 \\
sigma\_decay & 1 & 0.995 \\
\hline
\end{tabular}
\end{minipage}
\hfill
\begin{minipage}{0.48\textwidth}
\centering
\caption{jumanji/Snake-v1}
\begin{tabular}{lcc}
\hline
Hyperparameter & eggroll & open\_es \\ \hline
activation & pqn & pqn \\
deterministic\_policy & false & false \\
learning\_rate & 0.001 & 0.001 \\
lr\_decay & 0.9995 & 1 \\
layer\_size & 256 & 256 \\
n\_layers & 3 & 3 \\
n\_parallel\_evaluations & 8 & 1 \\
pop\_size & 4096 & 2048 \\
optimizer & adam & sgd \\
rank & 1 & / \\
rank\_transform & true & false \\
sigma & 0.05 & 0.2 \\
sigma\_decay & 0.9995 & 1 \\
\hline
\end{tabular}
\end{minipage}
\end{table}

\begin{table}[htb!]
\centering
\begin{minipage}{0.48\textwidth}
\centering
\caption{kinetix/l/hard\_pinball}
\begin{tabular}{lcc}
\hline
Hyperparameter & eggroll & open\_es \\ \hline
activation & pqn & pqn \\
deterministic\_policy & true & true \\
learning\_rate & 0.01 & 0.01 \\
lr\_decay & 0.995 & 1 \\
layer\_size & 256 & 256 \\
n\_layers & 3 & 3 \\
n\_parallel\_evaluations & 8 & 1 \\
pop\_size & 2048 & 512 \\
optimizer & sgd & sgd \\
rank & 4 & / \\
rank\_transform & true & true \\
sigma & 0.05 & 0.5 \\
sigma\_decay & 0.999 & 0.9995 \\
\hline
\end{tabular}
\end{minipage}
\hfill
\begin{minipage}{0.48\textwidth}
\centering
\caption{kinetix/m/h17\_thrustcontrol\_left}
\begin{tabular}{lcc}
\hline
Hyperparameter & eggroll & open\_es \\ \hline
activation & pqn & pqn \\
deterministic\_policy & false & false \\
learning\_rate & 0.1 & 0.001 \\
lr\_decay & 0.9995 & 1 \\
layer\_size & 256 & 256 \\
n\_layers & 3 & 3 \\
n\_parallel\_evaluations & 4 & 1 \\
pop\_size & 512 & 1024 \\
optimizer & sgd & adam \\
rank & 4 & / \\
rank\_transform & true & true \\
sigma & 0.5 & 0.5 \\
sigma\_decay & 1 & 0.999 \\
\hline
\end{tabular}
\end{minipage}
\end{table}

\begin{table}[htb!]
\centering
\begin{minipage}{0.48\textwidth}
\centering
\caption{kinetix/s/h1\_thrust\_over\_ball}
\begin{tabular}{lcc}
\hline
Hyperparameter & eggroll & open\_es \\ \hline
activation & pqn & pqn \\
deterministic\_policy & false & false \\
learning\_rate & 0.1 & 0.01 \\
lr\_decay & 0.995 & 0.995 \\
layer\_size & 256 & 256 \\
n\_layers & 3 & 3 \\
n\_parallel\_evaluations & 1 & 1 \\
pop\_size & 512 & 2048 \\
optimizer & adamw & sgd \\
rank & 1 & / \\
rank\_transform & true & true \\
sigma & 0.5 & 0.05 \\
sigma\_decay & 0.9995 & 1 \\
\hline
\end{tabular}
\end{minipage}
\hfill
\begin{minipage}{0.48\textwidth}
\centering
\caption{navix/Navix-DoorKey-8x8-v0}
\begin{tabular}{lcc}
\hline
Hyperparameter & eggroll & open\_es \\ \hline
activation & pqn & pqn \\
deterministic\_policy & false & false \\
learning\_rate & 0.01 & 0.01 \\
lr\_decay & 0.9995 & 1 \\
layer\_size & 256 & 256 \\
n\_layers & 3 & 3 \\
n\_parallel\_evaluations & 1 & 8 \\
pop\_size & 1024 & 2048 \\
optimizer & adamw & adam \\
rank & 1 & / \\
rank\_transform & false & true \\
sigma & 0.05 & 0.05 \\
sigma\_decay & 1 & 1 \\
\hline
\end{tabular}
\end{minipage}
\end{table}

\begin{table}[htb!]
\centering
\begin{minipage}{0.48\textwidth}
\centering
\caption{navix/Navix-Dynamic-Obstacles-6x6-Random-v0}
\begin{tabular}{lcc}
\hline
Hyperparameter & eggroll & open\_es \\ \hline
activation & pqn & pqn \\
deterministic\_policy & false & false \\
learning\_rate & 0.01 & 0.01 \\
lr\_decay & 0.999 & 1 \\
layer\_size & 256 & 256 \\
n\_layers & 3 & 3 \\
n\_parallel\_evaluations & 4 & 1 \\
pop\_size & 512 & 4096 \\
optimizer & adam & adam \\
rank & 2 & / \\
rank\_transform & false & false \\
sigma & 0.05 & 0.2 \\
sigma\_decay & 1 & 0.995 \\
\hline
\end{tabular}
\end{minipage}
\hfill
\begin{minipage}{0.48\textwidth}
\centering
\caption{navix/Navix-FourRooms-v0}
\begin{tabular}{lcc}
\hline
Hyperparameter & eggroll & open\_es \\ \hline
activation & pqn & pqn \\
deterministic\_policy & false & false \\
learning\_rate & 0.01 & 0.001 \\
lr\_decay & 0.999 & 0.9995 \\
layer\_size & 256 & 256 \\
n\_layers & 3 & 3 \\
n\_parallel\_evaluations & 4 & 4 \\
pop\_size & 2048 & 2048 \\
optimizer & sgd & adam \\
rank & 4 & / \\
rank\_transform & true & false \\
sigma & 0.05 & 0.05 \\
sigma\_decay & 0.9995 & 0.9995 \\
\hline
\end{tabular}
\end{minipage}
\end{table}

\begin{table}[htb!]
\footnotesize
\centering
\caption{PPO Hyperparameters (Set 1)}
\begin{tabular}{lcccccccc}
\hline
Hyperparameter
& CartPole & Pendulum & Ant & Humanoid
& IDP & CraftaxClassic & CraftaxSymbolic & Game2048 \\
\hline
activation & pqn & pqn & pqn & pqn & pqn & pqn & pqn & pqn \\
clip\_eps & 0.2 & 0.1 & 0.2 & 0.3 & 0.1 & 0.2 & 0.2 & 0.3 \\
ent\_coef & 0.0001 & 0.001 & 0 & 0.0001 & 0.0001 & 0.0001 & 0 & 0.001 \\
gae\_lambda & 0.9 & 0.95 & 0.95 & 0.9 & 0.98 & 0.98 & 0.9 & 0.9 \\
gamma & 0.995 & 0.999 & 0.995 & 0.95 & 0.99 & 0.95 & 0.95 & 0.99 \\
learning\_rate & 0.0003 & 0.0003 & 0.0003 & 0.0001 & 0.001 & 0.001 & 0.0003 & 0.0003 \\
max\_grad\_norm & 0.5 & 1 & 0.5 & 2 & 2 & 2 & 2 & 2 \\
layer\_size & 256 & 256 & 256 & 256 & 256 & 256 & 256 & 256 \\
n\_layers & 3 & 3 & 3 & 3 & 3 & 3 & 3 & 3 \\
normalize\_obs & true & true & true & true & true & true & true & true \\
normalize\_rew & false & false & false & false & false & false & false & false \\
num\_envs & 256 & 256 & 64 & 256 & 64 & 128 & 256 & 64 \\
num\_epochs & 4 & 16 & 8 & 4 & 4 & 4 & 4 & 8 \\
num\_minibatches & 32 & 16 & 32 & 64 & 64 & 32 & 32 & 16 \\
num\_steps & 128 & 256 & 128 & 64 & 128 & 128 & 64 & 64 \\
rew\_norm\_discount & 0.99 & 0.99 & 0.99 & 0.99 & 0.99 & 0.99 & 0.99 & 0.99 \\
skip\_initial\_eval & false & false & false & false & false & false & false & false \\
vf\_coef & 0.5 & 1 & 1 & 0.75 & 1 & 0.5 & 0.75 & 0.75 \\
\hline
\end{tabular}
\end{table}

\begin{table}[htb!]
\footnotesize

\centering
\caption{PPO Hyperparameters (Set 2)}
\begin{tabular}{lcccccccc}
\hline
Hyperparameter
& Knapsack & Snake & HardPinball & ThrustLeft
& ThrustBall & DoorKey & DynamicObs & FourRooms \\
\hline
activation & pqn & pqn & pqn & pqn & pqn & pqn & pqn & pqn \\
clip\_eps & 0.1 & 0.3 & 0.1 & 0.2 & 0.2 & 0.1 & 0.1 & 0.1 \\
ent\_coef & 0.0001 & 0.001 & 0.0001 & 0.0001 & 0.0001 & 0.0001 & 0.001 & 0.001 \\
gae\_lambda & 0.9 & 0.95 & 0.9 & 0.9 & 0.95 & 0.98 & 0.98 & 0.9 \\
gamma & 0.99 & 0.999 & 0.99 & 0.995 & 0.999 & 0.95 & 0.999 & 0.99 \\
learning\_rate & 0.0001 & 0.0001 & 0.0001 & 0.0001 & 0.0001 & 0.0003 & 0.001 & 0.001 \\
max\_grad\_norm & 0.5 & 0.5 & 1 & 2 & 0.5 & 0.5 & 1 & 1 \\
layer\_size & 256 & 256 & 256 & 256 & 256 & 256 & 256 & 256 \\
n\_layers & 3 & 3 & 3 & 3 & 3 & 3 & 3 & 3 \\
normalize\_obs & true & true & true & true & true & true & true & true \\
normalize\_rew & false & false & false & false & false & false & false & false \\
num\_envs & 256 & 128 & 256 & 256 & 64 & 64 & 128 & 256 \\
num\_epochs & 4 & 4 & 16 & 16 & 16 & 16 & 4 & 8 \\
num\_minibatches & 64 & 16 & 16 & 32 & 16 & 64 & 16 & 32 \\
num\_steps & 128 & 128 & 64 & 128 & 64 & 256 & 128 & 256 \\
rew\_norm\_discount & 0.99 & 0.99 & 0.99 & 0.99 & 0.99 & 0.99 & 0.99 & 0.99 \\
skip\_initial\_eval & false & false & false & false & false & false & false & false \\
vf\_coef & 0.75 & 0.75 & 0.5 & 0.5 & 0.5 & 0.75 & 0.5 & 0.75 \\
\hline
\end{tabular}
\end{table}
\newpage

\end{document}